\documentclass{article}

\PassOptionsToPackage{numbers}{natbib}

\usepackage[main, final]{neurips_2026}

\usepackage[utf8]{inputenc} 
\usepackage[T1]{fontenc}    
\usepackage{hyperref}       
\usepackage{url}            
\usepackage{booktabs}       
\usepackage{amsfonts}       
\usepackage{nicefrac}       
\usepackage{microtype}      
\usepackage{xcolor}         
\usepackage{graphicx}
\usepackage{multirow}
\usepackage{colortbl}
\usepackage{enumitem}

\usepackage{amsmath,amssymb,amsthm}
\usepackage{aliascnt}
\usepackage{subcaption}
\usepackage{algorithm}
\usepackage{algpseudocode}

\usepackage[capitalize,noabbrev,nameinlink]{cleveref}

\makeatletter
\AddToHook{cmd/appendix/before}{%
    \def\cref@section@alias{appendix}%
    \def\cref@subsection@alias{appendix}%
    \def\cref@subsubsection@alias{appendix}%
}
\makeatother

\usepackage{crossreftools}

\theoremstyle{plain}
\newtheorem{theorem}{Theorem}[section]
\newaliascnt{proposition}{theorem}
\newtheorem{proposition}[proposition]{Proposition}
\aliascntresetthe{proposition}
\newaliascnt{lemma}{theorem}

\aliascntresetthe{lemma}
\newaliascnt{corollary}{theorem}
\newtheorem{corollary}[corollary]{Corollary}
\aliascntresetthe{corollary}
\theoremstyle{definition}

\newtheorem{example}{Example}[section]
\newtheorem{assumption}{Assumption}[section]
\theoremstyle{remark}

\crefname{assumption}{Assumption}{Assumptions}
\crefname{theorem}{Theorem}{Theorems}
\crefname{proposition}{Proposition}{Propositions}
\crefname{lemma}{Lemma}{Lemmas}
\crefname{equation}{Equation}{Equations}
\crefname{section}{Section}{Sections}
\crefname{appendix}{Appendix}{Appendices}
\crefname{table}{Table}{Tables}
\crefname{figure}{Figure}{Figures}
\crefname{example}{Example}{Examples}
\crefname{corollary}{Corollary}{Corollaries}
\crefname{algorithm}{Algorithm}{Algorithms}
\crefname{definition}{Definition}{Definitions}

\DeclareRobustCommand{\methodPARmD}{PA-RmD}
\DeclareRobustCommand{\methodSINDy}{SINDy}
\DeclareRobustCommand{\methodNeuralODE}{Neural ODE}
\DeclareRobustCommand{\methodLEADS}{LEADS}
\DeclareRobustCommand{\methodCoDALOne}{CoDA-$\ell_1$}
\DeclareRobustCommand{\methodCoDALTwo}{CoDA-$\ell_2$}
\DeclareRobustCommand{\methodSIAR}{SIAR}

\title{Payoff-Aware Prediction of Population Game Dynamics}

\author{%
  Kang Wang$^{1}$ \quad
  Xiao Wang$^{1,2}$\thanks{Correspondence to: Renzhe Xu and Xiao Wang.} \quad
  Bo Li$^{3}$ \quad
  Renzhe Xu$^{1,2}$\footnotemark[\value{footnote}] \\
  $^{1}$School of Computing and Artificial Intelligence,
  Shanghai University of Finance and Economics \\
  $^{2}$MoE Key Laboratory of Interdisciplinary Research of Computation and Economics, \\Shanghai University of Finance and Economics\\
  $^{3}$School of Economics and Management, Tsinghua University \\
  \texttt{wangkang0330@gmail.com} \quad
  \texttt{wangxiao@sufe.edu.cn} \\
  \texttt{libo@sem.tsinghua.edu.cn} \quad
  \texttt{xurenzhe@sufe.edu.cn}
}

\begin{document}

\newcommand{\bfone}{\mathbf{1}}
\newcommand{\bbR}{\mathbb{R}}
\newcommand{\bbS}{\mathbb{S}}
\newcommand{\xt}{x^{(t)}}
\newcommand{\calT}{\mathcal{T}}
\newcommand{\calX}{\mathcal{X}}
\newcommand{\calU}{\mathcal{U}}
\newcommand{\test}{\mathrm{te}}
\newcommand{\train}{\mathrm{tr}}

\maketitle
    
\begin{abstract}
Population game dynamics describe how aggregate behavior evolves in response to payoff signals. In many applications, the payoff function may change, and the goal is to predict the population trajectory induced by a new payoff design before trajectory data under that design are available. We study this problem as \textbf{payoff-aware prediction of population game dynamics}. Standard dynamics learning methods fit the motion observed under training payoffs, but do not separate the payoff-dependent incentive from the payoff-independent response structure needed for prediction under new payoffs. Building on the existing framework of Riemannian game dynamics, we propose \methodPARmD{}, a payoff-aware learning method that keeps payoff information explicit and learns the payoff-independent response structure from data. Theoretically, we show that the learned model preserves fixed-payoff Nash equilibria and derive finite-time one-step prediction bounds under shared-response and payoff-coverage conditions. Empirically, we evaluate \methodPARmD{} on synthetic zero-sum and potential games, as well as a YouTube Trending multi-country dataset, showing that our method performs well under changing payoff designs.
\end{abstract}

\section{Introduction}

Population games study systems with many agents who repeatedly choose among a finite set of strategies. In such games, a payoff function specifies the benefit agents receive from each strategy, and the associated population game dynamics describe how the aggregate behavior of the population changes under these payoffs. This viewpoint is standard in large-population models of congestion systems, markets, and other settings in which individual identities are less important than aggregate behavior~\citep{sandholm2010population,hadikhanloo2022learning,quijano2017role}. The same aggregate perspective has also been adopted in models of online content platforms~\citep{xu2023competing,kiyohara2025policy}, where content creators repeatedly adjust what content to produce in response to platform-provided payoff signals, such as exposure or monetary rewards~\citep{yao2023bad,dean2025accounting,wang2026lower,eilat2023performative}.

Predicting such population game dynamics can provide important guidance for decision-making and mechanism design. In many design settings, the payoff function is not fixed: a planner or platform may alter the incentives faced by the population, and these changes can substantially reshape the induced population dynamics. Prediction in this setting must therefore be payoff-aware: the model should take a candidate payoff function as input and forecast the population trajectory it would induce. We refer to this problem as \textbf{payoff-aware prediction of population game dynamics}. The challenge is that the prediction is \emph{counterfactual}. For example, a content platform may want to know whether a new reward mechanism would shift creators toward different content categories and how this shift would affect aggregate platform outcomes. Before deployment, however, it has no trajectory data under that new mechanism. The operational question is therefore how to predict the trajectory that would arise under a new payoff function using only trajectories observed under previous payoff functions.

Existing work provides important context but does not directly address this prediction problem. The population game dynamics literature primarily studies specified dynamics under a given payoff function, with central questions including convergence and stability~\citep{sandholm2010population}. Data-driven work has studied learning, identifying, and steering game dynamics from observed play~\citep{sakos2024data,canyakmaz2025learning}, while general dynamical-system learning methods learn vector fields or governing equations from trajectory data~\citep{chen2018neural,brunton2016discovering}. These approaches can learn dynamics from observed trajectories, but they typically fit the realized motion under the observed payoff conditions. As a result, the learned dynamics may mix the payoff-dependent incentive with the population response structure, leaving no principled way to update the prediction when the payoff function changes.

To avoid this issue, we turn to structured population-game dynamics, where different dynamics encode different assumptions about how populations respond to payoffs~\citep{sandholm2010population}. Among these dynamics, Riemannian game dynamics (RmD)~\citep{mertikopoulos2018riemannian} provide a unified framework that includes commonly used models such as replicator dynamics~\citep{taylor1978evolutionary} and projection dynamics~\citep{nagurney1997projected}. RmD's key idea is to model population motion as the outcome of balancing a \emph{payoff-dependent gain from motion} against a \emph{payoff-independent cost from motion}. This gain-cost view is natural in platform settings \citep{xu2023competing,eilat2023performative,dean2025accounting}: creators may be attracted to content categories with higher exposure or rewards, but switching categories can be costly. These adjustment costs can often be viewed as stable across payoff designs because they reflect practical barriers such as learning new production skills~\citep{farrell2007coordination}. Motivated by this relative stability and following the RmD framework, we model the cost term as payoff-independent and use it as the shared population response structure for payoff-aware prediction. Once this cost term is learned from historical trajectories, a new payoff function can be plugged into the gain term to predict the induced population motion. We therefore propose \methodPARmD{}, a payoff-aware Riemannian dynamics learning model that implements this gain-cost decomposition from data.

We analyze the proposed \methodPARmD{} method both theoretically and empirically. On the theory side, we show that the learned model preserves the Nash equilibrium set of a fixed payoff function when all strategies retain positive population mass, and we prove finite-time one-step prediction bounds under a shared population response assumption. These bounds identify a payoff-coverage condition: prediction under a held-out payoff is controlled when its payoff direction is represented by the payoff directions observed in training. Empirically, we evaluate the framework on synthetic zero-sum games and potential games, as well as on a YouTube Trending multi-country dataset. The results show that the payoff-aware formulation can support prediction across held-out payoff designs.

To conclude, our main contributions are as follows:
\begin{itemize}[leftmargin=*, itemsep=0.15em, topsep=0.15em]
    \item We formulate payoff-aware prediction of population game dynamics: predicting population trajectories under held-out payoff designs from trajectories observed under previous payoff designs.
    \item We instantiate this formulation with \methodPARmD{}, a RmD-based framework that keeps payoff functions explicit and learns the shared population response structure from data.
    \item We show that the learned model preserves fixed-payoff Nash equilibria and prove finite-time one-step prediction bounds under shared population response and payoff-coverage conditions.
    \item We demonstrate the effectiveness of the framework on synthetic zero-sum games and potential games and a YouTube Trending multi-country dataset.
\end{itemize}

\section{Related Work}

\paragraph{Population games and dynamics.}
Population games provide a standard framework for modeling strategic interaction in large populations of anonymous agents through aggregate strategy shares and state-dependent payoffs~\citep{sandholm2010population}. This population-level viewpoint has been used in congestion and routing systems~\citep{nagurney1997projected,sandholm2015population}, distributed control~\citep{quijano2017role,barreiro2016constrained}, and evolutionary and biological systems~\citep{taylor1978evolutionary,hofbauer1998evolutionary}. We adopt the same aggregate perspective because our goal is to predict how population behavior changes when the payoff environment changes.

Game dynamics specify how a population or a set of players adapts over time in response to payoff information. In population games, these dynamics are often formalized as revision protocols that convert current payoff vectors into feasible motion. Canonical examples include replicator, projection, logit, and Riemannian game dynamics~\citep{taylor1978evolutionary,nagurney1997projected,mertikopoulos2016learning,mertikopoulos2018riemannian,hadikhanloo2022learning}. More broadly, dynamics in games are also used as algorithmic and conceptual tools for studying decentralized adaptation and equilibrium computation~\citep{papadimitriou2019game,hart2000simple,syrgkanis2015fast,freund1999adaptive}. Prior work typically starts from a specified update rule and studies its stationarity, stability, convergence, regret, or ability to accelerate equilibrium computation~\citep{sandholm2015population,fudenberg1998theory,harris2023metalearning}. In contrast, we treat the dynamic itself as the object to be learned from trajectory data, with the goal of predicting population-level evolution under held-out payoff environments.

\paragraph{Learning dynamical systems from data.}
Data-driven system identification learns dynamics from trajectories, usually as equations, vector fields, or transition maps. Sparse Identification of Nonlinear Dynamics discovers interpretable equations from candidate libraries~\citep{brunton2016discovering}, while neural ODEs parameterize continuous-time vector fields with neural networks~\citep{chen2018neural}. Other methods incorporate structure for particle systems~\citep{bunne2022proximal}, fluid dynamics~\citep{li2021fourier}, social networks~\citep{trivedi2019dyrep}, and linear dynamical systems~\citep{Ma,YanxiChen,Wang2024,Max}; multi-environment dynamics-learning methods study generalization across changing systems or latent environments~\citep{yin2021leads,kirchmeyer2022coda,liu2025environment}.

Recent work has also learned dynamics or payoff information from game data. \citet{lam2016learning} estimate learning dynamics in routing games, and the \methodSIAR{} framework~\citep{sakos2024data} identifies game dynamics from scarce trajectories using polynomial vector fields with side-information constraints. SIAR-MPC further combines learned controlled dynamics with model predictive control to steer game dynamics toward desirable outcomes~\citep{canyakmaz2025learning}. Relatedly, \citet{bonafini2023data} recover individual payoff functions from observed evolutionary trajectories. These works infer dynamics, learning rules, or payoffs within an observed game environment. Our task is different: given a new payoff function and no trajectory under it, the model must predict aggregate evolution from a learned population response.

\section{Preliminaries}
\paragraph{Notations}
For a positive integer $K$, let $[K]=\{1,\ldots,K\}$ and let $\bfone$ denote the all-ones vector in $\bbR^K$. We write $\Delta_K=\{x\in\bbR_{\ge 0}^K:\bfone^\top x=1\}$ for the simplex, $\Delta_K^\circ=\{x\in\bbR_{>0}^K:\bfone^\top x=1\}$ for its interior, and $\calT=\{z\in\bbR^K:\bfone^\top z=0\}$ for the tangent space of the simplex. For $d\ge 1$, $\bbS_+^d$ and $\bbS_{++}^d$ denote the sets of $d\times d$ symmetric positive semidefinite and positive definite matrices, respectively. We use $\|\cdot\|_2$ for the Euclidean norm and $\|\cdot\|_{\mathrm{op}}$ for the operator norm.

\subsection{Population Games and Dynamics}
We consider finite-strategy population games with strategy set $[K]$. A population state is a vector $x\in\Delta_K$, where $x_i$ denotes the mass, or share, of the population using strategy $i$. A population game is specified by a payoff function $v:\Delta_K\to\bbR^K$, where $v_i(x)$ is the payoff received by an agent choosing strategy $i$ when the aggregate population state is $x$. Let $\calU$ denote the set of payoff functions.

A population dynamic specifies how the population state changes in response to the current payoff vector. On a finite time interval $I=[0,T]$, we write the induced trajectory as
\begin{equation}\label{eq:population-dynamic}
    \dot x_t = F(x_t, v(x_t)),
\end{equation}
where $F(x_t,v(x_t))\in\calT$ and $\calT$ is the tangent space of the simplex. Thus, the dynamic converts the payoff vector at the current state into a feasible direction of motion for the population state.

We write $\mathcal X\subseteq\Delta_K$ for the reachable population-state domain generated by the dynamics under payoff functions $v\in\calU$ and admissible initial conditions. Throughout the paper, we restrict attention to interior trajectories and assume that $\calX\subseteq\Delta_K^\circ$.

For a payoff function $v:\mathcal X\to\bbR^K$, define its Nash equilibrium set on the interior domain $\mathcal X$ as
\begin{equation}
\label{eq:nash-set}
    \mathcal N(v)
    =
    \{x\in\mathcal X:v_i(x)=v_j(x),\ \forall i,j\in[K]\}.
\end{equation}
Indeed, the standard population-game Nash condition requires every strategy used with positive mass to attain maximal payoff. Since every strategy has positive mass on $\mathcal X\subseteq\Delta_K^\circ$, this condition is equivalent to payoff indifference across all strategies.

\subsection{Payoff-Aware Prediction of Population Game Dynamics} \label{sec:prediction-problem}
We observe population trajectories generated under multiple payoff functions. For each training instance $n\in[N]$, a payoff function $v^{(n)}\in\calU$ induces a trajectory $x_t^{(n)}\in\calX$ over a finite time interval $[0,T]$. The trajectories are observed at sampling interval $\delta>0$, with observation grid $\hat I_\delta=\{0,\delta,2\delta,\ldots,\lfloor T/\delta\rfloor\delta\}$. The training data are a collection of observed trajectories, each containing the population states and the corresponding payoff vectors along the trajectory: $\mathcal D_{\mathrm{obs}}=\{\{(x_t^{(n)},v^{(n)}(x_t^{(n)}))\}_{t\in\hat I_\delta}:n\in[N]\}$. For simplicity, we write the setup with a common observation grid across trajectories; the same formulation extends directly to trajectory-specific observation times.

At prediction time, we are given a payoff function $w\in\calU$ and an initial state $x_0^\test\in\calX$. The payoff function $w$ may differ from those observed during training. The goal is to predict the population trajectory $x_t^\test$ induced by $w$ from the initial state $x_0^\test$. Thus the learning problem is payoff-aware: the payoff function is kept as an explicit input, and the model learns how population states respond to payoff vectors from the observed trajectories.

\begin{example}[Online content platform] \label{exam:platform}
Consider an online content platform, such as TikTok or YouTube, on which creators choose among $K$ content categories. The state $x_i$ is the share of creators producing category $i$ content, and $v_i(x)$ is the expected payoff to producing category $i$ when the platform-wide content distribution is $x$. Each training payoff function $v^{(n)}$ corresponds to a historical ranking, exposure, or reward rule, and the observed trajectory $x_t^{(n)}$ records how creators reallocate across categories under that rule. At prediction time, $w$ represents the platform rule used for forecasting. A payoff-aware model learns the shared adjustment structure from historical trajectories and combines it with $w$ to predict the creator population trajectory from the initial state $x_0^\test$.
\end{example}

\section{Learning Payoff-Aware Riemannian Dynamics} \label{sec:model}
\subsection{Learning Procedure}
To address the payoff-aware prediction problem defined in \cref{sec:prediction-problem}, the model must explain how population motion changes when the payoff function changes. A model that only fits the realized motion in observed trajectories can mix the payoff-dependent incentive with the response rule of the population, leaving no clear way to update predictions under a new payoff function.

We therefore use the structure of population-game dynamics. Different dynamics encode different assumptions about how agents revise strategies in response to payoffs~\citep{sandholm2010population}; equivalently, the payoff function alone does not determine population motion, because the response rule is part of the model~\citep{papadimitriou2019game}. In our setting, the goal is to keep the payoff function explicit while learning the part of the population response that can be shared across payoff designs.

\paragraph{Payoff-aware Riemannian dynamics learning model.}
Riemannian game dynamics (RmD)~\citep{mertikopoulos2018riemannian} provide a natural formalism for this goal. We define a payoff-aware Riemannian dynamics learning model, denoted by \methodPARmD{}. The model keeps the same structural form as RmD, but parameterizes the unknown shared cost-from-motion structure by $G_\theta:\calX\to\bbS^K_{++}$ and learns it from observed trajectories. Specifically, for any payoff function $u\in\calU$, the induced population dynamic is
\begin{equation} \label{eq:argmax}
    \dot x_t = \underset{z \in \calT}{\operatorname{argmax}} \left\{u(x_t)^\top z - \frac{1}{2} z^\top G_{\theta}(x_t) z\right\}.
\end{equation}
The parameter $\theta$ is shared across payoff functions, while the payoff function $u$ remains an explicit input. Thus, \methodPARmD{} does not learn a separate dynamic for each payoff function; it learns the payoff-independent adjustment structure that converts payoff incentives into population motion.

\paragraph{Interpretation.}
The variational problem in \cref{eq:argmax} makes the gain-cost decomposition explicit. The linear term $u(x_t)^\top z$ is the \emph{gain from motion}: a direction $z$ is favored when it increases the mass of strategies with high current payoffs and decreases the mass of strategies with low current payoffs. The quadratic term $\frac{1}{2}z^\top G_\theta(x_t)z$ is the \emph{cost from motion}: the matrix $G_\theta(x_t)$ assigns a state-dependent cost to each feasible direction in $\calT$. In \methodPARmD{}, this payoff-independent cost term is the shared population response structure to be learned from data. It can represent adjustment barriers, responsiveness to payoff differences, and constraints on reallocation. In the content-platform example, such costs capture how difficult it is for creators to shift toward a newly rewarded content category, for instance because doing so may require learning new production skills or rebuilding an audience.

Riemannian game dynamics also admit an individual-level interpretation through revision protocols, in which agents revise strategies according to payoff-dependent switching rates and the aggregate inflows and outflows induce the population dynamic. Under this interpretation, $G_\theta$ encodes how agents adjust behavior at the individual level, rather than merely a formal geometric object. We give the details in \Cref{sect:microfoundations}.

\paragraph{Equivalent parametrization.}
The optimization problem in \cref{eq:argmax} admits an equivalent form that is easier to use for learning. Fix an orthonormal basis $q_1,\dots,q_{K-1}$ of $\calT$ and let
\[
U=[q_1,\dots,q_{K-1}]\in\bbR^{K\times (K-1)}.
\]
\begin{proposition} \label{prop:optimization}
Fix $U$ as above. The Riemannian model in \cref{eq:argmax} is equivalent to
\begin{equation} \label{eq:c-param}
    \dot x_t = U C_\theta(x_t)U^\top u(x_t),
\end{equation}
where $C_\theta:\calX\to\bbS^{K-1}_{++}$. Specifically, any metric $G_\theta:\calX\to\bbS^K_{++}$ induces this form with $C_\theta(x)=(U^\top G_\theta(x)U)^{-1}$.
Conversely, any $C_\theta:\calX\to\bbS^{K-1}_{++}$ can be represented by the Riemannian model in \cref{eq:argmax}; one compatible choice is $G_\theta(x)=UC_\theta(x)^{-1}U^\top+c\bfone\bfone^\top/K$, for any scalar $c>0$.
\end{proposition}
Thus, instead of learning the full metric $G_\theta$, we learn $C_\theta$ directly. The parametrization in \Cref{eq:c-param} is convenient because it replaces the constrained optimization over feasible simplex directions in \cref{eq:argmax} with a matrix-vector product.

\paragraph{Loss function.}
Let $\hat I_\delta^{-}=\hat I_\delta\setminus\{\lfloor T/\delta\rfloor\delta\}$ denote the observation times that admit a one-step successor. We learn $\theta$ by matching the finite-difference approximation of the time derivative with the motion predicted by \methodPARmD{}:
\begin{equation}\label{eq:loss}
\ell(\theta) = \sum_{n=1}^N \sum_{t \in \hat I_\delta^{-}} \left\|\frac{x_{t+\delta}^{(n)}-x_t^{(n)}}{\delta} - U C_\theta(x_t^{(n)}) U^\top v^{(n)}(x_t^{(n)})\right\|_2^2,
\end{equation}
with $C_\theta(x)\in\bbS^{K-1}_{++}$ for all $x\in\calX$.

\subsection{Theoretical Analysis}

For the theoretical analysis, we impose a stronger interiority condition.
\begin{assumption} \label{assum:boundedness}
    Assume that $\calX \subset \Delta_K^\circ$ is a compact set.
\end{assumption}
This assumption focuses the analysis on regimes in which every strategy retains a non-zero population share. This interior-domain restriction is standard in population and evolutionary game dynamics, and is natural in large-population applications where each action or category can continue to attract some agents even when its population share is small \citep{sandholm2010population,mckelvey1995quantal,goeree2005regular}.

\paragraph{Equilibrium preservation.}
We first provide a structural property of the model: for a fixed payoff function, the matrix $C$ affects the transient motion but not the set of interior stationary states. Following the population-dynamic notation in \cref{eq:population-dynamic}, for a population dynamic $F$ and a payoff function $u$, define the stationary set induced by $u$ as
\[
    \mathcal S(F,u)
    =
    \{x\in\mathcal X:F(x,u(x))=0\}.
\]
These are the states at which the population dynamic produces zero motion.

\begin{proposition}
\label{prop:nash-equilibrium-preservation}
Suppose \cref{assum:boundedness} holds. Fix a payoff function $u\in \calU$. For any continuous $\widetilde C:\mathcal X\to\bbS_{++}^{K-1}$, define $F_{\widetilde C}(x,a) = U\widetilde C(x)U^\top a$. Then
\[
    \mathcal S(F_{\widetilde C},u)
    =
    \mathcal N(u)
    =
    \{x\in\mathcal X:U^\top u(x)=0\}.
\]
Here $\mathcal N(u)$ denotes the Nash equilibrium set defined in \cref{eq:nash-set}.
\end{proposition}
\Cref{prop:nash-equilibrium-preservation} shows that changing $\widetilde C$ changes how the population moves, but the stationary states are still exactly the states at which all strategies receive the same payoff. Learning $\widetilde C$ therefore does not create or remove interior Nash equilibria for the same payoff function.

\paragraph{Prediction under changing payoff functions.}
We next study when the matrix learned from training payoff functions can be reused for a prediction-time payoff function. The key assumption is that payoff functions may change across tasks, but the same $C_\star$ converts payoff vectors into motion.

\begin{assumption}
\label{assum:population-learning}
There exists $C_\star:\mathcal X\to\bbS_+^{K-1}$ such that, for every payoff function $u\in\calU$, the population state evolves according to $\dot x_t=UC_\star(x_t)U^\top u(x_t)$. For each $x\in\mathcal X$, let $\mathcal D_{\mathrm{tr}}(\cdot\mid x)$ denote a conditional distribution over training payoff functions observed at $x$.
\end{assumption}
The first part of \cref{assum:population-learning} says that payoff functions may change across tasks, but $C_\star$ does not. The second part is a population-risk idealization of the training data, where sufficiently many one-step observations near the same state are represented by $\mathcal D_{\mathrm{tr}}(\cdot\mid x)$.

Let $\Psi_\delta^u(x)$ denote the time-$\delta$ flow map generated by the dynamics in \cref{assum:population-learning}. Because our training objective uses one-step finite differences, the population-risk target at sampling interval $\delta$ is the statewise minimizer
\[
    C_{\delta,\mathrm{tr}}^\star(x)
    \in
    \arg\min_{C\in\mathbb S_+^{K-1}}
    \mathbb E_{v\sim\mathcal D_{\mathrm{tr}}(\cdot\mid x)}
    \left[
    \left\|
    \frac{1}{\delta}
    \left(
    \Psi_\delta^v(x)-x
    \right)
    -
    U C U^\top v(x)
    \right\|_2^2
    \right].
\]
Equivalently, $C_{\delta,\mathrm{tr}}^\star(x)$ is the matrix that minimizes the conditional squared loss one would obtain with infinitely many one-step training samples starting from the same state $x$. We optimize over positive semidefinite matrices, rather than positive definite matrices, to ensure that such a population-risk minimizer exists. We also impose standard boundedness and Lipschitz regularity conditions on the payoff functions and the shared matrix $C_\star$.

\begin{assumption} \label{assum:regularity}
    Assume that there exist constants $\Lambda, L_u, M, L_C > 0$ such that for every payoff function $u \in \calU $ and every $x,y\in\mathcal X$, $\|u(x)\|_2\le \Lambda$, $\|u(x)-u(y)\|_2\le L_u\|x-y\|_2$, $\|C_\star(x)\|_{\mathrm{op}}\le M$, and $\|C_\star(x)-C_\star(y)\|_{\mathrm{op}}\le L_C\|x-y\|_2$.
\end{assumption}

The next result says that this population-risk target transfers to a prediction-time payoff function $w$ when the projected payoff direction $U^\top w(x)$ is covered by the projected training payoff directions $U^\top v(x)$. The coefficient $\gamma_w$ measures how hard it is to represent the prediction-time payoff direction from the training distribution.

\begin{theorem}
\label{thrm:finite-time-target-transfer-directional}
Suppose \cref{assum:boundedness,assum:population-learning,assum:regularity} hold (with the constants $M,L_C,L_u,\Lambda$). Fix a prediction-time payoff function $w\in\calU$. Suppose there exists $\gamma_w<\infty$ such
that, for every $x\in\mathcal X$, there exists a square-integrable
coefficient $\rho_x(v)$ satisfying
\[
    U^\top w(x)
    =
    \mathbb E_{v\sim\mathcal D_{\mathrm{tr}}(\cdot\mid x)}
    \left[
    \rho_x(v)U^\top v(x)
    \right] \quad \text{and} \quad \mathbb E_{v\sim\mathcal D_{\mathrm{tr}}(\cdot\mid x)}
    \left[
    \rho_x(v)^2
    \right]
    \le
    \gamma_w^2.
\]
Then
\[
\begin{aligned}
    \sup_{x\in\mathcal X}
    \left\|
    \Psi_\delta^w(x)-x
    -
    \delta U C_{\delta,\mathrm{tr}}^\star(x)U^\top w(x)
    \right\|_2
    \le
    B_\star(1+\gamma_w)\delta^2 ,
\end{aligned}
\]
where
\begin{equation}\label{eq:B-star}
    B_\star
    =
    \frac12
    M\Lambda
    \left(
    L_C\Lambda+ML_u
    \right).
\end{equation}
\end{theorem}
\Cref{thrm:finite-time-target-transfer-directional} compares the true one-step displacement $\Psi_\delta^w(x)-x$ under $w$ with the one-step displacement predicted by the training target $C_{\delta,\mathrm{tr}}^\star(x)$. The error is order $\delta^2$: for small sampling intervals, the learned matrix gives an accurate first-order prediction, and the constant worsens when the prediction-time payoff direction is poorly covered by training directions.

\begin{corollary}
\label{coro:finite-time-target-transfer}
Suppose \cref{assum:boundedness,assum:population-learning,assum:regularity} hold (with the constants $M,L_C,L_u,\Lambda$). Suppose the projected training payoff vectors uniformly cover the tangent space on $\mathcal X$: there exists $\mu>0$ such that, for every
$x\in\mathcal X$,
\[
    \Sigma_{\mathrm{tr}}(x)
    :=
    \mathbb E_{v\sim\mathcal D_{\mathrm{tr}}(\cdot\mid x)}
    \left[
    U^\top v(x)v(x)^\top U
    \right]
    \succeq
    \mu I_{K-1}.
\]

Then, for any prediction-time payoff function $w \in \calU$,
\[
\begin{aligned}
    \sup_{x\in\mathcal X}
    \left\|
    \Psi_\delta^w(x)-x
    -
    \delta U C_{\delta,\mathrm{tr}}^\star(x)U^\top w(x)
    \right\|_2
    \le
    B_\star 
    \left(
    1+\frac{\Lambda}{\sqrt{\mu}}
    \right)
    \delta^2,
\end{aligned}
\]
where $B_\star$ is defined in \cref{eq:B-star}.
\end{corollary}
\Cref{coro:finite-time-target-transfer} gives a simpler sufficient condition for the coverage requirement in \Cref{thrm:finite-time-target-transfer-directional}. If the training payoff functions span all tangent payoff directions uniformly, then every prediction-time payoff function in $\calU$ satisfies the theorem with a controlled constant.

\subsection{Practical Implementations}
In implementation, we enforce the positive-definiteness constraint in \cref{eq:loss} on $C_\theta$ by parameterizing
\[
C_{\theta}(x) = L_{\theta}(x) L_{\theta}(x)^\top + \epsilon I
\]
where $L_{\theta}(x)\in\bbR^{(K-1)\times(K-1)}$ is the output of a neural network, $\epsilon>0$ is fixed, and $I$ is the identity matrix. We then optimize the loss in the learning procedure by standard gradient-based methods. This parametrization keeps every predicted motion in the form $UC_\theta(x)U^\top v(x)$ while ensuring $C_\theta(x)\in\bbS_{++}^{K-1}$ throughout training.

\section{Experiments} \label{sec:experiments}

We evaluate the payoff-aware framework in both synthetic and real-data experiments. The source code is publicly available at \url{https://github.com/wangkng/PA-RmD}.

\paragraph{Baselines.}
We compare \methodPARmD{} against three groups of baselines: (i) standard dynamics prediction methods, including \methodSINDy{} \citep{brunton2016discovering} and \methodNeuralODE{} \citep{chen2018neural}; (ii) methods designed for dynamics prediction under distribution shift, including \methodLEADS{} \citep{yin2021leads}, \methodCoDALOne{}, and \methodCoDALTwo{} \citep{kirchmeyer2022coda}\footnote{%
Note that \methodLEADS{} and CoDA require ground-truth environment labels during training, whereas \methodPARmD{} does not. At test time, to align with our prediction task without target-specific adaptation, we adopt the \emph{Pre-trained-$f$-Only} variant of \methodLEADS{} \citep{yin2021leads} and the zero-shot mean-parameter protocol for CoDA following \citet{li2026imooe}. Comparisons between their few-shot variants and zero-shot \methodPARmD{} are provided in \Cref{app:few-shot-adaptation}.
}; and (iii) a game-structured dynamics identification method, \methodSIAR{} \citep{sakos2024data}. To ensure a fair comparison under the payoff-aware prediction setup in \Cref{sec:prediction-problem}, all baselines are given the same payoff function information as input. Implementation details are provided in \Cref{app:baselines}.

\subsection{Synthetic Experiments}
\label{sec:exp-synthetic}
\paragraph{Data-generating process.}
Each synthetic setting pairs a payoff function family with a population response dynamic; construction details are given in \Cref{app:synthetic-detail}. We use two payoff function classes, denoted by $\calU^{\mathrm{cyclic}}$ and $\calU^{\mathrm{congestion}}$.

\begin{enumerate}[leftmargin=2em, itemsep=0.15em, topsep=0.15em]
    \item \textbf{Zero-sum cyclic game class}~\citep{szolnoki2014cyclic}.
    This class generalizes rock--paper--scissors interactions and cyclic dominance in evolutionary games. Each payoff function in $\calU^{\mathrm{cyclic}}$ is a linear function $v(x)=Ax$ on the simplex. Starting from a directed cyclic dominance rule over $K$ strategies, we construct different payoff functions by permuting the strategy labels in $A$. Thus, all payoff functions share the same zero-sum cyclic structure, but differ in which strategy labels win or lose against one another. For $K=3$, this gives exactly two payoff functions: the usual rock--paper--scissors rule and the reversed rule, where every winning relation is flipped.

    \item \textbf{Potential congestion game class}~\citep{rosenthal1973class}.
    Each payoff function in $\calU^{\mathrm{congestion}}$ comes from a congestion game with a fixed route--resource incidence structure. The $K$ population strategies are routes. A population state specifies the mass assigned to each route and thereby determines the load on each resource. Each route receives a payoff equal to the negative total cost of the resources it uses. We construct different payoff functions by changing the affine cost parameters of the resources while keeping the route--resource structure fixed. Thus, all payoff functions share the same routes and resources, but differ in how costly each resource becomes as congestion increases.
\end{enumerate}

We combine each payoff family with one of three response dynamics: replicator dynamics~\citep{taylor1978evolutionary}, projection dynamics~\citep{nagurney1997projected}, and multiplicative weights update (MWU)~\citep{arora2012mwu}. These dynamics cover complementary regimes for evaluating payoff-aware prediction. Replicator and projection dynamics are canonical continuous-time population dynamics and correspond to different motion-cost structures within the RmD framework~\citep{mertikopoulos2018riemannian}. MWU is included as a widely used discrete-time learning rule in games; although it is outside the continuous-time RmD model used in our theory, it serves as a useful stress test for whether the learned gain-cost structure remains effective beyond the exact model class. For each payoff family $\calU\in\{\calU^{\mathrm{cyclic}},\calU^{\mathrm{congestion}}\}$ and each response dynamic, we split $\calU$ into training payoff functions $\calU^\train$ and held-out payoff functions $\calU^\test$. We then generate training trajectories from payoff functions in $\calU^\train$ and test trajectories from payoff functions in $\calU^\test$, using the same response dynamic within the setting.

\begin{table}[t]
\renewcommand{\underline}[1]{\begingroup\setbox0=\hbox{#1}\leavevmode\vtop{\offinterlineskip\copy0\kern0.25pt\hrule height0.3pt width\wd0}\endgroup}
\centering
\caption{Synthetic held-out results under replicator dynamics. Each entry is mean $\pm$ std over five runs on held-out payoff functions. In all tables, bold and underlined entries indicate the best and second-best results for each setting and metric, respectively.}
\label{tab:synthetic-replicator}
\scriptsize
\renewcommand{\arraystretch}{1.12}
\setlength{\tabcolsep}{3pt}
\resizebox{\linewidth}{!}{%
\begin{tabular}{c|c|ccc|ccc}
\hline
\multirow{2}{*}{Game} & \multirow{2}{*}{Method} & \multicolumn{3}{c|}{$K=3$} & \multicolumn{3}{c}{$K=4$} \\
\cline{3-8}
& & One-step ($\downarrow$) & Rollout RMSE ($\downarrow$) & Rollout@$T$ ($\downarrow$) & One-step ($\downarrow$) & Rollout RMSE ($\downarrow$) & Rollout@$T$ ($\downarrow$) \\
\hline
\multirow{7}{*}{\begin{tabular}[c]{@{}c@{}}Zero-sum\\cyclic\end{tabular}} & \methodSINDy{} & (4.05 $\pm$ 0.00)E-2 & (4.04 $\pm$ 0.00)E-1 & (5.23 $\pm$ 0.00)E-1 & (1.09 $\pm$ 0.00)E-1 & (2.88 $\pm$ 0.00)E-1 & (3.07 $\pm$ 0.00)E-1 \\
 & \methodSIAR{} & (5.03 $\pm$ 0.00)E-2 & (3.05 $\pm$ 0.00)E-1 & (3.99 $\pm$ 0.00)E-1 & (5.84 $\pm$ 0.00)E-2 & (2.11 $\pm$ 0.00)E-1 & (2.07 $\pm$ 0.00)E-1 \\
 & \methodNeuralODE{} & (1.45 $\pm$ 0.05)E-2 & (1.49 $\pm$ 0.16)E-1 & (2.09 $\pm$ 0.28)E-1 & \underline{(7.07 $\pm$ 0.20)E-3} & \underline{(8.72 $\pm$ 0.62)E-2} & \underline{(1.20 $\pm$ 0.07)E-1} \\
 & \methodLEADS{} & (1.67 $\pm$ 0.18)E-2 & (1.75 $\pm$ 0.30)E-1 & (2.62 $\pm$ 0.56)E-1 & (7.80 $\pm$ 1.24)E-3 & (9.71 $\pm$ 1.56)E-2 & (1.31 $\pm$ 0.16)E-1 \\
 & \methodCoDALOne{} & (1.43 $\pm$ 0.02)E-2 & (1.42 $\pm$ 0.13)E-1 & (2.05 $\pm$ 0.29)E-1 & (7.36 $\pm$ 0.98)E-3 & (9.38 $\pm$ 0.94)E-2 & (1.26 $\pm$ 0.15)E-1 \\
 & \methodCoDALTwo{} & \underline{(1.43 $\pm$ 0.02)E-2} & \underline{(1.41 $\pm$ 0.12)E-1} & \underline{(2.04 $\pm$ 0.28)E-1} & (7.34 $\pm$ 1.01)E-3 & (9.34 $\pm$ 0.99)E-2 & (1.25 $\pm$ 0.16)E-1 \\
\rowcolor{gray!15}
\cellcolor{white} & \methodPARmD{} & \textbf{(1.29 $\pm$ 0.05)E-3} & \textbf{(2.12 $\pm$ 0.73)E-2} & \textbf{(2.32 $\pm$ 0.52)E-2} & \textbf{(9.81 $\pm$ 0.34)E-4} & \textbf{(1.73 $\pm$ 0.09)E-2} & \textbf{(1.88 $\pm$ 0.09)E-2} \\
\hline
\multirow{7}{*}{\begin{tabular}[c]{@{}c@{}}Potential\\congestion\end{tabular}} & \methodSINDy{} & (3.32 $\pm$ 0.00)E-3 & (4.81 $\pm$ 0.00)E-2 & (7.19 $\pm$ 0.00)E-2 & (5.58 $\pm$ 0.00)E-3 & (7.07 $\pm$ 0.00)E-2 & (9.91 $\pm$ 0.00)E-2 \\
 & \methodSIAR{} & (7.15 $\pm$ 0.00)E-3 & (8.77 $\pm$ 0.00)E-2 & (1.38 $\pm$ 0.00)E-1 & (4.59 $\pm$ 0.00)E-3 & (5.17 $\pm$ 0.00)E-2 & (7.52 $\pm$ 0.00)E-2 \\
 & \methodNeuralODE{} & \underline{(1.62 $\pm$ 0.13)E-3} & (2.07 $\pm$ 0.18)E-2 & \underline{(2.90 $\pm$ 0.19)E-2} & \underline{(1.84 $\pm$ 0.06)E-3} & \underline{(2.21 $\pm$ 0.07)E-2} & \underline{(3.41 $\pm$ 0.12)E-2} \\
 & \methodLEADS{} & (1.63 $\pm$ 0.12)E-3 & \underline{(2.07 $\pm$ 0.14)E-2} & (2.92 $\pm$ 0.20)E-2 & (2.05 $\pm$ 0.15)E-3 & (2.42 $\pm$ 0.17)E-2 & (3.79 $\pm$ 0.32)E-2 \\
 & \methodCoDALOne{} & (1.63 $\pm$ 0.07)E-3 & (2.10 $\pm$ 0.11)E-2 & (2.92 $\pm$ 0.10)E-2 & (1.97 $\pm$ 0.17)E-3 & (2.38 $\pm$ 0.19)E-2 & (3.66 $\pm$ 0.34)E-2 \\
 & \methodCoDALTwo{} & (1.63 $\pm$ 0.07)E-3 & (2.10 $\pm$ 0.11)E-2 & (2.92 $\pm$ 0.10)E-2 & (1.97 $\pm$ 0.18)E-3 & (2.38 $\pm$ 0.20)E-2 & (3.66 $\pm$ 0.35)E-2 \\
\rowcolor{gray!15}
\cellcolor{white} & \methodPARmD{} & \textbf{(3.48 $\pm$ 0.31)E-4} & \textbf{(5.83 $\pm$ 0.28)E-3} & \textbf{(6.95 $\pm$ 0.55)E-3} & \textbf{(3.41 $\pm$ 0.33)E-4} & \textbf{(5.13 $\pm$ 0.50)E-3} & \textbf{(6.56 $\pm$ 0.63)E-3} \\
\hline
\end{tabular}%
}
\end{table}

\begin{figure}[t]
\centering
\includegraphics[width=\textwidth]{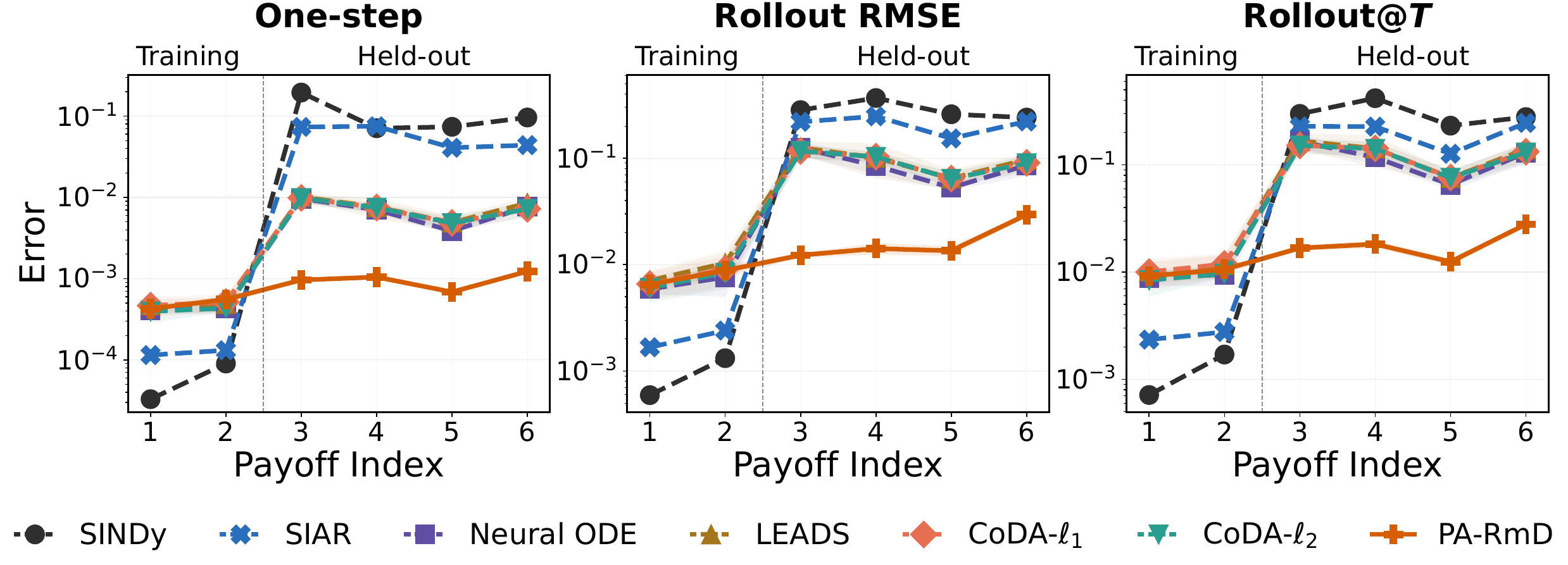}
\caption{Error by payoff function index for the $K=4$ zero-sum cyclic game under replicator dynamics.}
\label{fig:app-payoff-index-cyclic-replicator}
\end{figure}

\begin{figure}[t]
\centering
\includegraphics[width=\textwidth]{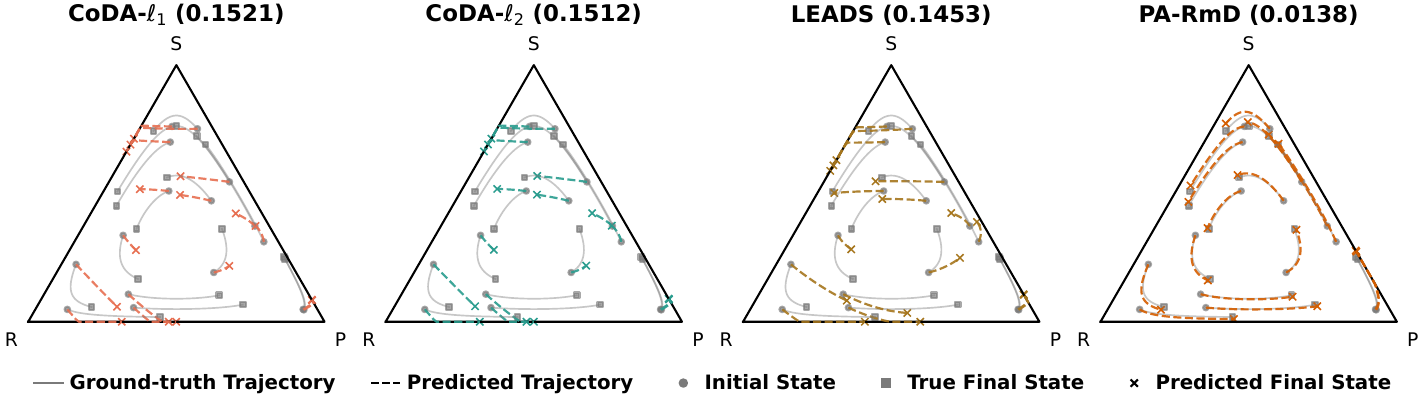}
\caption{Held-out rollout trajectories under replicator dynamics for the four best methods in the zero-sum cyclic game, ordered from fourth-best to best by rollout RMSE. Each panel shows the model name followed by the rollout RMSE for the illustrated run in parentheses. Gray curves represent ground-truth trajectories, and colored dashed curves represent model rollouts.}
\label{fig:synthetic-rollout-cyclic-replicator-top4}
\end{figure}

\paragraph{Evaluation.}
We evaluate out-of-distribution (OOD) prediction under payoff functions held out from training. \cref{app:synthetic-results-details} provides additional in-distribution (ID) results under seen payoff functions and overall results across both seen and held-out payoff functions. All evaluations use test trajectories with independently sampled initial conditions. For one-step evaluation, we predict each next state from the corresponding ground-truth state and report the mean $\ell_2$ error over all test transitions. For rollout evaluation, each model is initialized at the ground-truth initial state and rolled out recursively over the full test horizon. We report the full-trajectory rollout RMSE and final-state $\ell_2$ error (rollout@$\!T$). Rollout metrics are averaged over test trajectories, and all metrics over five independent runs.

\paragraph{Results.}
We first evaluate the performance of different methods on both payoff families, all three response dynamics, and $K\in\{3,4\}$. \Cref{tab:synthetic-replicator} presents the results for replicator dynamics in held-out payoff environments, while \Cref{tab:synthetic-projection,tab:synthetic-mwu} provide the corresponding results for projection and MWU dynamics in \cref{app:synthetic-results-heldout}. Across these settings, \methodPARmD{} consistently achieves the lowest one-step error, rollout RMSE, and rollout@$\!T$, showing that the learned payoff-aware response structure transfers across held-out payoff functions. Additional tables in \cref{app:synthetic-results-overall-seen} report performance in seen training payoff environments and overall across both seen and held-out payoff environments.

We further evaluate every trained model on trajectories generated by each payoff function in the class, including both training and held-out payoff functions. \Cref{fig:app-payoff-index-cyclic-replicator,fig:app-payoff-index-congestion-replicator,fig:app-payoff-index-cyclic-projection,fig:app-payoff-index-congestion-projection,fig:app-payoff-index-cyclic-mwu,fig:app-payoff-index-congestion-mwu} report the error for each payoff function index. Some baselines fit trajectories from training payoff functions well, but their errors increase sharply on trajectories from held-out payoff functions. In contrast, \methodPARmD{} maintains low error across the payoff function class, demonstrating stronger payoff-aware generalization.

Finally, \Cref{fig:synthetic-rollout-cyclic-replicator-top4} visualizes held-out rollouts for a single run under replicator dynamics for the $K=3$ zero-sum cyclic game. Their corresponding results over all five runs are summarized in \cref{tab:synthetic-replicator}. \methodPARmD{} closely tracks the ground-truth trajectories over the full horizon, while the other methods drift away over time. Full rollout comparisons for all methods and synthetic settings are provided in \Cref{fig:app-cyclic-rollout-replicator-full,fig:app-congestion-rollout-replicator-full,fig:app-cyclic-rollout-projection-full,fig:app-congestion-rollout-projection-full,fig:app-cyclic-rollout-mwu-full,fig:app-congestion-rollout-mwu-full}.

\subsection{Real-world Experiments}
\label{sec:exp-real}

\paragraph{Data and split.}
We use a multi-country YouTube Trending panel\footnote{\url{https://www.kaggle.com/datasets/datasnaek/youtube-new}} with 10 countries, 122 aligned dates, and $K=15$ shared content categories. Each country provides one trajectory, where the population state is the distribution of trending videos across content categories. We train models on trajectories from one subset of countries and test them only on trajectories from countries not used for training: for each $q\in\{3,\ldots,9\}$, we train on $q$ countries and evaluate on the remaining countries.
For each value of $q$, we randomly sample 10 training-country sets from all $\binom{10}{q}$ possible choices. In each run, we first average errors across the held-out countries within each training-country set and then across the 10 sampled sets. We report the mean and standard deviation across five independent runs.

\begin{figure}[t]
\centering
\includegraphics[width=\textwidth]{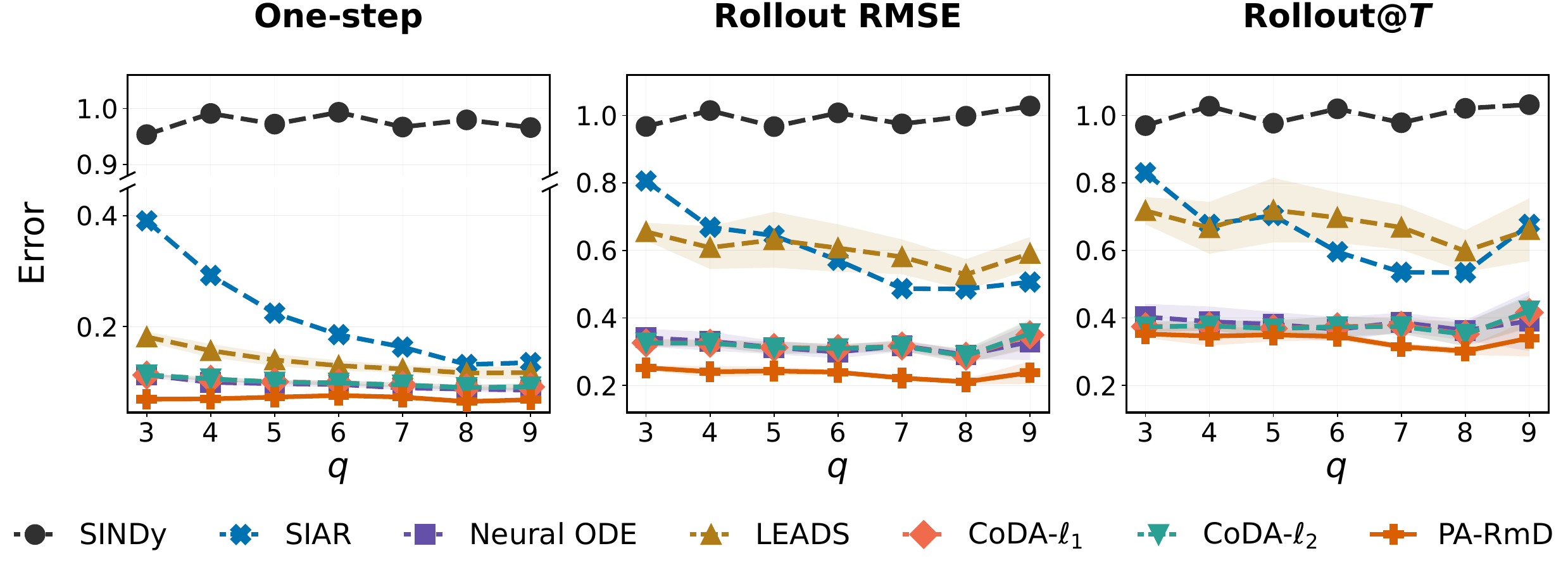}
\caption{YouTube held-out country prediction errors as the number $q$ of training countries varies. One-step errors use observed engagement-derived payoff vectors, while rollout metrics use estimated country-specific payoff proxies.}
\label{fig:youtube-k-curves}
\end{figure}

\paragraph{Evaluation.}
The dataset provides neither explicit payoff values nor a payoff function that can be queried at arbitrary states. We therefore construct an observed payoff vector from user engagement for each country and date \citep{yao2023bad, kayal2025large, mcgrady2025one}. These vectors are used directly for one-step payoff-aware prediction: given the observed state and payoff on a test day, each model predicts the next-day state. Closed-loop rollout additionally requires payoffs at model-predicted states, so for each country we fit a payoff proxy using the first $S$ observed state--payoff pairs and use it to provide payoffs during rollout. Details of the payoff construction and rollout protocol are provided in \Cref{app:real-world-detail}.

\paragraph{Results.}
\Cref{fig:youtube-k-curves} summarizes one-step and closed-loop rollout errors on held-out countries as the number $q$ of training countries varies. The corresponding numerical results are reported in \Cref{tab:exp-youtube} in \Cref{app:youtube-complete}. Across all values of $q$, \methodPARmD{} consistently achieves the lowest one-step error, rollout RMSE, and rollout@$T$ error. These consistent gains demonstrate the effectiveness of \methodPARmD{} for predicting real-world population dynamics.

\section{Conclusions}

We studied payoff-aware prediction of population game dynamics by keeping payoff signals explicit and learning a reusable payoff-independent response structure from state--payoff--transition data. \methodPARmD{} implements this idea through a Riemannian dynamics model that separates payoff-dependent gain from payoff-independent movement cost. We showed that the learned structure preserves fixed-payoff Nash equilibria and derived finite-time one-step prediction guarantees under shared-response and payoff-coverage conditions. Empirically, \methodPARmD{} performs consistently well on synthetic held-out payoff function tasks across multiple game classes and response dynamics, and improves held-out-country one-step prediction on YouTube Trending data with engagement payoff proxies. 
\section*{Acknowledgments}
Renzhe Xu's research was supported by the National Natural Science Foundation of China (No. 72442024), the Shanghai Sailing Program (No. 24YF2711600), and the CCF-Huawei Populus Grove Fund. Xiao Wang's research was supported by the National Key R\&D Program of China (No. 2023YFA1009500), the Fundamental Research Funds for the Central Universities, the Shanghai Sci-Tech Co-Research Program (No. 25HB2703800), and the National Natural Science Foundation of China (No. 62572291). Bo Li's research was supported by the National Natural Science Foundation of China (No. 72171131).

\bibliographystyle{plainnat}
\bibliography{reference}

\newpage
\appendix

\section{Microfoundations of Riemannian Game Dynamics} \label{sect:microfoundations}
This section summarizes the revision-protocol interpretation of Riemannian game dynamics. The material is included to make the population-level model in the main text self-contained; the constructions below follow the standard microfoundation of evolutionary dynamics by revision protocols \citep{sandholm2010population} and the revision-protocol results for Riemannian game dynamics in \citet{mertikopoulos2018riemannian}.

\paragraph{Revision protocols.}
A revision protocol specifies how individual agents switch strategies. To distinguish time from strategy components, write $x_{t,i}$ for the $i$-th coordinate of the population state $x_t$. For $i\neq j$, let $q_{ij}(x_t,v)\ge 0$ denote the instantaneous rate at which an agent currently choosing strategy $i$ switches to strategy $j$, when the population state is $x_t$ and the payoff vector is $v(x_t)$. The induced population dynamic is the mass-balance equation
\[
    \dot x_{t,i}
    =
    \sum_{j\neq i}x_{t,j} q_{ji}(x_t,v)
    -
    x_{t,i}\sum_{j\neq i}q_{ij}(x_t,v),
    \qquad i\in[K].
\]
The first term is the inflow into strategy $i$ from agents currently using other strategies, and the second term is the outflow from strategy $i$ to other strategies. Thus, a population dynamic can be derived from individual revision rates.

\paragraph{Riemannian metrics as revision protocols.}
Proposition 5.1 of \citet{mertikopoulos2018riemannian} shows that, on the interior of the simplex, Riemannian game dynamics admit such a revision-protocol interpretation. In our notation, let
\[
    A_t=G(x_t)^{-1}.
\]
If $A_t$ has nonnegative entries, then, up to a positive state-dependent change of speed, the Riemannian dynamic
\[
    \dot x_t
    =
    A_t v(x_t)
    -
    \frac{\bfone^\top A_t v(x_t)}
    {\bfone^\top A_t\bfone}
    A_t\bfone
\]
can be generated by revision rates determined by $A_t$. For example, after a payoff shift if needed so that $A_t v(x_t)\ge 0$, one such protocol is
\[
    q_{ij}(x_t,v)
    =
    \frac{(A_t\bfone)_i}{x_{t,i}}
    \big(A_t v(x_t)\big)_j,
    \qquad i\neq j.
\]
Substituting these rates into the mass-balance equation gives
\[
    \dot x_t
    =
    \big(\bfone^\top A_t\bfone\big)A_t v(x_t)
    -
    \big(\bfone^\top A_t v(x_t)\big)A_t\bfone,
\]
which is the Riemannian dynamic above after rescaling time by the positive factor $\bfone^\top A_t\bfone$. Hence the metric $G$, or equivalently its inverse $A_t$, can be interpreted as encoding payoff-independent behavioral structure that shapes how agents revise their strategies.

When $A_t$ is diagonal, the same result also yields a pairwise-comparison revision protocol:
\[
    q_{ij}(x_t,v)
    =
    \frac{(A_t)_{ii}(A_t)_{jj}}{x_{t,i}}
    [v_j(x_t)-v_i(x_t)]_+,
    \qquad i\neq j,
\]
where $[a]_+=\max\{a,0\}$. This form is especially useful for interpreting classical dynamics.

\begin{example}[Replicator dynamics~\citep{taylor1978evolutionary}]
The replicator dynamics correspond to the Shahshahani metric~\citep{shahshahani1979new}
\[
    G(x_t)=\operatorname{diag}\left(\frac{1}{x_{t,1}},\ldots,\frac{1}{x_{t,K}}\right),
    \qquad
    A_t=\operatorname{diag}(x_{t,1},\ldots,x_{t,K}).
\]
The diagonal pairwise-comparison protocol becomes
\[
    q_{ij}(x_t,v)
    =
    x_{t,j}[v_j(x_t)-v_i(x_t)]_+,
    \qquad i\neq j.
\]
Substituting into the mass-balance equation gives
\[
    \dot x_{t,i}
    =
    x_{t,i}\left(v_i(x_t)-\sum_{j=1}^K x_{t,j}v_j(x_t)\right),
\]
so agents tend to switch toward strategies with higher current payoffs, and the growth rate of each strategy is its payoff advantage over the population average.
\end{example}

\begin{example}[Projection dynamics~\citep{nagurney1997projected}]
The Euclidean projection dynamics correspond to the Euclidean metric
\[
    G(x_t)=I,
    \qquad
    A_t=I.
\]
The diagonal pairwise-comparison protocol becomes, after a normalization by $K$,
\[
    q_{ij}(x_t,v)
    =
    \frac{1}{Kx_{t,i}}[v_j(x_t)-v_i(x_t)]_+,
    \qquad i\neq j.
\]
Substituting into the mass-balance equation gives
\[
    \dot x_{t,i}
    =
    v_i(x_t)-\frac{1}{K}\sum_{j=1}^K v_j(x_t),
\]
which is the projection of the payoff vector onto the tangent space $\calT$. Thus projection dynamics provide the Euclidean counterpart to replicator dynamics within the same revision-protocol view.
\end{example}

\section{Omitted Proofs} \label{sec:proofs}

\subsection{Proof of \Cref{prop:optimization}}
\begin{proof}
The equivalence is pointwise in $x$, so fix a state $x\in\calX$ and write $G=G_\theta(x)$, $a=u(x)$. Every feasible direction $z\in\calT$ can be written uniquely as $z=Uy$, where $y\in\bbR^{K-1}$. Since $U$ has orthonormal columns, the objective in \cref{eq:argmax} becomes
\[
    a^\top Uy-\frac{1}{2}y^\top U^\top G U y.
\]
The matrix $U^\top G U$ is positive definite, so this strictly concave quadratic has the unique maximizer
\[
    y^\star=(U^\top G U)^{-1}U^\top a.
\]
Substituting $z^\star=Uy^\star$ gives
\[
    z^\star
    =
    U(U^\top G U)^{-1}U^\top a.
\]
Thus the Riemannian form gives $C_\theta(x)=(U^\top G_\theta(x)U)^{-1}$. Conversely, fix any $C_\theta(x)\in\bbS^{K-1}_{++}$. Let $P=\bfone\bfone^\top/K$, choose any scalar $c>0$, and define
\[
    G_\theta(x):=UC_\theta(x)^{-1}U^\top+cP.
\]
The matrix $G_\theta(x)$ is symmetric positive definite because it is positive definite on $\calT$ through $UC_\theta(x)^{-1}U^\top$ and positive definite on $\operatorname{span}\{\bfone\}$ through $cP$. Moreover, $U^\top G_\theta(x)U=C_\theta(x)^{-1}$, so
\[
    U(U^\top G_\theta(x)U)^{-1}U^\top a
    =
    UC_\theta(x)U^\top a
\]
for every payoff vector $a\in\bbR^K$. Since the argument holds at each $x$, the two parametrizations define the same class of vector fields.
\end{proof}

\subsection{Proof of \Cref{prop:nash-equilibrium-preservation}}
\begin{proof}
If $U^\top u(x)=0$, then $F_{\widetilde C}(x,u(x))=0$. Conversely, if
$F_{\widetilde C}(x,u(x))=0$, then
\[
    U\widetilde C(x)U^\top u(x)=0.
\]
Because $U$ has full column rank, $\widetilde C(x)U^\top u(x)=0$. Since
$\widetilde C(x)$ is positive definite, it is invertible, so
$U^\top u(x)=0$.

It remains to identify this condition with Nash equilibrium on $\mathcal X$.
Because the columns of $U$ span $\calT=\{z:\bfone^\top z=0\}$, the condition
$U^\top u(x)=0$ is equivalent to $u(x)\in\calT^\perp=\operatorname{span}\{\bfone\}$.
Hence all pure strategies have the same payoff at $x$. Since
$\mathcal X\subset\Delta_K^\circ$, this is exactly the Nash equilibrium
condition on $\mathcal X$, so
\[
    \mathcal S(F_{\widetilde C},u)
    =
    \mathcal N(u)
    =
    \{x\in\mathcal X:U^\top u(x)=0\}.
\]
\end{proof}

\subsection{Proof of \Cref{thrm:finite-time-target-transfer-directional}}
\begin{proof}
Fix a utility $u$ and a state $x\in\mathcal X$. Since $U$ has
orthonormal columns, we have
\[
    U^\top U=I_{K-1},
    \qquad
    \|U\|_{\mathrm{op}}=\|U^\top\|_{\mathrm{op}}=1.
\]
Therefore,
\[
\begin{aligned}
    \|f_\star^u(x)\|_2
    &=
    \|UC_\star(x)U^\top u(x)\|_2        \\
    &\le
    \|U\|_{\mathrm{op}}
    \|C_\star(x)\|_{\mathrm{op}}
    \|U^\top\|_{\mathrm{op}}
    \|u(x)\|_2                          \\
    &\le
    M\Lambda .
\end{aligned}
\]
Similarly, for any $x,y\in\mathcal X$,
\[
\begin{aligned}
    &
    \|f_\star^u(x)-f_\star^u(y)\|_2                                      \\
    &=
    \left\|
    UC_\star(x)U^\top u(x)-UC_\star(y)U^\top u(y)
    \right\|_2                                                           \\
    &=
    \left\|
    U\left[
    C_\star(x)U^\top u(x)-C_\star(y)U^\top u(y)
    \right]
    \right\|_2                                                           \\
    &\le
    \left\|
    C_\star(x)U^\top u(x)-C_\star(y)U^\top u(y)
    \right\|_2                                                           \\
    &\le
    \left\|
    C_\star(x)U^\top\big(u(x)-u(y)\big)
    \right\|_2
    +
    \left\|
    \big(C_\star(x)-C_\star(y)\big)U^\top u(y)
    \right\|_2                                                           \\
    &\le
    M L_u\|x-y\|_2
    +
    L_C\Lambda\|x-y\|_2                                                   \\
    &=
    \left(
    ML_u+L_C\Lambda
    \right)
    \|x-y\|_2 .
\end{aligned}
\]
Thus the vector field $f_\star^u$ is uniformly bounded by $M\Lambda$ and
uniformly Lipschitz with constant $ML_u+L_C\Lambda$.

Define
\[
    B_\star
    :=
    \frac12
    M\Lambda
    \left(
    ML_u+L_C\Lambda
    \right).
\]
Let $z(s)$ be the solution of
\[
    \dot z(s)=f_\star^u(z(s)),
    \qquad
    z(0)=x .
\]
By definition of the flow map,
\[
    \Psi_\delta^u(x)
    =
    z(\delta)
    =
    x+\int_0^\delta f_\star^u(z(s))\,ds .
\]
Therefore,
\[
\begin{aligned}
    &
    \left\|
    \frac{1}{\delta}
    \left(
    \Psi_\delta^u(x)-x
    \right)
    -
    f_\star^u(x)
    \right\|_2                                                     \\
    &=
    \left\|
    \frac{1}{\delta}
    \int_0^\delta
    \left(
    f_\star^u(z(s))-f_\star^u(x)
    \right)
    ds
    \right\|_2                                                     \\
    &\le
    \frac{1}{\delta}
    \int_0^\delta
    \left\|
    f_\star^u(z(s))-f_\star^u(x)
    \right\|_2
    ds                                                            \\
    &\le
    \frac{ML_u+L_C\Lambda}{\delta}
    \int_0^\delta
    \|z(s)-x\|_2
    ds .
\end{aligned}
\]
Since all flow segments remain in $\mathcal X$,
\[
    \|f_\star^u(z(r))\|_2\le M\Lambda
    \qquad
    \text{for all }r\in[0,\delta].
\]
Hence
\[
\begin{aligned}
    \|z(s)-x\|_2
    &=
    \left\|
    \int_0^s f_\star^u(z(r))\,dr
    \right\|_2                                      \\
    &\le
    \int_0^s
    \|f_\star^u(z(r))\|_2\,dr                       \\
    &\le
    M\Lambda s .
\end{aligned}
\]
Substituting this into the previous display gives
\[
\begin{aligned}
    &
    \left\|
    \frac{1}{\delta}
    \left(
    \Psi_\delta^u(x)-x
    \right)
    -
    f_\star^u(x)
    \right\|_2                                      \\
    &\le
    \frac{ML_u+L_C\Lambda}{\delta}
    \int_0^\delta M\Lambda s\,ds                   \\
    &=
    \frac12
    M\Lambda
    \left(
    ML_u+L_C\Lambda
    \right)
    \delta                                         \\
    &=
    B_\star\delta .
\end{aligned}
\]

Now fix $x\in\mathcal X$. Because the vector field is tangent to the
simplex, every displacement
\[
    \Psi_\delta^u(x)-x
\]
lies in the tangent space
\[
    T=\{z\in\mathbb R^K:\mathbf 1^\top z=0\}.
\]
Also, $UCU^\top u(x)\in T$ for every $C$, since the columns of $U$
span $T$. For any $z\in T$,
\[
    z=UU^\top z,
    \qquad
    \|z\|_2=\|U^\top z\|_2.
\]
Therefore the $K$-dimensional objective defining
$C_{\delta,\mathrm{tr}}^\star(x)$ is equivalent to the tangent-coordinate
objective
\[
    C_{\delta,\mathrm{tr}}^\star(x)
    \in
    \arg\min_{C\in\mathbb S_+^{K-1}}
    \mathbb E_{v\sim\mathcal D_{\mathrm{tr}}(\cdot\mid x)}
    \left[
    \left\|
    \frac{1}{\delta}
    U^\top
    \left(
    \Psi_\delta^v(x)-x
    \right)
    -
    CU^\top v(x)
    \right\|_2^2
    \right].
\]

Define the finite-time bias
\[
    b_\delta^u(x)
    :=
    \frac{1}{\delta}
    U^\top
    \left(
    \Psi_\delta^u(x)-x
    \right)
    -
    C_\star(x)U^\top u(x).
\]
Since
\[
    U^\top f_\star^u(x)
    =
    U^\top U C_\star(x)U^\top u(x)
    =
    C_\star(x)U^\top u(x),
\]
we can write
\[
    b_\delta^u(x)
    =
    U^\top
    \left[
    \frac{1}{\delta}
    \left(
    \Psi_\delta^u(x)-x
    \right)
    -
    f_\star^u(x)
    \right].
\]
Thus, using $\|U^\top\|_{\mathrm{op}}=1$,
\[
    \|b_\delta^u(x)\|_2
    \le
    B_\star\delta .
\]

Let
\[
    \widehat C(x)=C_{\delta,\mathrm{tr}}^\star(x),
    \qquad
    E(x)=\widehat C(x)-C_\star(x).
\]
Because
\[
    \frac{1}{\delta}
    U^\top
    \left(
    \Psi_\delta^v(x)-x
    \right)
    =
    C_\star(x)U^\top v(x)+b_\delta^v(x),
\]
the tangent-coordinate objective can be rewritten as
\[
    \widehat C(x)
    \in
    \arg\min_{C\in\mathbb S_+^{K-1}}
    \mathbb E_{v\sim\mathcal D_{\mathrm{tr}}(\cdot\mid x)}
    \left[
    \left\|
    b_\delta^v(x)
    -
    \left(C-C_\star(x)\right)U^\top v(x)
    \right\|_2^2
    \right].
\]

Since $C_\star(x)\in\mathbb S_+^{K-1}$, it is feasible. The feasible set
$\mathbb S_+^{K-1}$ is convex, so the line segment
\[
    C_s(x)
    =
    (1-s)\widehat C(x)+sC_\star(x),
    \qquad
    s\in[0,1],
\]
is feasible. By optimality of $\widehat C(x)$, the function
\[
    \varphi(s)
    =
    \mathbb E_{v\sim\mathcal D_{\mathrm{tr}}(\cdot\mid x)}
    \left[
    \left\|
    b_\delta^v(x)
    -
    \left(C_s(x)-C_\star(x)\right)U^\top v(x)
    \right\|_2^2
    \right]
\]
has a minimum at $s=0$. Hence
\[
    \varphi'(0^+)\ge 0.
\]
But
\[
    C_s(x)-C_\star(x)
    =
    (1-s)E(x).
\]
Therefore
\[
    \varphi(s)
    =
    \mathbb E
    \left[
    \left\|
    b_\delta^v(x)
    -
    (1-s)E(x)U^\top v(x)
    \right\|_2^2
    \right],
\]
where the expectation is over $v\sim\mathcal D_{\mathrm{tr}}(\cdot\mid x)$.
Differentiating at $s=0$ gives
\[
\begin{aligned}
    \varphi'(0^+)
    &=
    2
    \mathbb E
    \left[
    \left\langle
    b_\delta^v(x)-E(x)U^\top v(x),
    E(x)U^\top v(x)
    \right\rangle
    \right].
\end{aligned}
\]
Since $\varphi'(0^+)\ge0$, we obtain
\[
    \mathbb E
    \left[
    \|E(x)U^\top v(x)\|_2^2
    \right]
    \le
    \mathbb E
    \left[
    \left\langle
    b_\delta^v(x),
    E(x)U^\top v(x)
    \right\rangle
    \right].
\]
By Cauchy--Schwarz,
\[
\begin{aligned}
    &
    \mathbb E
    \left[
    \left\langle
    b_\delta^v(x),
    E(x)U^\top v(x)
    \right\rangle
    \right]                                      \\
    &\le
    \left(
    \mathbb E
    \left[
    \|b_\delta^v(x)\|_2^2
    \right]
    \right)^{1/2}
    \left(
    \mathbb E
    \left[
    \|E(x)U^\top v(x)\|_2^2
    \right]
    \right)^{1/2}.
\end{aligned}
\]
Since $\|b_\delta^v(x)\|_2\le B_\star\delta$ uniformly,
\[
    \left(
    \mathbb E
    \left[
    \|b_\delta^v(x)\|_2^2
    \right]
    \right)^{1/2}
    \le
    B_\star\delta .
\]
Therefore,
\[
    \left(
    \mathbb E
    \left[
    \|E(x)U^\top v(x)\|_2^2
    \right]
    \right)^{1/2}
    \le
    B_\star\delta .
\]

Now use the directional coverage assumption. Since
\[
    U^\top w(x)
    =
    \mathbb E
    \left[
    \rho_x(v)U^\top v(x)
    \right],
\]
we have
\[
\begin{aligned}
    \|E(x)U^\top w(x)\|_2
    &=
    \left\|
    \mathbb E
    \left[
    \rho_x(v)E(x)U^\top v(x)
    \right]
    \right\|_2                                      \\
    &\le
    \mathbb E
    \left[
    |\rho_x(v)|\,
    \|E(x)U^\top v(x)\|_2
    \right]                                        \\
    &\le
    \left(
    \mathbb E[\rho_x(v)^2]
    \right)^{1/2}
    \left(
    \mathbb E
    \left[
    \|E(x)U^\top v(x)\|_2^2
    \right]
    \right)^{1/2}                                  \\
    &\le
    \gamma_w B_\star\delta .
\end{aligned}
\]

Finally, for the test utility $w$,
\[
\begin{aligned}
    &
    \Psi_\delta^w(x)-x
    -
    \delta U\widehat C(x)U^\top w(x)                 \\
    &=
    \delta U
    \left[
    b_\delta^w(x)
    -
    E(x)U^\top w(x)
    \right].
\end{aligned}
\]
Hence
\[
\begin{aligned}
    &
    \left\|
    \Psi_\delta^w(x)-x
    -
    \delta U\widehat C(x)U^\top w(x)
    \right\|_2                                      \\
    &\le
    \delta
    \left(
    \|b_\delta^w(x)\|_2
    +
    \|E(x)U^\top w(x)\|_2
    \right)                                        \\
    &\le
    \delta
    \left(
    B_\star\delta
    +
    \gamma_w B_\star\delta
    \right)                                        \\
    &=
    B_\star(1+\gamma_w)\delta^2 .
\end{aligned}
\]
Taking the supremum over $x\in\mathcal X$ proves the theorem.
\end{proof}

\subsection{Proof of \Cref{coro:finite-time-target-transfer}}
\begin{proof}
It suffices to verify the directional coverage condition required by
\cref{thrm:finite-time-target-transfer-directional}. Fix any
$x\in\mathcal X$. Since
\[
    \Sigma_{\mathrm{tr}}(x)
    \succeq
    \mu I_{K-1},
\]
the matrix $\Sigma_{\mathrm{tr}}(x)$ is invertible. Define
\[
    \rho_x(v)
    =
    v(x)^\top U
    \Sigma_{\mathrm{tr}}(x)^{-1}
    U^\top w(x).
\]
Then
\[
\begin{aligned}
    \mathbb E_{v\sim\mathcal D_{\mathrm{tr}}(\cdot\mid x)}
    \left[
    \rho_x(v)U^\top v(x)
    \right]
    &=
    \mathbb E_{v\sim\mathcal D_{\mathrm{tr}}(\cdot\mid x)}
    \left[
    U^\top v(x)v(x)^\top U
    \right]
    \Sigma_{\mathrm{tr}}(x)^{-1}
    U^\top w(x)                                      \\
    &=
    \Sigma_{\mathrm{tr}}(x)
    \Sigma_{\mathrm{tr}}(x)^{-1}
    U^\top w(x)                                      \\
    &=
    U^\top w(x).
\end{aligned}
\]
Thus $U^\top w(x)$ can be represented by the projected training utilities
with coefficient $\rho_x$.

It remains to bound the size of this coefficient. We have
\[
\begin{aligned}
    \mathbb E_{v\sim\mathcal D_{\mathrm{tr}}(\cdot\mid x)}
    \left[
    \rho_x(v)^2
    \right]
    &=
    w(x)^\top U
    \Sigma_{\mathrm{tr}}(x)^{-1}
    \mathbb E_{v\sim\mathcal D_{\mathrm{tr}}(\cdot\mid x)}
    \left[
    U^\top v(x)v(x)^\top U
    \right]
    \Sigma_{\mathrm{tr}}(x)^{-1}
    U^\top w(x)                                      \\
    &=
    w(x)^\top U
    \Sigma_{\mathrm{tr}}(x)^{-1}
    U^\top w(x).
\end{aligned}
\]
Since
\[
    \Sigma_{\mathrm{tr}}(x)
    \succeq
    \mu I_{K-1},
\]
we have
\[
    \Sigma_{\mathrm{tr}}(x)^{-1}
    \preceq
    \frac{1}{\mu}I_{K-1}.
\]
Therefore,
\[
\begin{aligned}
    \mathbb E_{v\sim\mathcal D_{\mathrm{tr}}(\cdot\mid x)}
    \left[
    \rho_x(v)^2
    \right]
    &\le
    \frac{1}{\mu}
    \left\|
    U^\top w(x)
    \right\|_2^2                                      \\
    &\le
    \frac{1}{\mu}
    \left\|
    w(x)
    \right\|_2^2                                      \\
    &\le
    \frac{\Lambda^2}{\mu},
\end{aligned}
\]
where the second inequality uses $\|U^\top\|_{\mathrm{op}}=1$, and the
last inequality follows from \cref{assum:regularity}.

Hence the directional coverage condition holds with
\[
    \gamma_w
    =
    \frac{\Lambda}{\sqrt{\mu}}.
\]
Applying \cref{thrm:finite-time-target-transfer-directional} gives
\[
\begin{aligned}
    \sup_{x\in\mathcal X}
    \left\|
    \Psi_\delta^w(x)-x
    -
    \delta U C_{\delta,\mathrm{tr}}^\star(x)U^\top w(x)
    \right\|_2
    &\le
    B_\star(1+\gamma_w)\delta^2                                      \\
    &=
    B_\star
    \left(
    1+\frac{\Lambda}{\sqrt{\mu}}
    \right)
    \delta^2 .
\end{aligned}
\]
This proves the corollary.
\end{proof}

\section{Additional Experiments} \label{sec:appendix-exp}

\subsection{Experimental Details}
\label{app:exp-details}
\subsubsection{Baselines and Training Details} \label{app:baselines}
For fairness, all baselines use the same payoff-aware input tuple as \methodPARmD{}, namely the current population state, the payoff observation, and the sampling interval.
\methodSINDy{} is a sparse-regression baseline that fits a parsimonious dynamics model from a predefined library of candidate features \citep{brunton2016discovering}.
\methodNeuralODE{} is a payoff-conditioned neural vector-field model that learns $f_\theta(x,v)$ and integrates it forward in time \citep{chen2018neural}.
\methodLEADS{} is a dynamics-generalization baseline designed to transfer learned dynamics across environments \citep{yin2021leads}.
\methodCoDALOne{} and \methodCoDALTwo{} are context-informed dynamics adaptation baselines with different regularization choices \citep{kirchmeyer2022coda}. \methodSIAR{} is a game-structured dynamics identification baseline that fits polynomial vector fields under side-information constraints \citep{sakos2024data}.
The neural models are optimized using Adam, while \methodSINDy{} and \methodSIAR{} are fit by sparse regression and sum-of-squares optimization, respectively.
Unless otherwise stated, results are averaged over five independent runs.

For compute resources, each method on each experimental setting can be run on a single NVIDIA RTX 4090 GPU with 24GB memory; no multi-GPU or distributed training is required.

\subsubsection{Synthetic Data Details} \label{app:synthetic-detail}

\paragraph{Zero-sum cyclic games.}
For each $K\in\{3,4\}$, let
$A_{\mathrm{base}}^{(K)}\in\bbR^{K\times K}$ denote the
antisymmetric payoff matrix associated with a directed cycle:
\[
    (A_{\mathrm{base}}^{(K)})_{ij}
    =
    \begin{cases}
        -1, & j\equiv i+1 \pmod K,\\
        1,  & i\equiv j+1 \pmod K,\\
        0,  & \text{otherwise}.
    \end{cases}
\]
Thus, strategy $i$ loses to the next strategy on the cycle and
defeats the previous strategy.

To generate distinct cyclic payoff environments without counting
global rotations multiple times, we keep strategy $1$ fixed and
permute the remaining $K-1$ strategy labels. Let
$\pi_1,\ldots,\pi_{E_K}$ denote these permutations, where
\[
    E_K=(K-1)!.
\]
For each environment $e\in[E_K]$, define the permuted payoff matrix and its associated payoff mapping by
\[
    (A_e^{(K)})_{ij}
    =
    (A_{\mathrm{base}}^{(K)})_{\pi_e(i),\pi_e(j)},
    \qquad
    v_e^{(K)}(x)=A_e^{(K)}x,
    \quad x\in\Delta_K.
\]
The resulting cyclic payoff family is
\[
    \calU_K^{\mathrm{cyclic}}
    =
    \{v_e^{(K)}:e\in[E_K]\}.
\]
Since every $A_e^{(K)}$ is antisymmetric, every environment corresponds to a zero-sum game.

\paragraph{Potential congestion games.}
For each $K\in\{3,4\}$, we fix $K$ labeled routes and $K$
resources. All environments share the ring incidence matrix
$B^{(K)}\in\{0,1\}^{K\times K}$, defined by
\[
    B^{(K)}_{ri}
    =
    \mathbf{1}\{i=r\}
    +
    \mathbf{1}\{i\equiv r+1 \pmod K\}.
\]
Thus, route $i$ uses resources $i-1$ and $i$, with indices
interpreted modulo $K$. At population state $x\in\Delta_K$, the
resource-load vector is
\[
    \ell(x)=B^{(K)}x,
    \qquad
    \ell_r(x)
    =
    (B^{(K)}x)_r
    =
    \sum_{i=1}^K B^{(K)}_{ri}x_i.
\]

To match the size of the cyclic payoff family, we sample
$E_K=(K-1)!$ congestion environments. For each environment
$e\in[E_K]$, we sample vectors $a_e,b_e\in\bbR^K$, with all
coordinates drawn independently according to
\[
    a_{e,r}\sim\operatorname{Unif}(0.8,2.2),
    \qquad
    b_{e,r}\sim\operatorname{Unif}(0,0.3),
    \qquad r\in[K].
\]
We define the scaling factor
\[
    s_e
    =
    \left\|
    (B^{(K)})^\top
    \operatorname{diag}(a_e)
    B^{(K)}
    \right\|_F^{-1}.
\]
The resulting payoff mapping is
\[
    v_e^{(K)}(x)
    =
    -s_e(B^{(K)})^\top
    \left[
        \operatorname{diag}(a_e)B^{(K)}x+b_e
    \right],
    \qquad x\in\Delta_K.
\]
Here, resource $r$ has affine cost
$s_e(a_{e,r}\ell_r(x)+b_{e,r})$, and the payoff of a route is the
negative total cost of the resources it uses.

The resulting payoff family is
\[
    \calU_K^{\mathrm{congestion}}
    =
    \{v_e^{(K)}:e\in[E_K]\}.
\]
The factor $s_e$ normalizes the Frobenius norm of the state-dependent linear component to one. Since its linear coefficient matrix $-s_e(B^{(K)})^\top\operatorname{diag}(a_e)B^{(K)}$ is symmetric and negative semidefinite, $v_e^{(K)}$ is the gradient of a concave quadratic potential; hence every environment is an exact potential game.

\paragraph{Training and held-out environments.}
As constructed, both payoff families contain $E_K=(K-1)!$ payoff environments, indexed by $e\in[E_K]$. For $K=3$, we use environment $e=1$ for training
and hold out the other environment. For $K=4$, we use environments
$e=1,2$ for training and hold out the remaining four. We use the
same environment split for all response dynamics.

For each training payoff environment, we generate 16 training trajectories. A separate evaluation pool also contains 16 trajectories per payoff environment, including both training and held-out environments. Each trajectory contains 20 transitions.

\paragraph{Response dynamics.}
For each payoff function $v$ in either payoff class, trajectories are generated using one of three response dynamics.
Replicator dynamics evolve as
\[
    \dot x_i
    =
    x_i\left(v_i(x)-x^\top v(x)\right),
    \qquad i\in[K].
\]
Projection dynamics use the Euclidean projection of the payoff vector onto the simplex tangent space:
\[
    \dot x
    =
    \Pi_{\calT}v(x),
    \qquad
    \Pi_{\calT}
    =
    I-\frac{1}{K}\bfone\bfone^\top .
\]
Multiplicative weights update (MWU) is the discrete-time update
\[
    x_i^{(m+1)}
    =
    \frac{x_i^{(m)}\exp(\eta v_i(x^{(m)}))}
    {\sum_{j=1}^K x_j^{(m)}\exp(\eta v_j(x^{(m)}))},
    \qquad i\in[K],
\]
with a fixed step size $\eta$ across all MWU experiments.

\paragraph{Synthetic transition interval.}
All synthetic transitions used for training and testing use $\delta=0.1$.
For continuous-time dynamics, trajectories are numerically integrated using RK4 \citep{hairer1993solving} and observed every $\delta=0.1$ time units. For MWU, each observed transition consists of two consecutive discrete iterates. To retain the common transition-tuple format, we encode one MWU update with $\delta=0.1$.

\subsubsection{Real-world Data Details} \label{app:real-world-detail}

\paragraph{YouTube state and observed payoff.}
The YouTube experiment uses a multi-country panel of daily trending videos.
For country $c$, day $t$, and category $i$, let $\mathcal V_{c,t,i}$ denote the set of trending videos in that category.
We define the population state as the empirical distribution of trending videos across the $K$ categories:
\[
    x^{(c)}_{t,i}
    =
    \frac{|\mathcal V_{c,t,i}|}
    {\sum_{j=1}^K |\mathcal V_{c,t,j}|},
    \qquad
    x^{(c)}_t\in\Delta_K.
\]

Because the dataset provides no explicit payoffs, we construct an engagement-derived payoff vector from video view counts.
For each video observation $r$, we define the log-transformed view score \citep{mcgrady2025one}:
\[
    \phi(r)
    =
    \log\!\bigl(1+\mathrm{views}(r)\bigr).
\]
For each nonempty country--day--category cell, we aggregate these scores using the median:
\[
    m^{(c)}_{t,i}
    =
    \operatorname*{median}_{r\in\mathcal V_{c,t,i}}
    \phi(r).
\]
Within each country and category, we forward-fill missing scores over time and backfill any leading missing values using the first available score, yielding $\tilde m^{(c)}_{t,i}$.
Finally, for each country and day, we define the observed payoff vector by centering the filled scores across categories:
\[
    v^{(c)}_{t,i}
    =
    \tilde m^{(c)}_{t,i}
    -
    \frac{1}{K}\sum_{j=1}^K \tilde m^{(c)}_{t,j}.
\]
By construction, $\sum_{i=1}^K v^{(c)}_{t,i}=0$.

\paragraph{Country-specific payoff proxy and rollout protocol.}
Closed-loop prediction requires payoffs at model-predicted states, but the engagement-derived payoffs above are observed only along realized trajectories.
For each held-out country $c$, we use the first $S$ observed state--payoff pairs, $\{(x_s^{(c)},v_s^{(c)})\}_{s=0}^{S-1}$, as a support prefix. From this prefix, we estimate a country-specific payoff proxy using Nadaraya--Watson kernel regression with a Gaussian kernel, a standard nonparametric method that predicts the payoff at a query state by averaging payoff observations at nearby states \citep{tosatto2020nonparametric,loh2026practical}.
For a query state $x$, the normalized kernel weights and the resulting payoff estimate are
\[
    a_s^{(c)}(x)
    =
    \frac{\exp\!\left(-\|x-x_s^{(c)}\|_2^2/(2h_c^2)\right)}
    {\sum_{r=0}^{S-1}\exp\!\left(-\|x-x_r^{(c)}\|_2^2/(2h_c^2)\right)},
    \qquad
    \bar v_c(x)
    =
    \sum_{s=0}^{S-1}a_s^{(c)}(x)v_s^{(c)}.
\]
We choose the bandwidth using the median-distance heuristic, i.e., $h_c = \operatorname{median}_{0\leq r<s<S} \|x_r^{(c)}-x_s^{(c)}\|_2$.
To match the centering convention of the observed payoffs, we define the final country-specific proxy as
\[
    \hat v_c(x)
    =
    \bar v_c(x)
    -
    \frac{\mathbf{1}^{\top}\bar v_c(x)}{K}\mathbf{1},
\]
which ensures $\mathbf{1}^{\top}\hat v_c(x)=0$.

Starting from the first state after the support prefix, we initialize $\hat x_S^{(c)}=x_S^{(c)}$ and roll out each fitted model over the remaining steps. At each step $t=S,\ldots,T-1$, the model evaluates the country-specific proxy at its current predicted state $\hat x_{t}^{(c)}$ and uses $\hat v_c(\hat x_t^{(c)})$ to predict the next-day state $\hat x_{t+1}^{(c)}$. The reported rollout results use $S=40$. With 122 aligned dates, this leaves 81 transitions for evaluation per country, and the support prefix is excluded from the rollout metrics.

The payoff proxy is used only for closed-loop rollout. One-step errors are evaluated separately over the full held-out trajectories using the observed states and payoff vectors and are therefore independent of $\hat v_c$.

\subsection{Detailed Synthetic Results}
\label{app:synthetic-results-details}

\subsubsection{Additional Held-out Results}
\label{app:synthetic-results-heldout}

\begin{table}[H]
\renewcommand{\underline}[1]{\begingroup\setbox0=\hbox{#1}\leavevmode\vtop{\offinterlineskip\copy0\kern0.25pt\hrule height0.3pt width\wd0}\endgroup}
\centering
\caption{Synthetic held-out results under projection dynamics. Each entry is mean $\pm$ std over five runs on held-out payoff functions.}
\label{tab:synthetic-projection}
\scriptsize
\renewcommand{\arraystretch}{1.12}
\setlength{\tabcolsep}{3pt}
\resizebox{\linewidth}{!}{%
\begin{tabular}{c|c|ccc|ccc}
\hline
\multirow{2}{*}{Game} & \multirow{2}{*}{Method} & \multicolumn{3}{c|}{$K=3$} & \multicolumn{3}{c}{$K=4$} \\
\cline{3-8}
& & One-step ($\downarrow$) & Rollout RMSE ($\downarrow$) & Rollout@$T$ ($\downarrow$) & One-step ($\downarrow$) & Rollout RMSE ($\downarrow$) & Rollout@$T$ ($\downarrow$) \\
\hline
\multirow{7}{*}{\begin{tabular}[c]{@{}c@{}}Zero-sum\\cyclic\end{tabular}} & \methodSINDy{} & (3.85 $\pm$ 0.00)E-1 & (6.83 $\pm$ 0.00)E-1 & (7.83 $\pm$ 0.00)E-1 & (8.38 $\pm$ 0.00)E-2 & (4.65 $\pm$ 0.00)E-1 & (6.04 $\pm$ 0.00)E-1 \\
 & \methodSIAR{} & (7.68 $\pm$ 0.00)E-1 & (6.11 $\pm$ 0.00)E-1 & (7.16 $\pm$ 0.00)E-1 & (1.50 $\pm$ 0.00)E-1 & (2.77 $\pm$ 0.00)E-1 & (2.79 $\pm$ 0.00)E-1 \\
 & \methodNeuralODE{} & (3.19 $\pm$ 0.41)E-2 & \underline{(3.55 $\pm$ 0.71)E-1} & \underline{(5.58 $\pm$ 1.12)E-1} & (1.36 $\pm$ 0.20)E-2 & (1.83 $\pm$ 0.29)E-1 & (2.55 $\pm$ 0.43)E-1 \\
 & \methodLEADS{} & (3.53 $\pm$ 0.30)E-2 & (4.41 $\pm$ 0.74)E-1 & (6.88 $\pm$ 1.08)E-1 & \underline{(1.24 $\pm$ 0.12)E-2} & \underline{(1.60 $\pm$ 0.21)E-1} & \underline{(2.24 $\pm$ 0.31)E-1} \\
 & \methodCoDALOne{} & (3.12 $\pm$ 0.38)E-2 & (3.66 $\pm$ 0.41)E-1 & (5.88 $\pm$ 0.56)E-1 & (1.30 $\pm$ 0.09)E-2 & (1.70 $\pm$ 0.14)E-1 & (2.37 $\pm$ 0.20)E-1 \\
 & \methodCoDALTwo{} & \underline{(3.12 $\pm$ 0.38)E-2} & (3.66 $\pm$ 0.41)E-1 & (5.87 $\pm$ 0.56)E-1 & (1.32 $\pm$ 0.12)E-2 & (1.75 $\pm$ 0.17)E-1 & (2.48 $\pm$ 0.25)E-1 \\
\rowcolor{gray!15}
\cellcolor{white} & \methodPARmD{} & \textbf{(9.61 $\pm$ 0.21)E-3} & \textbf{(1.37 $\pm$ 0.06)E-1} & \textbf{(2.08 $\pm$ 0.09)E-1} & \textbf{(7.01 $\pm$ 0.04)E-3} & \textbf{(1.08 $\pm$ 0.06)E-1} & \textbf{(1.47 $\pm$ 0.06)E-1} \\
\hline
\multirow{7}{*}{\begin{tabular}[c]{@{}c@{}}Potential\\congestion\end{tabular}} & \methodSINDy{} & (4.47 $\pm$ 0.00)E-3 & (5.38 $\pm$ 0.00)E-2 & (7.93 $\pm$ 0.00)E-2 & (8.70 $\pm$ 0.00)E-3 & (8.02 $\pm$ 0.00)E-2 & (1.12 $\pm$ 0.00)E-1 \\
 & \methodSIAR{} & (4.77 $\pm$ 0.00)E-3 & (4.60 $\pm$ 0.00)E-2 & (7.44 $\pm$ 0.00)E-2 & (2.27 $\pm$ 0.00)E-2 & (1.54 $\pm$ 0.00)E-1 & (2.01 $\pm$ 0.00)E-1 \\
 & \methodNeuralODE{} & (3.44 $\pm$ 0.31)E-3 & (3.47 $\pm$ 0.32)E-2 & (5.42 $\pm$ 0.50)E-2 & \underline{(4.37 $\pm$ 0.90)E-3} & \underline{(4.18 $\pm$ 0.86)E-2} & \underline{(6.33 $\pm$ 1.36)E-2} \\
 & \methodLEADS{} & (3.40 $\pm$ 0.17)E-3 & (3.43 $\pm$ 0.17)E-2 & (5.36 $\pm$ 0.27)E-2 & (5.23 $\pm$ 0.40)E-3 & (4.86 $\pm$ 0.36)E-2 & (7.48 $\pm$ 0.58)E-2 \\
 & \methodCoDALOne{} & \underline{(3.22 $\pm$ 0.38)E-3} & \underline{(3.24 $\pm$ 0.40)E-2} & \underline{(5.06 $\pm$ 0.62)E-2} & (4.44 $\pm$ 0.74)E-3 & (4.20 $\pm$ 0.74)E-2 & (6.41 $\pm$ 1.16)E-2 \\
 & \methodCoDALTwo{} & (3.25 $\pm$ 0.34)E-3 & (3.27 $\pm$ 0.35)E-2 & (5.11 $\pm$ 0.55)E-2 & (4.72 $\pm$ 0.82)E-3 & (4.44 $\pm$ 0.82)E-2 & (6.81 $\pm$ 1.28)E-2 \\
\rowcolor{gray!15}
\cellcolor{white} & \methodPARmD{} & \textbf{(1.33 $\pm$ 0.11)E-5} & \textbf{(1.46 $\pm$ 0.13)E-4} & \textbf{(2.12 $\pm$ 0.16)E-4} & \textbf{(5.78 $\pm$ 0.33)E-5} & \textbf{(8.73 $\pm$ 0.61)E-4} & \textbf{(8.31 $\pm$ 0.41)E-4} \\
\hline
\end{tabular}%
}
\end{table}

\begin{table}[H]
\renewcommand{\underline}[1]{\begingroup\setbox0=\hbox{#1}\leavevmode\vtop{\offinterlineskip\copy0\kern0.25pt\hrule height0.3pt width\wd0}\endgroup}
\centering
\caption{Synthetic held-out results under MWU dynamics. Each entry is mean $\pm$ std over five runs on held-out payoff functions.}
\label{tab:synthetic-mwu}
\scriptsize
\renewcommand{\arraystretch}{1.12}
\setlength{\tabcolsep}{3pt}
\resizebox{\linewidth}{!}{%
\begin{tabular}{c|c|ccc|ccc}
\hline
\multirow{2}{*}{Game} & \multirow{2}{*}{Method} & \multicolumn{3}{c|}{$K=3$} & \multicolumn{3}{c}{$K=4$} \\
\cline{3-8}
& & One-step ($\downarrow$) & Rollout RMSE ($\downarrow$) & Rollout@$T$ ($\downarrow$) & One-step ($\downarrow$) & Rollout RMSE ($\downarrow$) & Rollout@$T$ ($\downarrow$) \\
\hline
\multirow{7}{*}{\begin{tabular}[c]{@{}c@{}}Zero-sum\\cyclic\end{tabular}} & \methodSINDy{} & (1.45 $\pm$ 0.00)E-1 & (5.43 $\pm$ 0.00)E-1 & (5.78 $\pm$ 0.00)E-1 & (2.00 $\pm$ 0.00)E-1 & (3.96 $\pm$ 0.00)E-1 & (4.83 $\pm$ 0.00)E-1 \\
 & \methodSIAR{} & (1.24 $\pm$ 0.00)E-1 & (3.77 $\pm$ 0.00)E-1 & (4.28 $\pm$ 0.00)E-1 & (1.30 $\pm$ 0.00)E-1 & (2.66 $\pm$ 0.00)E-1 & (2.83 $\pm$ 0.00)E-1 \\
 & \methodNeuralODE{} & (3.09 $\pm$ 0.35)E-2 & (3.02 $\pm$ 0.77)E-1 & (4.57 $\pm$ 1.54)E-1 & (1.52 $\pm$ 0.05)E-2 & (1.79 $\pm$ 0.10)E-1 & (2.55 $\pm$ 0.17)E-1 \\
 & \methodLEADS{} & (3.28 $\pm$ 0.32)E-2 & (3.29 $\pm$ 0.69)E-1 & (5.00 $\pm$ 1.21)E-1 & (1.58 $\pm$ 0.17)E-2 & (1.73 $\pm$ 0.12)E-1 & (2.42 $\pm$ 0.15)E-1 \\
 & \methodCoDALOne{} & (2.94 $\pm$ 0.12)E-2 & (2.66 $\pm$ 0.58)E-1 & (4.01 $\pm$ 1.15)E-1 & (1.49 $\pm$ 0.18)E-2 & (1.66 $\pm$ 0.23)E-1 & (2.35 $\pm$ 0.36)E-1 \\
 & \methodCoDALTwo{} & \underline{(2.93 $\pm$ 0.12)E-2} & \underline{(2.63 $\pm$ 0.58)E-1} & \underline{(3.96 $\pm$ 1.14)E-1} & \underline{(1.48 $\pm$ 0.17)E-2} & \underline{(1.64 $\pm$ 0.22)E-1} & \underline{(2.30 $\pm$ 0.36)E-1} \\
\rowcolor{gray!15}
\cellcolor{white} & \methodPARmD{} & \textbf{(2.65 $\pm$ 0.05)E-3} & \textbf{(3.04 $\pm$ 0.29)E-2} & \textbf{(2.71 $\pm$ 0.18)E-2} & \textbf{(1.63 $\pm$ 0.07)E-3} & \textbf{(2.58 $\pm$ 0.11)E-2} & \textbf{(2.51 $\pm$ 0.16)E-2} \\
\hline
\multirow{7}{*}{\begin{tabular}[c]{@{}c@{}}Potential\\congestion\end{tabular}} & \methodSINDy{} & (9.56 $\pm$ 0.00)E-3 & (1.15 $\pm$ 0.00)E-1 & (1.45 $\pm$ 0.00)E-1 & (1.71 $\pm$ 0.00)E-2 & (1.84 $\pm$ 0.00)E-1 & (2.49 $\pm$ 0.00)E-1 \\
 & \methodSIAR{} & (1.50 $\pm$ 0.00)E-2 & (1.70 $\pm$ 0.00)E-1 & (2.59 $\pm$ 0.00)E-1 & (1.15 $\pm$ 0.00)E-2 & (1.04 $\pm$ 0.00)E-1 & (1.44 $\pm$ 0.00)E-1 \\
 & \methodNeuralODE{} & (2.44 $\pm$ 0.21)E-3 & (2.99 $\pm$ 0.23)E-2 & (4.10 $\pm$ 0.24)E-2 & \underline{(3.68 $\pm$ 0.05)E-3} & \underline{(4.10 $\pm$ 0.07)E-2} & \underline{(6.23 $\pm$ 0.12)E-2} \\
 & \methodLEADS{} & (2.51 $\pm$ 0.36)E-3 & (2.99 $\pm$ 0.37)E-2 & (4.20 $\pm$ 0.51)E-2 & (3.91 $\pm$ 0.34)E-3 & (4.34 $\pm$ 0.39)E-2 & (6.60 $\pm$ 0.69)E-2 \\
 & \methodCoDALOne{} & \underline{(2.27 $\pm$ 0.10)E-3} & \underline{(2.79 $\pm$ 0.09)E-2} & \underline{(3.85 $\pm$ 0.15)E-2} & (3.83 $\pm$ 0.20)E-3 & (4.28 $\pm$ 0.26)E-2 & (6.54 $\pm$ 0.41)E-2 \\
 & \methodCoDALTwo{} & (2.29 $\pm$ 0.09)E-3 & (2.81 $\pm$ 0.09)E-2 & (3.87 $\pm$ 0.15)E-2 & (3.84 $\pm$ 0.22)E-3 & (4.29 $\pm$ 0.27)E-2 & (6.56 $\pm$ 0.45)E-2 \\
\rowcolor{gray!15}
\cellcolor{white} & \methodPARmD{} & \textbf{(5.02 $\pm$ 0.52)E-4} & \textbf{(9.37 $\pm$ 0.65)E-3} & \textbf{(1.03 $\pm$ 0.10)E-2} & \textbf{(4.14 $\pm$ 0.41)E-4} & \textbf{(6.57 $\pm$ 0.65)E-3} & \textbf{(7.71 $\pm$ 0.79)E-3} \\
\hline
\end{tabular}%
}
\end{table}

\begin{figure}[H]
\centering
\includegraphics[width=\textwidth]{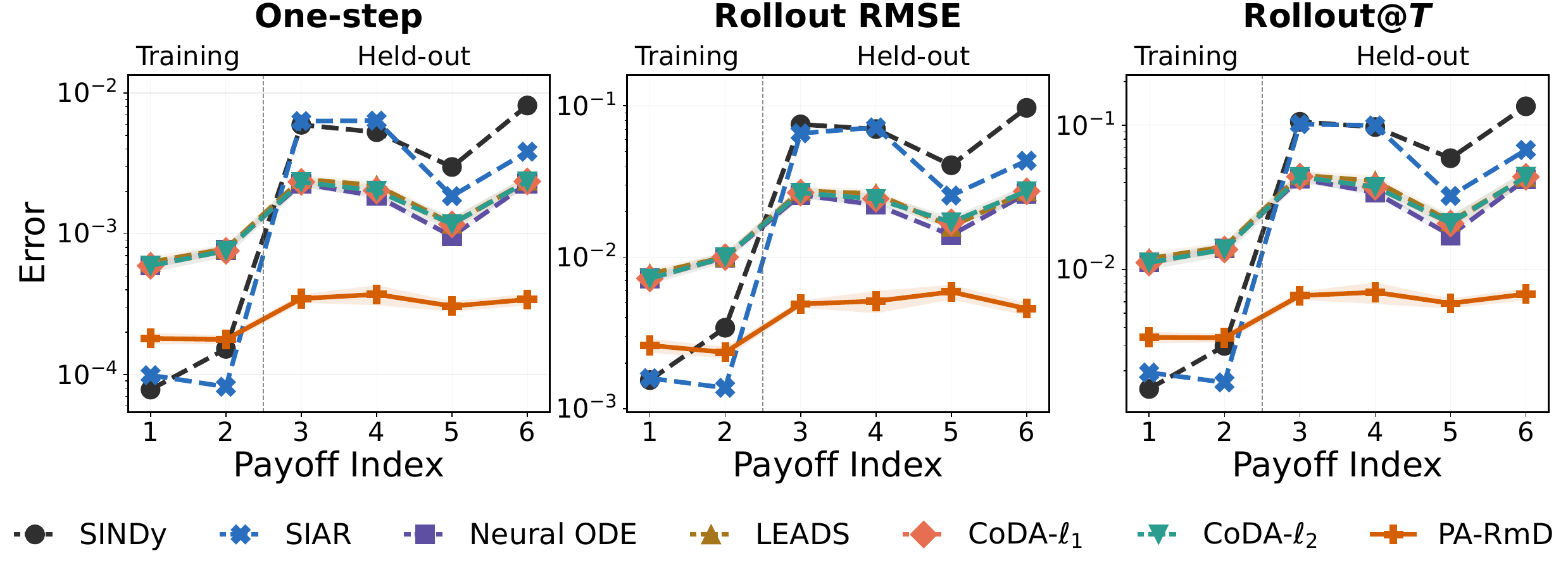}
\caption{Error by payoff function index for the $K=4$ potential congestion game under replicator dynamics.}
\label{fig:app-payoff-index-congestion-replicator}
\end{figure}

\begin{figure}[H]
\centering
\includegraphics[width=\textwidth]{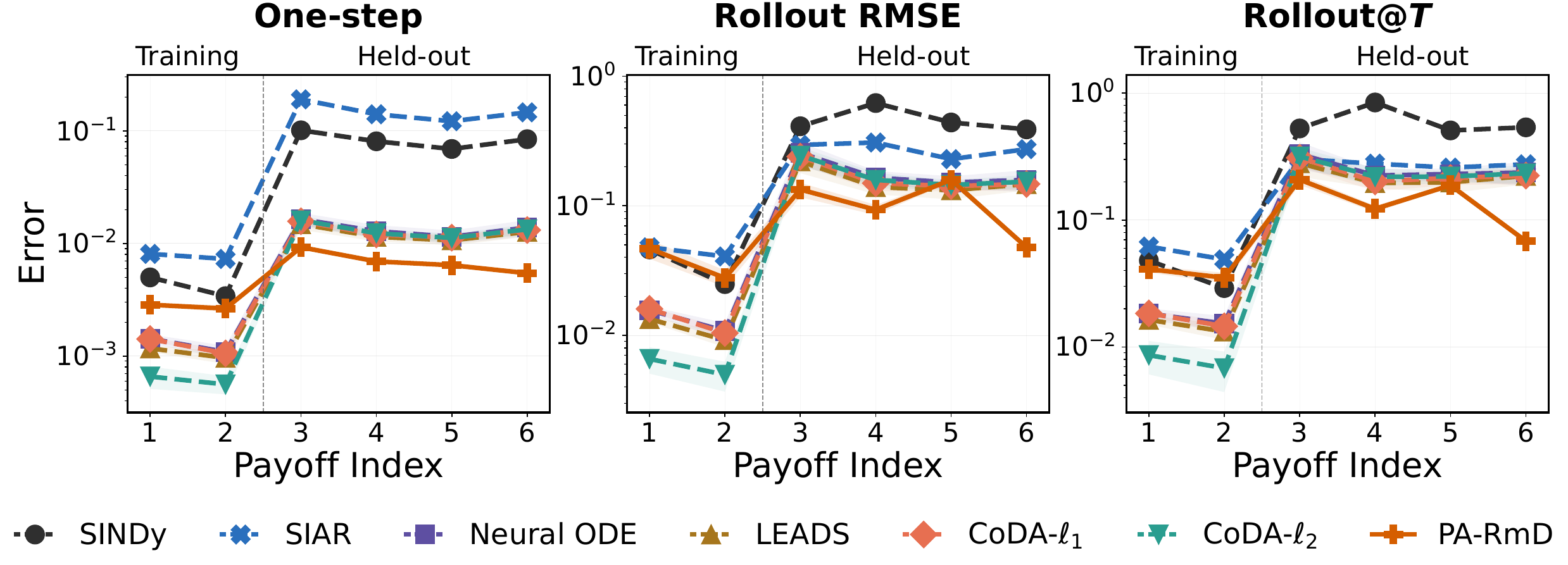}
\caption{Error by payoff function index for the $K=4$ zero-sum cyclic game under projection dynamics.}
\label{fig:app-payoff-index-cyclic-projection}
\end{figure}

\begin{figure}[H]
\centering
\includegraphics[width=\textwidth]{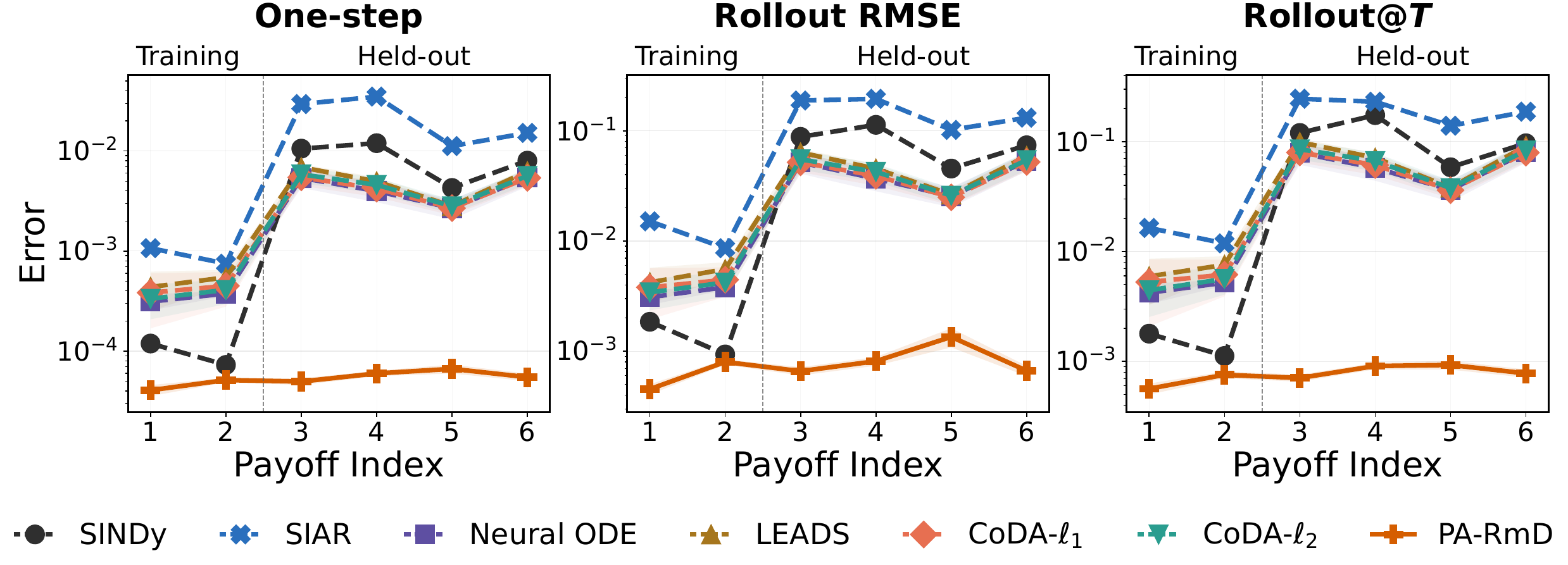}
\caption{Error by payoff function index for the $K=4$ potential congestion game under projection dynamics.}
\label{fig:app-payoff-index-congestion-projection}
\end{figure}

\begin{figure}[H]
\centering
\includegraphics[width=\textwidth]{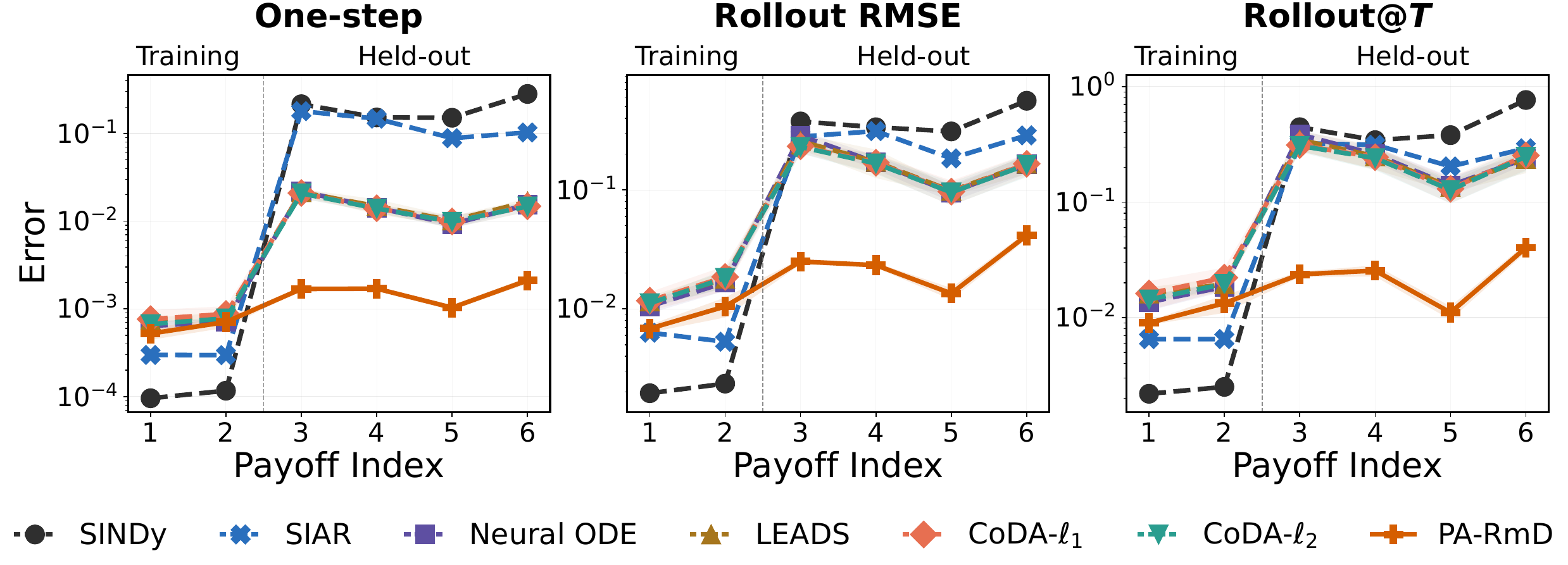}
\caption{Error by payoff function index for the $K=4$ zero-sum cyclic game under MWU dynamics.}
\label{fig:app-payoff-index-cyclic-mwu}
\end{figure}

\begin{figure}[H]
\centering
\includegraphics[width=\textwidth]{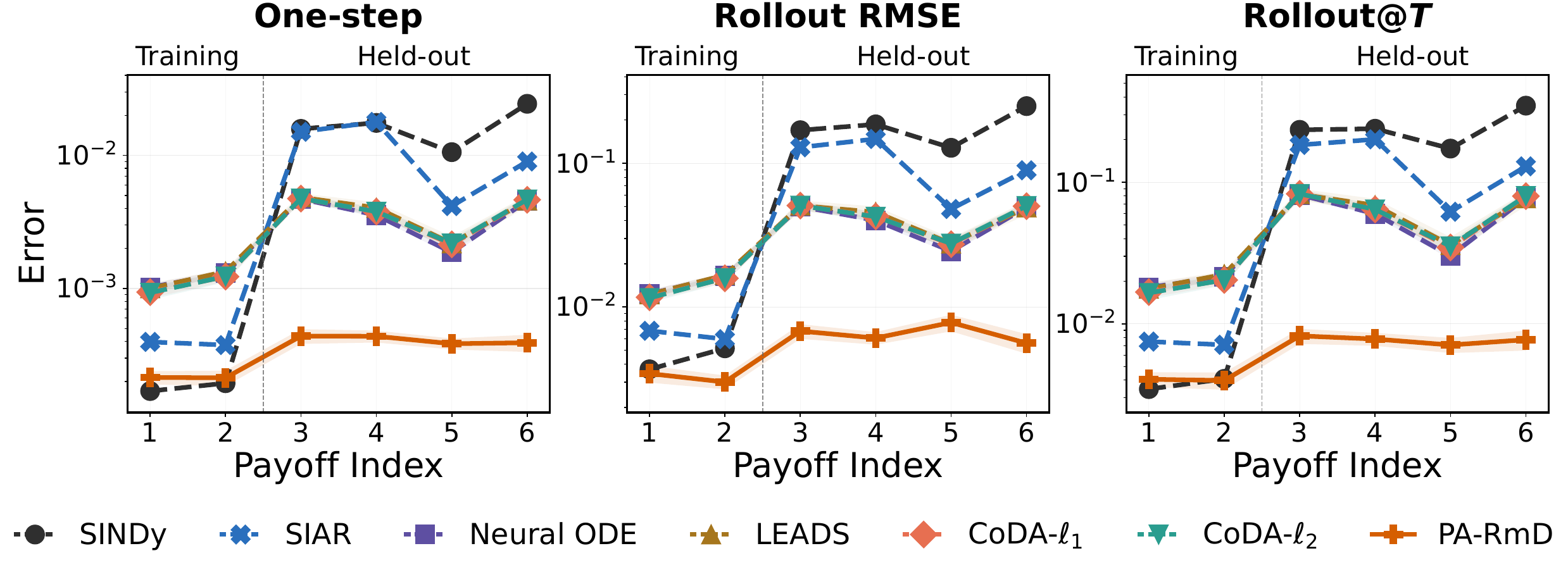}
\caption{Error by payoff function index for the $K=4$ potential congestion game under MWU dynamics.}
\label{fig:app-payoff-index-congestion-mwu}
\end{figure}

\begin{figure}[H]
\centering
\includegraphics[width=\textwidth]{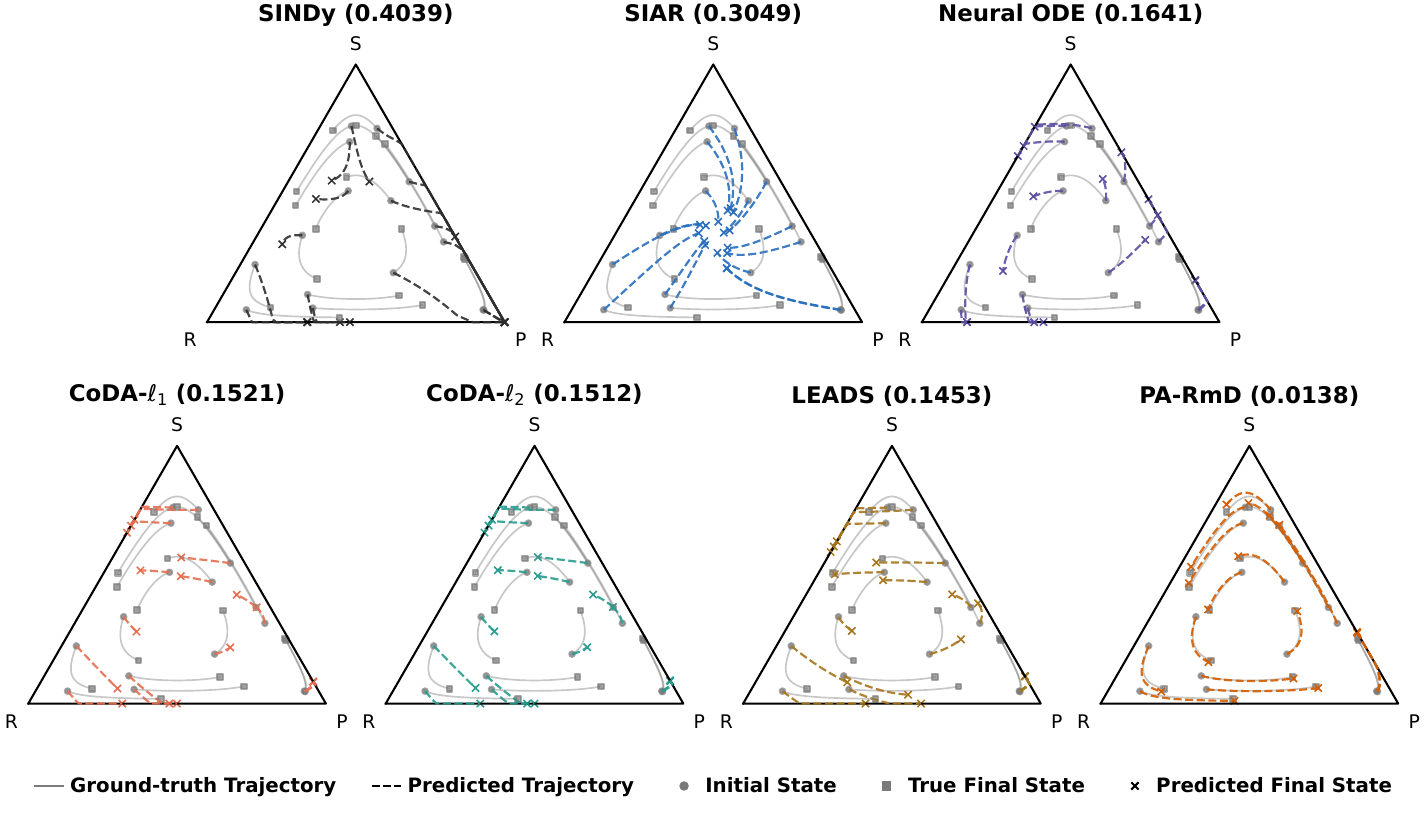}
\caption{Full held-out rollout trajectories for the $K=3$ zero-sum cyclic game under replicator dynamics.}
\label{fig:app-cyclic-rollout-replicator-full}
\end{figure}

\begin{figure}[H]
\centering
\includegraphics[width=\textwidth]{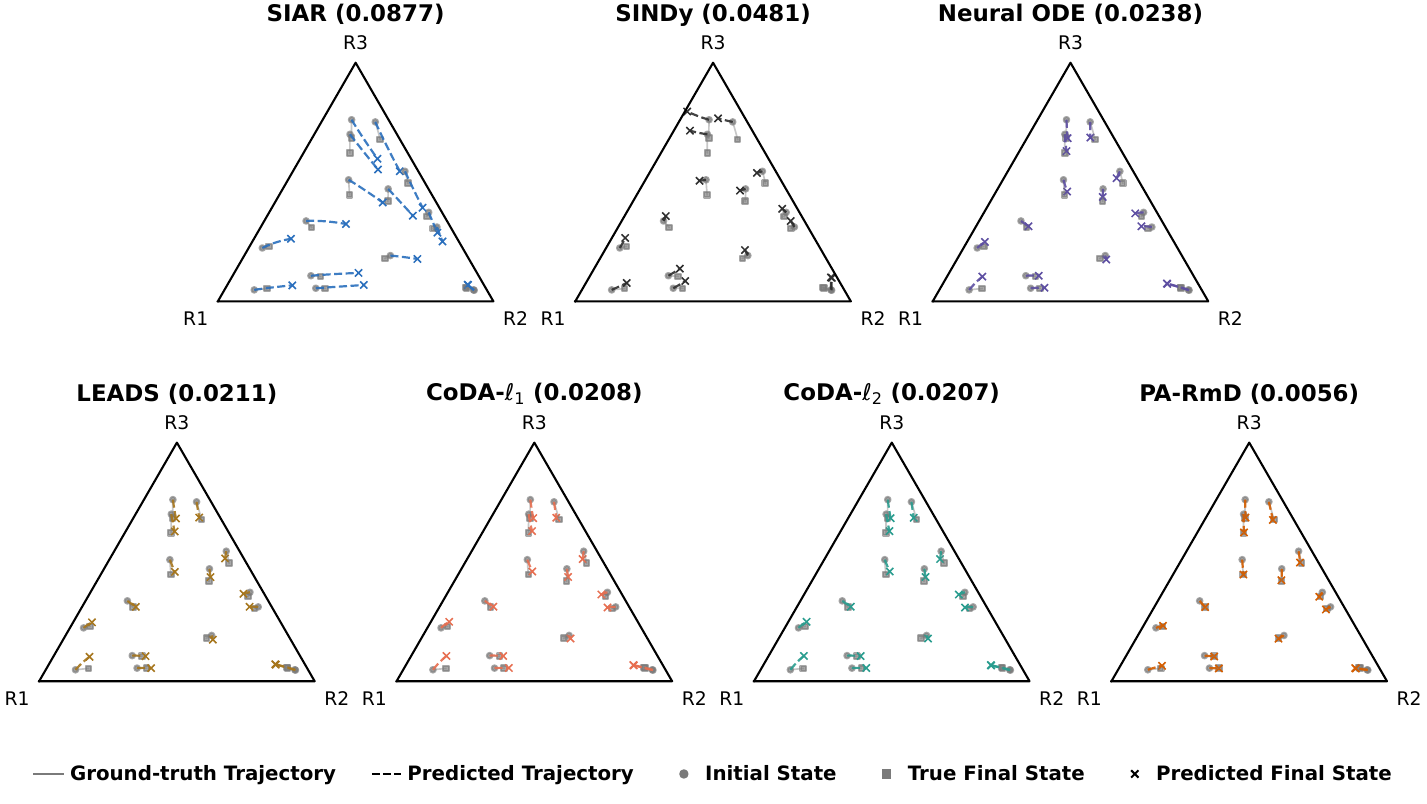}
\caption{Full held-out rollout trajectories for the $K=3$ potential congestion game under replicator dynamics.}
\label{fig:app-congestion-rollout-replicator-full}
\end{figure}

\begin{figure}[H]
\centering
\includegraphics[width=\textwidth]{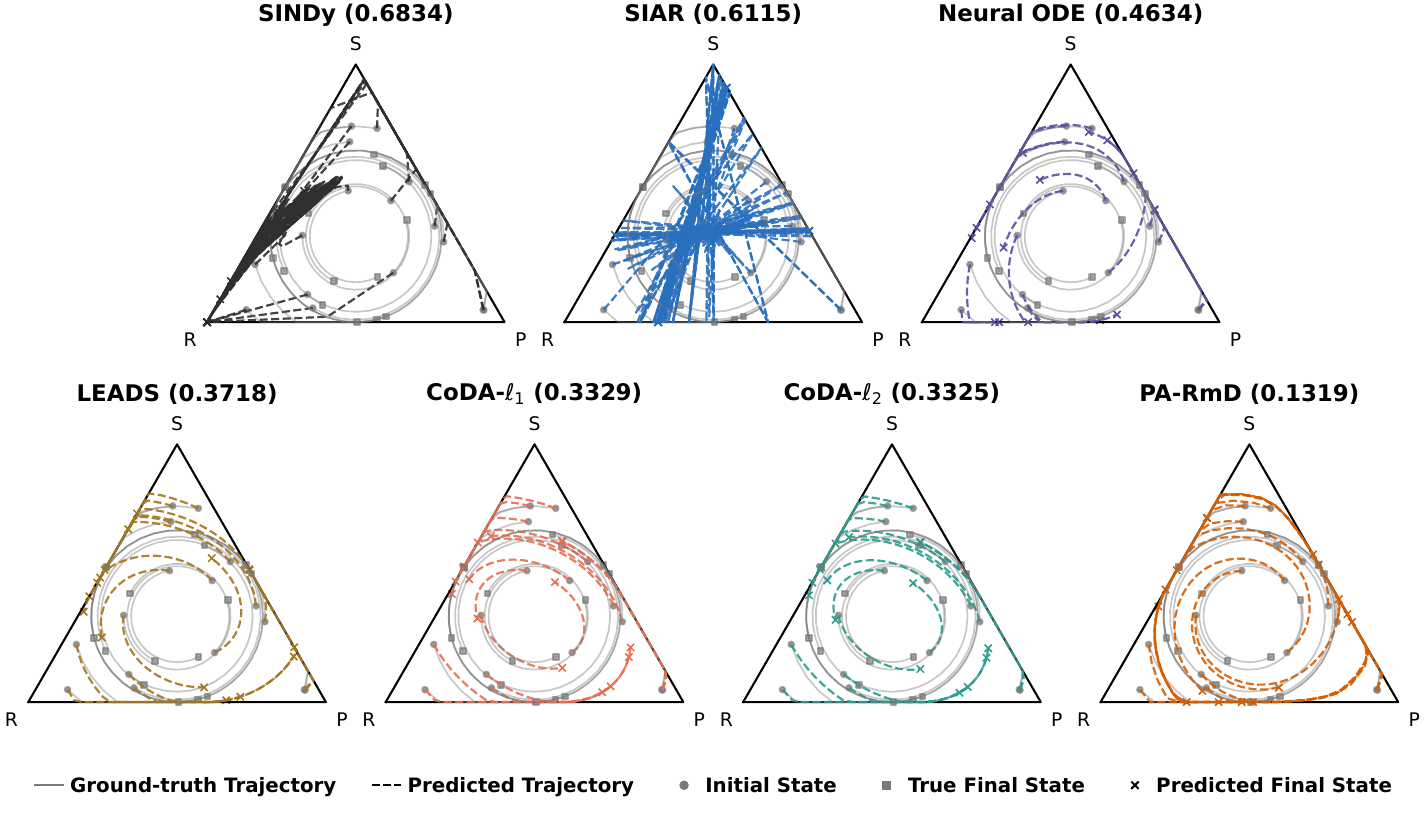}
\caption{Full held-out rollout trajectories for the $K=3$ zero-sum cyclic game under projection dynamics.}
\label{fig:app-cyclic-rollout-projection-full}
\end{figure}

\begin{figure}[H]
\centering
\includegraphics[width=\textwidth]{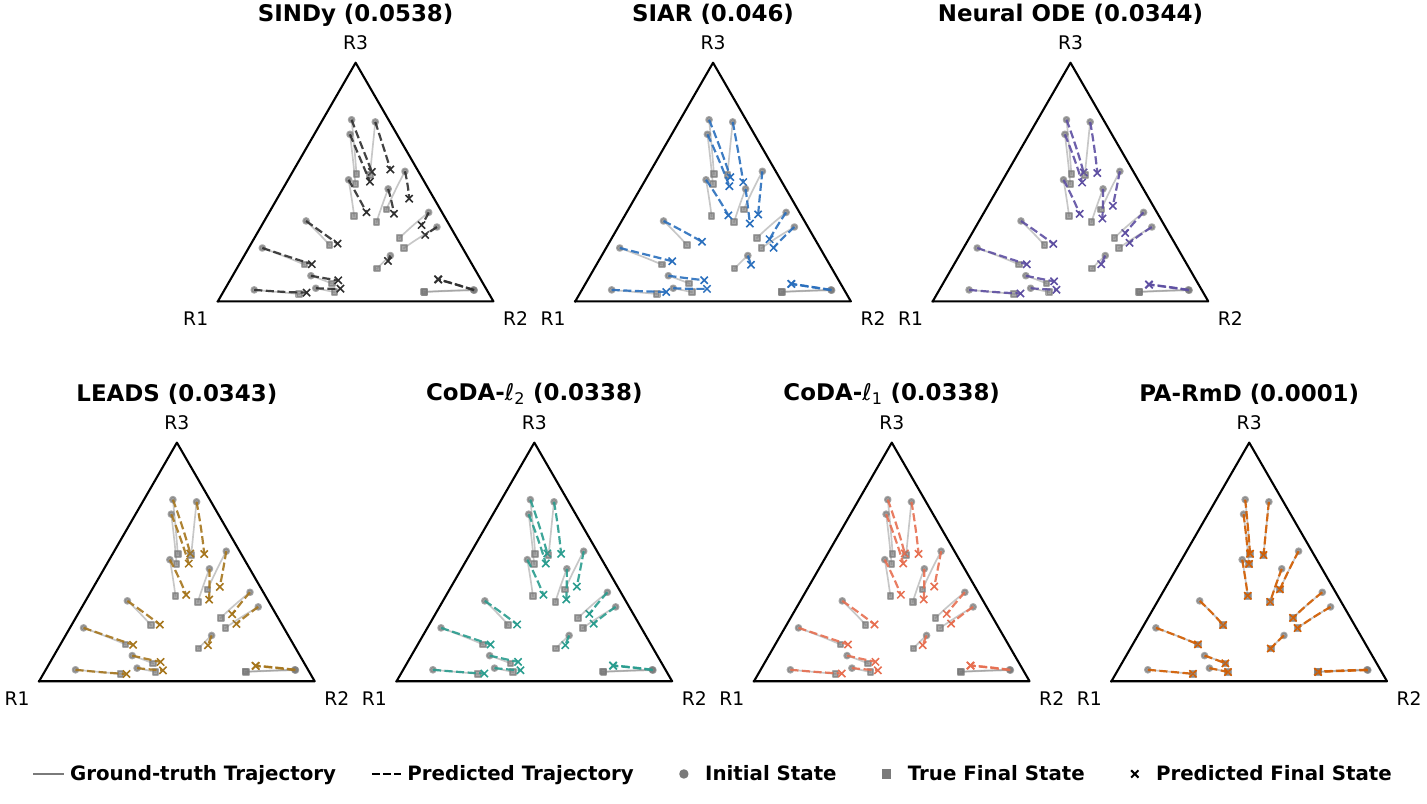}
\caption{Full held-out rollout trajectories for the $K=3$ potential congestion game under projection dynamics.}
\label{fig:app-congestion-rollout-projection-full}
\end{figure}

\begin{figure}[H]
\centering
\includegraphics[width=\textwidth]{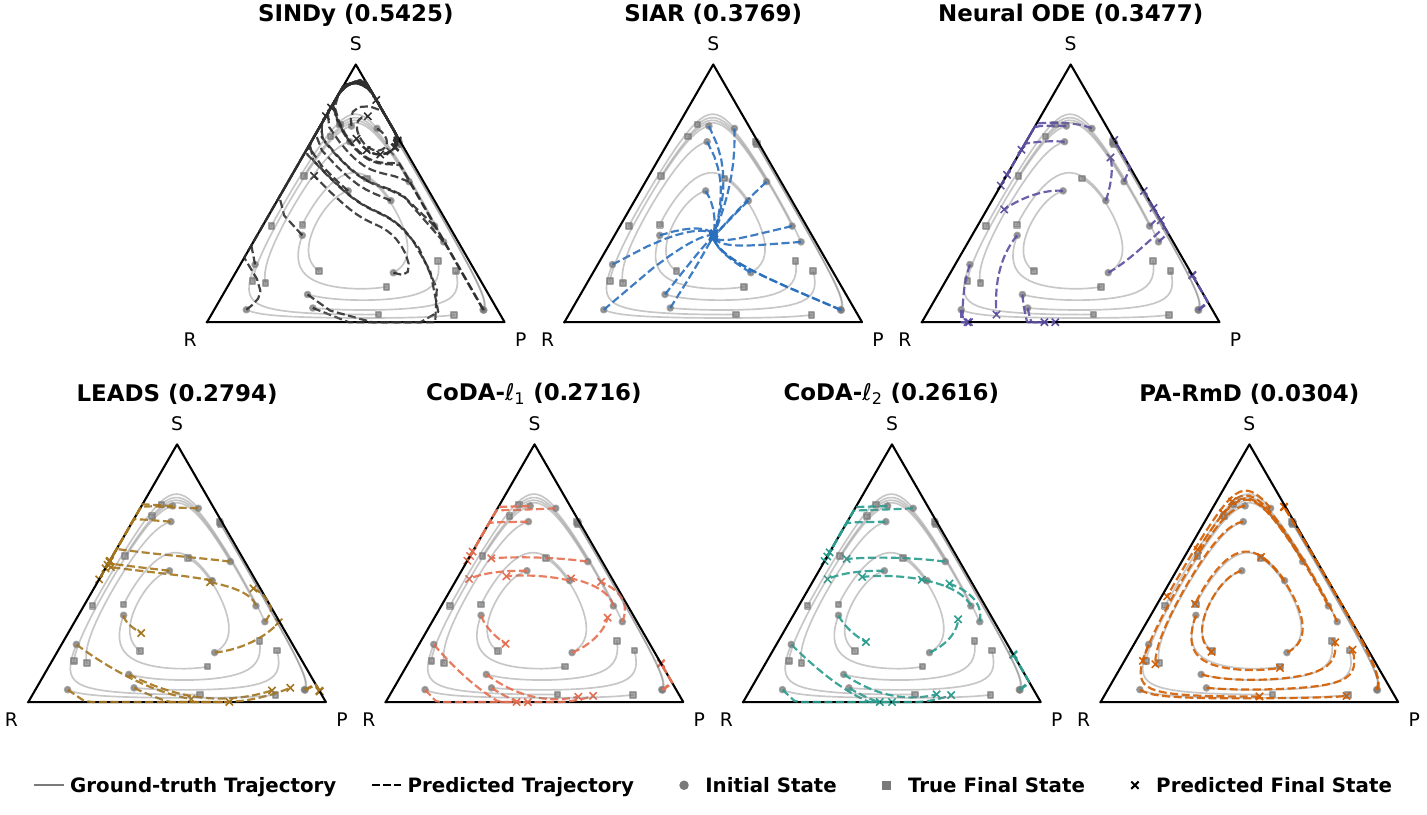}
\caption{Full held-out rollout trajectories for the $K=3$ zero-sum cyclic game under MWU dynamics.}
\label{fig:app-cyclic-rollout-mwu-full}
\end{figure}

\begin{figure}[H]
\centering
\includegraphics[width=\textwidth]{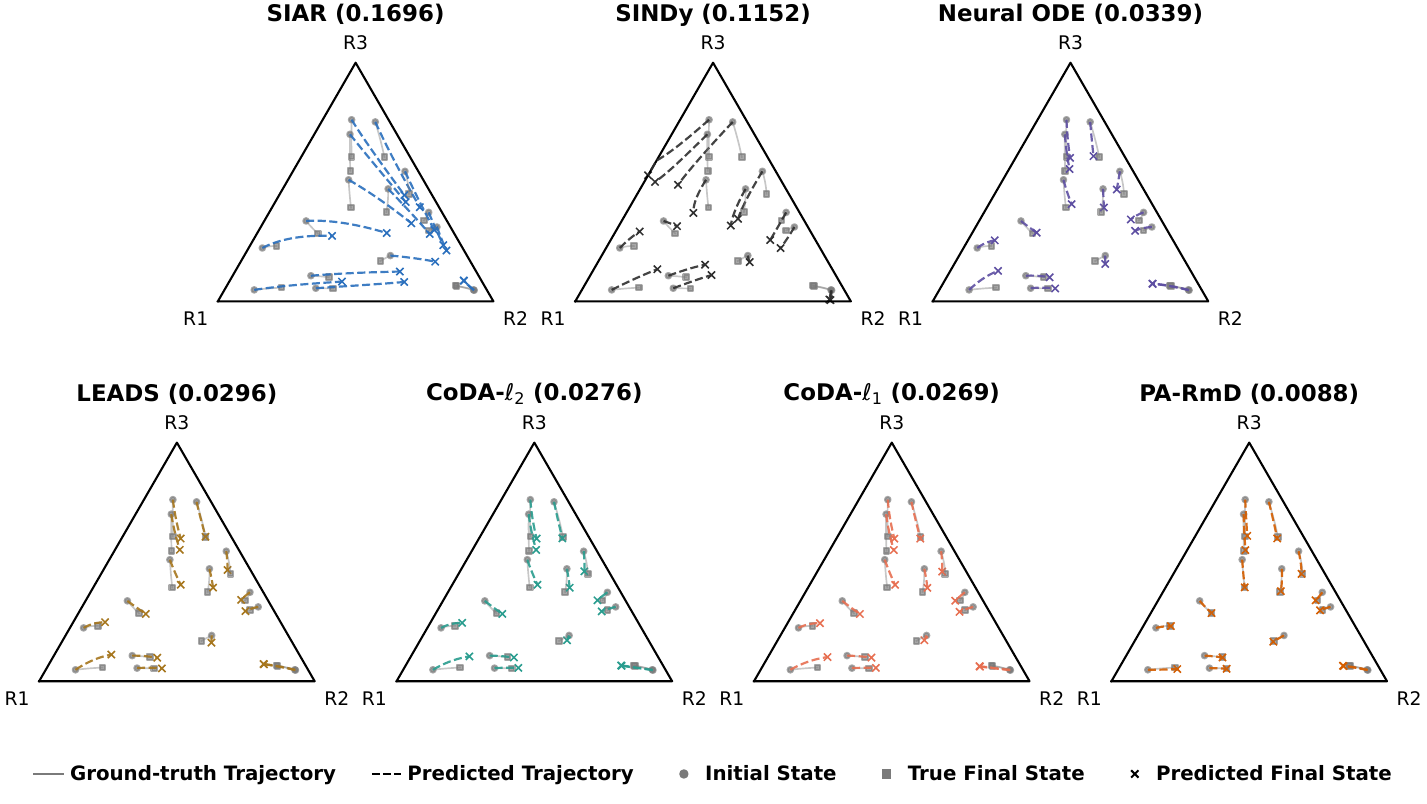}
\caption{Full held-out rollout trajectories for the $K=3$ potential congestion game under MWU dynamics.}
\label{fig:app-congestion-rollout-mwu-full}
\end{figure}

\subsubsection{Results on Overall and Seen Payoff Environments}
\label{app:synthetic-results-overall-seen}

\begin{table}[H]
\renewcommand{\underline}[1]{\begingroup\setbox0=\hbox{#1}\leavevmode\vtop{\offinterlineskip\copy0\kern0.25pt\hrule height0.3pt width\wd0}\endgroup}
\centering
\caption{Overall results across both seen and held-out payoff environments under replicator dynamics. Each entry is mean $\pm$ std over five runs.}
\label{tab:app-synthetic-overall-replicator}
\scriptsize
\renewcommand{\arraystretch}{1.12}
\setlength{\tabcolsep}{3pt}
\resizebox{\linewidth}{!}{%
\begin{tabular}{c|c|ccc|ccc}
\hline
\multirow{2}{*}{Game} & \multirow{2}{*}{Method} & \multicolumn{3}{c|}{$K=3$} & \multicolumn{3}{c}{$K=4$} \\
\cline{3-8}
& & One-step ($\downarrow$) & Rollout RMSE ($\downarrow$) & Rollout@$T$ ($\downarrow$) & One-step ($\downarrow$) & Rollout RMSE ($\downarrow$) & Rollout@$T$ ($\downarrow$) \\
\hline
\multirow{7}{*}{\begin{tabular}[c]{@{}c@{}}Zero-sum\\cyclic\end{tabular}} & \methodSINDy{} & (2.02 $\pm$ 0.00)E-2 & (2.02 $\pm$ 0.00)E-1 & (2.62 $\pm$ 0.00)E-1 & (7.28 $\pm$ 0.00)E-2 & (1.92 $\pm$ 0.00)E-1 & (2.05 $\pm$ 0.00)E-1 \\
 & \methodSIAR{} & (2.52 $\pm$ 0.00)E-2 & (1.52 $\pm$ 0.00)E-1 & (2.00 $\pm$ 0.00)E-1 & (3.90 $\pm$ 0.00)E-2 & (1.41 $\pm$ 0.00)E-1 & (1.39 $\pm$ 0.00)E-1 \\
 & \methodNeuralODE{} & (7.44 $\pm$ 0.19)E-3 & (7.70 $\pm$ 0.79)E-2 & (1.09 $\pm$ 0.14)E-1 & \underline{(4.85 $\pm$ 0.12)E-3} & \underline{(6.04 $\pm$ 0.44)E-2} & \underline{(8.27 $\pm$ 0.48)E-2} \\
 & \methodLEADS{} & (8.52 $\pm$ 0.91)E-3 & (9.01 $\pm$ 1.52)E-2 & (1.35 $\pm$ 0.28)E-1 & (5.36 $\pm$ 0.82)E-3 & (6.77 $\pm$ 1.04)E-2 & (9.11 $\pm$ 1.06)E-2 \\
 & \methodCoDALOne{} & \underline{(7.29 $\pm$ 0.09)E-3} & (7.28 $\pm$ 0.67)E-2 & (1.05 $\pm$ 0.15)E-1 & (5.07 $\pm$ 0.64)E-3 & (6.52 $\pm$ 0.60)E-2 & (8.75 $\pm$ 0.99)E-2 \\
 & \methodCoDALTwo{} & (7.29 $\pm$ 0.11)E-3 & \underline{(7.27 $\pm$ 0.60)E-2} & \underline{(1.05 $\pm$ 0.14)E-1} & (5.03 $\pm$ 0.68)E-3 & (6.47 $\pm$ 0.66)E-2 & (8.62 $\pm$ 1.11)E-2 \\
\rowcolor{gray!15}
\cellcolor{white} & \methodPARmD{} & \textbf{(7.69 $\pm$ 0.24)E-4} & \textbf{(1.24 $\pm$ 0.36)E-2} & \textbf{(1.43 $\pm$ 0.25)E-2} & \textbf{(8.17 $\pm$ 0.27)E-4} & \textbf{(1.41 $\pm$ 0.06)E-2} & \textbf{(1.59 $\pm$ 0.06)E-2} \\
\hline
\multirow{7}{*}{\begin{tabular}[c]{@{}c@{}}Potential\\congestion\end{tabular}} & \methodSINDy{} & (1.66 $\pm$ 0.00)E-3 & (2.41 $\pm$ 0.00)E-2 & (3.60 $\pm$ 0.00)E-2 & (3.76 $\pm$ 0.00)E-3 & (4.80 $\pm$ 0.00)E-2 & (6.68 $\pm$ 0.00)E-2 \\
 & \methodSIAR{} & (3.59 $\pm$ 0.00)E-3 & (4.41 $\pm$ 0.00)E-2 & (6.92 $\pm$ 0.00)E-2 & (3.09 $\pm$ 0.00)E-3 & (3.49 $\pm$ 0.00)E-2 & (5.07 $\pm$ 0.00)E-2 \\
 & \methodNeuralODE{} & \underline{(1.23 $\pm$ 0.09)E-3} & (1.60 $\pm$ 0.14)E-2 & \underline{(2.25 $\pm$ 0.16)E-2} & \underline{(1.45 $\pm$ 0.04)E-3} & \underline{(1.76 $\pm$ 0.05)E-2} & \underline{(2.69 $\pm$ 0.09)E-2} \\
 & \methodLEADS{} & (1.23 $\pm$ 0.08)E-3 & \underline{(1.59 $\pm$ 0.09)E-2} & (2.26 $\pm$ 0.13)E-2 & (1.60 $\pm$ 0.11)E-3 & (1.91 $\pm$ 0.12)E-2 & (2.97 $\pm$ 0.23)E-2 \\
 & \methodCoDALOne{} & (1.23 $\pm$ 0.03)E-3 & (1.61 $\pm$ 0.06)E-2 & (2.26 $\pm$ 0.05)E-2 & (1.54 $\pm$ 0.13)E-3 & (1.88 $\pm$ 0.15)E-2 & (2.85 $\pm$ 0.26)E-2 \\
 & \methodCoDALTwo{} & (1.23 $\pm$ 0.03)E-3 & (1.61 $\pm$ 0.06)E-2 & (2.26 $\pm$ 0.05)E-2 & (1.54 $\pm$ 0.13)E-3 & (1.87 $\pm$ 0.15)E-2 & (2.85 $\pm$ 0.27)E-2 \\
\rowcolor{gray!15}
\cellcolor{white} & \methodPARmD{} & \textbf{(2.40 $\pm$ 0.26)E-4} & \textbf{(4.02 $\pm$ 0.31)E-3} & \textbf{(4.76 $\pm$ 0.48)E-3} & \textbf{(2.87 $\pm$ 0.25)E-4} & \textbf{(4.25 $\pm$ 0.39)E-3} & \textbf{(5.51 $\pm$ 0.47)E-3} \\
\hline
\end{tabular}%
}
\end{table}

\begin{table}[H]
\renewcommand{\underline}[1]{\begingroup\setbox0=\hbox{#1}\leavevmode\vtop{\offinterlineskip\copy0\kern0.25pt\hrule height0.3pt width\wd0}\endgroup}
\centering
\caption{Overall results across both seen and held-out payoff environments under projection dynamics. Each entry is mean $\pm$ std over five runs.}
\label{tab:app-synthetic-overall-projection}
\scriptsize
\renewcommand{\arraystretch}{1.12}
\setlength{\tabcolsep}{3pt}
\resizebox{\linewidth}{!}{%
\begin{tabular}{c|c|ccc|ccc}
\hline
\multirow{2}{*}{Game} & \multirow{2}{*}{Method} & \multicolumn{3}{c|}{$K=3$} & \multicolumn{3}{c}{$K=4$} \\
\cline{3-8}
& & One-step ($\downarrow$) & Rollout RMSE ($\downarrow$) & Rollout@$T$ ($\downarrow$) & One-step ($\downarrow$) & Rollout RMSE ($\downarrow$) & Rollout@$T$ ($\downarrow$) \\
\hline
\multirow{7}{*}{\begin{tabular}[c]{@{}c@{}}Zero-sum\\cyclic\end{tabular}} & \methodSINDy{} & (1.94 $\pm$ 0.00)E-1 & (3.50 $\pm$ 0.00)E-1 & (4.04 $\pm$ 0.00)E-1 & (5.73 $\pm$ 0.00)E-2 & (3.22 $\pm$ 0.00)E-1 & (4.16 $\pm$ 0.00)E-1 \\
 & \methodSIAR{} & (3.88 $\pm$ 0.00)E-1 & (3.30 $\pm$ 0.00)E-1 & (3.89 $\pm$ 0.00)E-1 & (1.02 $\pm$ 0.00)E-1 & (1.99 $\pm$ 0.00)E-1 & (2.04 $\pm$ 0.00)E-1 \\
 & \methodNeuralODE{} & (1.61 $\pm$ 0.21)E-2 & \underline{(1.79 $\pm$ 0.36)E-1} & \underline{(2.81 $\pm$ 0.56)E-1} & (9.51 $\pm$ 1.31)E-3 & (1.26 $\pm$ 0.19)E-1 & (1.76 $\pm$ 0.28)E-1 \\
 & \methodLEADS{} & (1.80 $\pm$ 0.15)E-2 & (2.23 $\pm$ 0.37)E-1 & (3.47 $\pm$ 0.54)E-1 & \underline{(8.62 $\pm$ 0.80)E-3} & \underline{(1.10 $\pm$ 0.14)E-1} & \underline{(1.55 $\pm$ 0.21)E-1} \\
 & \methodCoDALOne{} & \underline{(1.57 $\pm$ 0.19)E-2} & (1.84 $\pm$ 0.21)E-1 & (2.95 $\pm$ 0.28)E-1 & (9.10 $\pm$ 0.59)E-3 & (1.17 $\pm$ 0.09)E-1 & (1.64 $\pm$ 0.14)E-1 \\
 & \methodCoDALTwo{} & (1.58 $\pm$ 0.19)E-2 & (1.84 $\pm$ 0.20)E-1 & (2.95 $\pm$ 0.28)E-1 & (9.02 $\pm$ 0.83)E-3 & (1.19 $\pm$ 0.12)E-1 & (1.68 $\pm$ 0.17)E-1 \\
\rowcolor{gray!15}
\cellcolor{white} & \methodPARmD{} & \textbf{(5.31 $\pm$ 0.24)E-3} & \textbf{(7.21 $\pm$ 0.26)E-2} & \textbf{(1.09 $\pm$ 0.04)E-1} & \textbf{(5.59 $\pm$ 0.04)E-3} & \textbf{(8.47 $\pm$ 0.47)E-2} & \textbf{(1.11 $\pm$ 0.05)E-1} \\
\hline
\multirow{7}{*}{\begin{tabular}[c]{@{}c@{}}Potential\\congestion\end{tabular}} & \methodSINDy{} & (2.24 $\pm$ 0.00)E-3 & (2.69 $\pm$ 0.00)E-2 & (3.96 $\pm$ 0.00)E-2 & (5.83 $\pm$ 0.00)E-3 & (5.39 $\pm$ 0.00)E-2 & (7.53 $\pm$ 0.00)E-2 \\
 & \methodSIAR{} & (2.39 $\pm$ 0.00)E-3 & (2.32 $\pm$ 0.00)E-2 & (3.73 $\pm$ 0.00)E-2 & (1.54 $\pm$ 0.00)E-2 & (1.07 $\pm$ 0.00)E-1 & (1.38 $\pm$ 0.00)E-1 \\
 & \methodNeuralODE{} & (1.76 $\pm$ 0.15)E-3 & (1.78 $\pm$ 0.15)E-2 & (2.77 $\pm$ 0.24)E-2 & \underline{(3.03 $\pm$ 0.59)E-3} & \underline{(2.90 $\pm$ 0.57)E-2} & \underline{(4.38 $\pm$ 0.89)E-2} \\
 & \methodLEADS{} & (1.74 $\pm$ 0.08)E-3 & (1.76 $\pm$ 0.08)E-2 & (2.74 $\pm$ 0.13)E-2 & (3.65 $\pm$ 0.27)E-3 & (3.40 $\pm$ 0.24)E-2 & (5.21 $\pm$ 0.39)E-2 \\
 & \methodCoDALOne{} & \underline{(1.64 $\pm$ 0.17)E-3} & \underline{(1.66 $\pm$ 0.18)E-2} & \underline{(2.59 $\pm$ 0.27)E-2} & (3.10 $\pm$ 0.49)E-3 & (2.94 $\pm$ 0.49)E-2 & (4.46 $\pm$ 0.76)E-2 \\
 & \methodCoDALTwo{} & (1.65 $\pm$ 0.17)E-3 & (1.66 $\pm$ 0.18)E-2 & (2.60 $\pm$ 0.27)E-2 & (3.27 $\pm$ 0.56)E-3 & (3.09 $\pm$ 0.56)E-2 & (4.71 $\pm$ 0.87)E-2 \\
\rowcolor{gray!15}
\cellcolor{white} & \methodPARmD{} & \textbf{(7.41 $\pm$ 1.39)E-6} & \textbf{(8.27 $\pm$ 1.61)E-5} & \textbf{(1.17 $\pm$ 0.21)E-4} & \textbf{(5.39 $\pm$ 0.29)E-5} & \textbf{(7.91 $\pm$ 0.48)E-4} & \textbf{(7.74 $\pm$ 0.38)E-4} \\
\hline
\end{tabular}%
}
\end{table}

\begin{table}[H]
\renewcommand{\underline}[1]{\begingroup\setbox0=\hbox{#1}\leavevmode\vtop{\offinterlineskip\copy0\kern0.25pt\hrule height0.3pt width\wd0}\endgroup}
\centering
\caption{Overall results across both seen and held-out payoff environments under MWU dynamics. Each entry is mean $\pm$ std over five runs.}
\label{tab:app-synthetic-overall-mwu}
\scriptsize
\renewcommand{\arraystretch}{1.12}
\setlength{\tabcolsep}{3pt}
\resizebox{\linewidth}{!}{%
\begin{tabular}{c|c|ccc|ccc}
\hline
\multirow{2}{*}{Game} & \multirow{2}{*}{Method} & \multicolumn{3}{c|}{$K=3$} & \multicolumn{3}{c}{$K=4$} \\
\cline{3-8}
& & One-step ($\downarrow$) & Rollout RMSE ($\downarrow$) & Rollout@$T$ ($\downarrow$) & One-step ($\downarrow$) & Rollout RMSE ($\downarrow$) & Rollout@$T$ ($\downarrow$) \\
\hline
\multirow{7}{*}{\begin{tabular}[c]{@{}c@{}}Zero-sum\\cyclic\end{tabular}} & \methodSINDy{} & (7.24 $\pm$ 0.00)E-2 & (2.72 $\pm$ 0.00)E-1 & (2.91 $\pm$ 0.00)E-1 & (1.34 $\pm$ 0.00)E-1 & (2.65 $\pm$ 0.00)E-1 & (3.23 $\pm$ 0.00)E-1 \\
 & \methodSIAR{} & (6.21 $\pm$ 0.00)E-2 & (1.91 $\pm$ 0.00)E-1 & (2.18 $\pm$ 0.00)E-1 & (8.65 $\pm$ 0.00)E-2 & (1.79 $\pm$ 0.00)E-1 & (1.91 $\pm$ 0.00)E-1 \\
 & \methodNeuralODE{} & (1.57 $\pm$ 0.17)E-2 & (1.55 $\pm$ 0.37)E-1 & (2.35 $\pm$ 0.76)E-1 & (1.04 $\pm$ 0.04)E-2 & (1.24 $\pm$ 0.07)E-1 & (1.76 $\pm$ 0.12)E-1 \\
 & \methodLEADS{} & (1.67 $\pm$ 0.17)E-2 & (1.69 $\pm$ 0.35)E-1 & (2.57 $\pm$ 0.61)E-1 & (1.08 $\pm$ 0.12)E-2 & (1.20 $\pm$ 0.08)E-1 & (1.67 $\pm$ 0.10)E-1 \\
 & \methodCoDALOne{} & (1.49 $\pm$ 0.06)E-2 & (1.36 $\pm$ 0.29)E-1 & (2.05 $\pm$ 0.57)E-1 & (1.02 $\pm$ 0.12)E-2 & (1.16 $\pm$ 0.15)E-1 & (1.63 $\pm$ 0.24)E-1 \\
 & \methodCoDALTwo{} & \underline{(1.49 $\pm$ 0.06)E-2} & \underline{(1.35 $\pm$ 0.28)E-1} & \underline{(2.03 $\pm$ 0.57)E-1} & \underline{(1.01 $\pm$ 0.11)E-2} & \underline{(1.14 $\pm$ 0.15)E-1} & \underline{(1.59 $\pm$ 0.24)E-1} \\
\rowcolor{gray!15}
\cellcolor{white} & \methodPARmD{} & \textbf{(1.42 $\pm$ 0.02)E-3} & \textbf{(1.68 $\pm$ 0.11)E-2} & \textbf{(1.53 $\pm$ 0.05)E-2} & \textbf{(1.29 $\pm$ 0.06)E-3} & \textbf{(2.01 $\pm$ 0.08)E-2} & \textbf{(2.05 $\pm$ 0.12)E-2} \\
\hline
\multirow{7}{*}{\begin{tabular}[c]{@{}c@{}}Potential\\congestion\end{tabular}} & \methodSINDy{} & (4.79 $\pm$ 0.00)E-3 & (5.77 $\pm$ 0.00)E-2 & (7.28 $\pm$ 0.00)E-2 & (1.15 $\pm$ 0.00)E-2 & (1.24 $\pm$ 0.00)E-1 & (1.67 $\pm$ 0.00)E-1 \\
 & \methodSIAR{} & (7.51 $\pm$ 0.00)E-3 & (8.51 $\pm$ 0.00)E-2 & (1.30 $\pm$ 0.00)E-1 & (7.80 $\pm$ 0.00)E-3 & (7.13 $\pm$ 0.00)E-2 & (9.85 $\pm$ 0.00)E-2 \\
 & \methodNeuralODE{} & (1.67 $\pm$ 0.22)E-3 & (2.07 $\pm$ 0.27)E-2 & (2.84 $\pm$ 0.31)E-2 & \underline{(2.84 $\pm$ 0.03)E-3} & \underline{(3.22 $\pm$ 0.05)E-2} & \underline{(4.82 $\pm$ 0.08)E-2} \\
 & \methodLEADS{} & (1.72 $\pm$ 0.32)E-3 & (2.09 $\pm$ 0.36)E-2 & (2.90 $\pm$ 0.51)E-2 & (3.00 $\pm$ 0.24)E-3 & (3.37 $\pm$ 0.27)E-2 & (5.07 $\pm$ 0.48)E-2 \\
 & \methodCoDALOne{} & \underline{(1.50 $\pm$ 0.07)E-3} & \underline{(1.86 $\pm$ 0.09)E-2} & \underline{(2.54 $\pm$ 0.10)E-2} & (2.91 $\pm$ 0.16)E-3 & (3.31 $\pm$ 0.19)E-2 & (4.98 $\pm$ 0.32)E-2 \\
 & \methodCoDALTwo{} & (1.52 $\pm$ 0.08)E-3 & (1.89 $\pm$ 0.10)E-2 & (2.57 $\pm$ 0.11)E-2 & (2.92 $\pm$ 0.18)E-3 & (3.32 $\pm$ 0.21)E-2 & (4.99 $\pm$ 0.35)E-2 \\
\rowcolor{gray!15}
\cellcolor{white} & \methodPARmD{} & \textbf{(3.16 $\pm$ 0.39)E-4} & \textbf{(5.98 $\pm$ 0.66)E-3} & \textbf{(6.43 $\pm$ 0.81)E-3} & \textbf{(3.47 $\pm$ 0.33)E-4} & \textbf{(5.46 $\pm$ 0.52)E-3} & \textbf{(6.48 $\pm$ 0.65)E-3} \\
\hline
\end{tabular}%
}
\end{table}

\begin{table}[H]
\renewcommand{\underline}[1]{\begingroup\setbox0=\hbox{#1}\leavevmode\vtop{\offinterlineskip\copy0\kern0.25pt\hrule height0.3pt width\wd0}\endgroup}
\centering
\caption{Results on seen training payoff environments under replicator dynamics. Each entry is mean $\pm$ std over five runs.}
\label{tab:app-synthetic-seen-replicator}
\scriptsize
\renewcommand{\arraystretch}{1.12}
\setlength{\tabcolsep}{3pt}
\resizebox{\linewidth}{!}{%
\begin{tabular}{c|c|ccc|ccc}
\hline
\multirow{2}{*}{Game} & \multirow{2}{*}{Method} & \multicolumn{3}{c|}{$K=3$} & \multicolumn{3}{c}{$K=4$} \\
\cline{3-8}
& & One-step ($\downarrow$) & Rollout RMSE ($\downarrow$) & Rollout@$T$ ($\downarrow$) & One-step ($\downarrow$) & Rollout RMSE ($\downarrow$) & Rollout@$T$ ($\downarrow$) \\
\hline
\multirow{7}{*}{\begin{tabular}[c]{@{}c@{}}Zero-sum\\cyclic\end{tabular}} & \methodSINDy{} & \textbf{(2.81 $\pm$ 0.00)E-6} & \textbf{(3.78 $\pm$ 0.00)E-5} & \textbf{(4.67 $\pm$ 0.00)E-5} & \textbf{(6.15 $\pm$ 0.00)E-5} & \textbf{(9.58 $\pm$ 0.00)E-4} & \textbf{(1.21 $\pm$ 0.00)E-3} \\
 & \methodSIAR{} & \underline{(4.06 $\pm$ 0.00)E-6} & \underline{(5.55 $\pm$ 0.00)E-5} & \underline{(7.46 $\pm$ 0.00)E-5} & \underline{(1.23 $\pm$ 0.00)E-4} & \underline{(2.04 $\pm$ 0.00)E-3} & \underline{(2.56 $\pm$ 0.00)E-3} \\
 & \methodNeuralODE{} & (3.84 $\pm$ 1.11)E-4 & (5.10 $\pm$ 1.58)E-3 & (8.13 $\pm$ 2.93)E-3 & (4.19 $\pm$ 0.46)E-4 & (6.76 $\pm$ 1.45)E-3 & (9.14 $\pm$ 1.43)E-3 \\
 & \methodLEADS{} & (3.35 $\pm$ 0.93)E-4 & (5.28 $\pm$ 1.22)E-3 & (7.55 $\pm$ 2.57)E-3 & (4.78 $\pm$ 0.78)E-4 & (8.71 $\pm$ 1.61)E-3 & (1.08 $\pm$ 0.19)E-2 \\
 & \methodCoDALOne{} & (2.80 $\pm$ 0.68)E-4 & (3.99 $\pm$ 1.03)E-3 & (5.94 $\pm$ 1.49)E-3 & (4.91 $\pm$ 1.18)E-4 & (7.85 $\pm$ 1.57)E-3 & (1.10 $\pm$ 0.28)E-2 \\
 & \methodCoDALTwo{} & (3.05 $\pm$ 0.56)E-4 & (4.44 $\pm$ 1.05)E-3 & (6.50 $\pm$ 1.59)E-3 & (4.14 $\pm$ 0.72)E-4 & (7.21 $\pm$ 0.79)E-3 & (9.04 $\pm$ 1.20)E-3 \\
\rowcolor{gray!15}
\cellcolor{white} & \methodPARmD{} & (2.53 $\pm$ 0.41)E-4 & (3.62 $\pm$ 0.71)E-3 & (5.41 $\pm$ 0.95)E-3 & (4.91 $\pm$ 0.35)E-4 & (7.66 $\pm$ 0.57)E-3 & (9.89 $\pm$ 0.56)E-3 \\
\hline
\multirow{7}{*}{\begin{tabular}[c]{@{}c@{}}Potential\\congestion\end{tabular}} & \methodSINDy{} & \textbf{(7.25 $\pm$ 0.00)E-6} & \textbf{(1.10 $\pm$ 0.00)E-4} & \textbf{(1.43 $\pm$ 0.00)E-4} & \underline{(1.15 $\pm$ 0.00)E-4} & \underline{(2.49 $\pm$ 0.00)E-3} & \underline{(2.23 $\pm$ 0.00)E-3} \\
 & \methodSIAR{} & \underline{(2.80 $\pm$ 0.00)E-5} & \underline{(5.27 $\pm$ 0.00)E-4} & \underline{(5.53 $\pm$ 0.00)E-4} & \textbf{(9.03 $\pm$ 0.00)E-5} & \textbf{(1.48 $\pm$ 0.00)E-3} & \textbf{(1.80 $\pm$ 0.00)E-3} \\
 & \methodNeuralODE{} & (8.41 $\pm$ 0.89)E-4 & (1.13 $\pm$ 0.12)E-2 & (1.61 $\pm$ 0.17)E-2 & (6.82 $\pm$ 0.33)E-4 & (8.61 $\pm$ 0.38)E-3 & (1.27 $\pm$ 0.06)E-2 \\
 & \methodLEADS{} & (8.33 $\pm$ 0.48)E-4 & (1.12 $\pm$ 0.07)E-2 & (1.59 $\pm$ 0.09)E-2 & (7.06 $\pm$ 0.47)E-4 & (8.91 $\pm$ 0.52)E-3 & (1.32 $\pm$ 0.10)E-2 \\
 & \methodCoDALOne{} & (8.34 $\pm$ 0.30)E-4 & (1.12 $\pm$ 0.04)E-2 & (1.59 $\pm$ 0.06)E-2 & (6.74 $\pm$ 0.63)E-4 & (8.63 $\pm$ 0.71)E-3 & (1.25 $\pm$ 0.12)E-2 \\
 & \methodCoDALTwo{} & (8.34 $\pm$ 0.30)E-4 & (1.12 $\pm$ 0.04)E-2 & (1.59 $\pm$ 0.06)E-2 & (6.75 $\pm$ 0.62)E-4 & (8.63 $\pm$ 0.70)E-3 & (1.25 $\pm$ 0.12)E-2 \\
\rowcolor{gray!15}
\cellcolor{white} & \methodPARmD{} & (1.31 $\pm$ 0.22)E-4 & (2.21 $\pm$ 0.37)E-3 & (2.56 $\pm$ 0.43)E-3 & (1.79 $\pm$ 0.14)E-4 & (2.49 $\pm$ 0.22)E-3 & (3.39 $\pm$ 0.27)E-3 \\
\hline
\end{tabular}%
}
\end{table}

\begin{table}[H]
\renewcommand{\underline}[1]{\begingroup\setbox0=\hbox{#1}\leavevmode\vtop{\offinterlineskip\copy0\kern0.25pt\hrule height0.3pt width\wd0}\endgroup}
\centering
\caption{Results on seen training payoff environments under projection dynamics. Each entry is mean $\pm$ std over five runs.}
\label{tab:app-synthetic-seen-projection}
\scriptsize
\renewcommand{\arraystretch}{1.12}
\setlength{\tabcolsep}{3pt}
\resizebox{\linewidth}{!}{%
\begin{tabular}{c|c|ccc|ccc}
\hline
\multirow{2}{*}{Game} & \multirow{2}{*}{Method} & \multicolumn{3}{c|}{$K=3$} & \multicolumn{3}{c}{$K=4$} \\
\cline{3-8}
& & One-step ($\downarrow$) & Rollout RMSE ($\downarrow$) & Rollout@$T$ ($\downarrow$) & One-step ($\downarrow$) & Rollout RMSE ($\downarrow$) & Rollout@$T$ ($\downarrow$) \\
\hline
\multirow{7}{*}{\begin{tabular}[c]{@{}c@{}}Zero-sum\\cyclic\end{tabular}} & \methodSINDy{} & (2.55 $\pm$ 0.00)E-3 & (1.74 $\pm$ 0.00)E-2 & (2.47 $\pm$ 0.00)E-2 & (4.20 $\pm$ 0.00)E-3 & (3.55 $\pm$ 0.00)E-2 & (3.86 $\pm$ 0.00)E-2 \\
 & \methodSIAR{} & (8.69 $\pm$ 0.00)E-3 & (4.88 $\pm$ 0.00)E-2 & (6.09 $\pm$ 0.00)E-2 & (7.68 $\pm$ 0.00)E-3 & (4.43 $\pm$ 0.00)E-2 & (5.54 $\pm$ 0.00)E-2 \\
 & \methodNeuralODE{} & (3.68 $\pm$ 1.19)E-4 & (2.91 $\pm$ 0.95)E-3 & (3.84 $\pm$ 1.30)E-3 & (1.26 $\pm$ 0.13)E-3 & (1.32 $\pm$ 0.14)E-2 & (1.68 $\pm$ 0.18)E-2 \\
 & \methodLEADS{} & (5.95 $\pm$ 2.30)E-4 & (4.75 $\pm$ 1.95)E-3 & (6.32 $\pm$ 2.69)E-3 & \underline{(1.07 $\pm$ 0.09)E-3} & \underline{(1.13 $\pm$ 0.09)E-2} & \underline{(1.48 $\pm$ 0.16)E-2} \\
 & \methodCoDALOne{} & \textbf{(2.63 $\pm$ 1.01)E-4} & \textbf{(1.98 $\pm$ 0.85)E-3} & \textbf{(2.70 $\pm$ 1.03)E-3} & (1.24 $\pm$ 0.05)E-3 & (1.32 $\pm$ 0.07)E-2 & (1.65 $\pm$ 0.11)E-2 \\
 & \methodCoDALTwo{} & \underline{(3.27 $\pm$ 0.59)E-4} & \underline{(2.50 $\pm$ 0.53)E-3} & \underline{(3.40 $\pm$ 0.70)E-3} & \textbf{(6.09 $\pm$ 1.23)E-4} & \textbf{(5.78 $\pm$ 1.30)E-3} & \textbf{(7.72 $\pm$ 2.08)E-3} \\
\rowcolor{gray!15}
\cellcolor{white} & \methodPARmD{} & (1.02 $\pm$ 0.26)E-3 & (7.26 $\pm$ 2.46)E-3 & (1.08 $\pm$ 0.39)E-2 & (2.75 $\pm$ 0.12)E-3 & (3.73 $\pm$ 0.52)E-2 & (3.81 $\pm$ 0.27)E-2 \\
\hline
\multirow{7}{*}{\begin{tabular}[c]{@{}c@{}}Potential\\congestion\end{tabular}} & \methodSINDy{} & \textbf{(5.32 $\pm$ 0.00)E-7} & \textbf{(7.74 $\pm$ 0.00)E-6} & \textbf{(7.50 $\pm$ 0.00)E-6} & \underline{(9.60 $\pm$ 0.00)E-5} & \underline{(1.39 $\pm$ 0.00)E-3} & \underline{(1.46 $\pm$ 0.00)E-3} \\
 & \methodSIAR{} & (1.53 $\pm$ 0.00)E-5 & (3.19 $\pm$ 0.00)E-4 & (2.29 $\pm$ 0.00)E-4 & (9.09 $\pm$ 0.00)E-4 & (1.19 $\pm$ 0.00)E-2 & (1.41 $\pm$ 0.00)E-2 \\
 & \methodNeuralODE{} & (0.82 $\pm$ 1.13)E-4 & (0.85 $\pm$ 1.12)E-3 & (1.32 $\pm$ 1.84)E-3 & (3.44 $\pm$ 0.34)E-4 & (3.45 $\pm$ 0.24)E-3 & (4.71 $\pm$ 0.52)E-3 \\
 & \methodLEADS{} & (7.84 $\pm$ 4.51)E-5 & (8.20 $\pm$ 4.63)E-4 & (1.24 $\pm$ 0.73)E-3 & (4.92 $\pm$ 1.32)E-4 & (4.89 $\pm$ 1.05)E-3 & (6.75 $\pm$ 1.88)E-3 \\
 & \methodCoDALOne{} & (7.22 $\pm$ 5.49)E-5 & (7.51 $\pm$ 5.15)E-4 & (1.13 $\pm$ 0.87)E-3 & (4.16 $\pm$ 1.90)E-4 & (4.13 $\pm$ 1.56)E-3 & (5.72 $\pm$ 2.68)E-3 \\
 & \methodCoDALTwo{} & (5.12 $\pm$ 0.96)E-5 & (5.48 $\pm$ 0.74)E-4 & (8.19 $\pm$ 1.50)E-4 & (3.72 $\pm$ 1.13)E-4 & (3.81 $\pm$ 1.00)E-3 & (5.04 $\pm$ 1.63)E-3 \\
\rowcolor{gray!15}
\cellcolor{white} & \methodPARmD{} & \underline{(1.55 $\pm$ 1.84)E-6} & \underline{(1.94 $\pm$ 2.11)E-5} & \underline{(2.27 $\pm$ 2.83)E-5} & \textbf{(4.61 $\pm$ 0.31)E-5} & \textbf{(6.27 $\pm$ 0.24)E-4} & \textbf{(6.60 $\pm$ 0.54)E-4} \\
\hline
\end{tabular}%
}
\end{table}

\begin{table}[H]
\renewcommand{\underline}[1]{\begingroup\setbox0=\hbox{#1}\leavevmode\vtop{\offinterlineskip\copy0\kern0.25pt\hrule height0.3pt width\wd0}\endgroup}
\centering
\caption{Results on seen training payoff environments under MWU dynamics. Each entry is mean $\pm$ std over five runs.}
\label{tab:app-synthetic-seen-mwu}
\scriptsize
\renewcommand{\arraystretch}{1.12}
\setlength{\tabcolsep}{3pt}
\resizebox{\linewidth}{!}{%
\begin{tabular}{c|c|ccc|ccc}
\hline
\multirow{2}{*}{Game} & \multirow{2}{*}{Method} & \multicolumn{3}{c|}{$K=3$} & \multicolumn{3}{c}{$K=4$} \\
\cline{3-8}
& & One-step ($\downarrow$) & Rollout RMSE ($\downarrow$) & Rollout@$T$ ($\downarrow$) & One-step ($\downarrow$) & Rollout RMSE ($\downarrow$) & Rollout@$T$ ($\downarrow$) \\
\hline
\multirow{7}{*}{\begin{tabular}[c]{@{}c@{}}Zero-sum\\cyclic\end{tabular}} & \methodSINDy{} & \underline{(1.99 $\pm$ 0.00)E-4} & \textbf{(1.78 $\pm$ 0.00)E-3} & \textbf{(2.83 $\pm$ 0.00)E-3} & \textbf{(1.06 $\pm$ 0.00)E-4} & \textbf{(2.16 $\pm$ 0.00)E-3} & \textbf{(2.35 $\pm$ 0.00)E-3} \\
 & \methodSIAR{} & (3.84 $\pm$ 0.00)E-4 & (5.28 $\pm$ 0.00)E-3 & (7.54 $\pm$ 0.00)E-3 & \underline{(2.98 $\pm$ 0.00)E-4} & \underline{(5.81 $\pm$ 0.00)E-3} & \underline{(6.51 $\pm$ 0.00)E-3} \\
 & \methodNeuralODE{} & (5.11 $\pm$ 2.34)E-4 & (7.77 $\pm$ 4.44)E-3 & (1.23 $\pm$ 0.73)E-2 & (6.76 $\pm$ 0.61)E-4 & (1.36 $\pm$ 0.17)E-2 & (1.60 $\pm$ 0.17)E-2 \\
 & \methodLEADS{} & (5.31 $\pm$ 1.72)E-4 & (8.38 $\pm$ 3.61)E-3 & (1.34 $\pm$ 0.64)E-2 & (7.89 $\pm$ 0.53)E-4 & (1.47 $\pm$ 0.12)E-2 & (1.78 $\pm$ 0.15)E-2 \\
 & \methodCoDALOne{} & (4.42 $\pm$ 1.25)E-4 & (6.54 $\pm$ 3.24)E-3 & (9.59 $\pm$ 4.90)E-3 & (8.20 $\pm$ 1.94)E-4 & (1.52 $\pm$ 0.26)E-2 & (1.92 $\pm$ 0.42)E-2 \\
 & \methodCoDALTwo{} & (4.40 $\pm$ 1.63)E-4 & (6.53 $\pm$ 3.56)E-3 & (9.72 $\pm$ 5.38)E-3 & (7.25 $\pm$ 0.30)E-4 & (1.47 $\pm$ 0.20)E-2 & (1.69 $\pm$ 0.10)E-2 \\
\rowcolor{gray!15}
\cellcolor{white} & \methodPARmD{} & \textbf{(1.83 $\pm$ 0.49)E-4} & \underline{(3.20 $\pm$ 1.10)E-3} & \underline{(3.51 $\pm$ 0.91)E-3} & (6.15 $\pm$ 0.86)E-4 & (8.66 $\pm$ 0.87)E-3 & (1.12 $\pm$ 0.11)E-2 \\
\hline
\multirow{7}{*}{\begin{tabular}[c]{@{}c@{}}Potential\\congestion\end{tabular}} & \methodSINDy{} & \textbf{(1.55 $\pm$ 0.00)E-5} & \textbf{(2.55 $\pm$ 0.00)E-4} & \textbf{(3.03 $\pm$ 0.00)E-4} & \textbf{(1.82 $\pm$ 0.00)E-4} & \underline{(4.42 $\pm$ 0.00)E-3} & \textbf{(3.77 $\pm$ 0.00)E-3} \\
 & \methodSIAR{} & \underline{(2.74 $\pm$ 0.00)E-5} & \underline{(5.75 $\pm$ 0.00)E-4} & \underline{(5.40 $\pm$ 0.00)E-4} & (3.87 $\pm$ 0.00)E-4 & (6.41 $\pm$ 0.00)E-3 & (7.30 $\pm$ 0.00)E-3 \\
 & \methodNeuralODE{} & (8.99 $\pm$ 2.63)E-4 & (1.16 $\pm$ 0.36)E-2 & (1.57 $\pm$ 0.49)E-2 & (1.16 $\pm$ 0.07)E-3 & (1.44 $\pm$ 0.07)E-2 & (1.98 $\pm$ 0.12)E-2 \\
 & \methodLEADS{} & (9.23 $\pm$ 2.77)E-4 & (1.19 $\pm$ 0.37)E-2 & (1.60 $\pm$ 0.53)E-2 & (1.17 $\pm$ 0.07)E-3 & (1.45 $\pm$ 0.07)E-2 & (2.00 $\pm$ 0.14)E-2 \\
 & \methodCoDALOne{} & (7.24 $\pm$ 0.84)E-4 & (9.35 $\pm$ 1.10)E-3 & (1.24 $\pm$ 0.14)E-2 & (1.08 $\pm$ 0.09)E-3 & (1.38 $\pm$ 0.08)E-2 & (1.86 $\pm$ 0.14)E-2 \\
 & \methodCoDALTwo{} & (7.45 $\pm$ 0.95)E-4 & (9.60 $\pm$ 1.27)E-3 & (1.28 $\pm$ 0.17)E-2 & (1.08 $\pm$ 0.11)E-3 & (1.37 $\pm$ 0.10)E-2 & (1.85 $\pm$ 0.17)E-2 \\
\rowcolor{gray!15}
\cellcolor{white} & \methodPARmD{} & (1.29 $\pm$ 0.28)E-4 & (2.60 $\pm$ 0.82)E-3 & (2.55 $\pm$ 0.67)E-3 & \underline{(2.14 $\pm$ 0.25)E-4} & \textbf{(3.22 $\pm$ 0.38)E-3} & \underline{(4.02 $\pm$ 0.48)E-3} \\
\hline
\end{tabular}%
}
\end{table}

\subsubsection{Comparison with Few-Shot Adapted Baselines}
\label{app:few-shot-adaptation}

To evaluate the baselines under test-time adaptation, we benchmark their few-shot variants against strictly zero-shot \methodPARmD{}. Specifically, each few-shot baseline receives one independent support trajectory per held-out payoff function and undergoes 500 adaptation steps, while \methodPARmD{} receives no target data and performs no adaptation. All methods are evaluated on identical test trajectories. \Cref{tab:app-synthetic-few-shot-summary} shows that even without test-time adaptation, zero-shot \methodPARmD{} remains highly competitive, outperforming all few-shot baselines in nearly all settings and placing second in the remaining ones.

\begin{table}[H]
\renewcommand{\underline}[1]{\begingroup\setbox0=\hbox{#1}\leavevmode\vtop{\offinterlineskip\copy0\kern0.25pt\hrule height0.3pt width\wd0}\endgroup}
\centering
\caption{Rollout RMSE on held-out payoffs across three dynamics. Each entry is mean $\pm$ std over five runs.}
\label{tab:app-synthetic-few-shot-summary}
\scriptsize
\renewcommand{\arraystretch}{1.12}
\setlength{\tabcolsep}{3pt}
\resizebox{\linewidth}{!}{%
\begin{tabular}{c|c|cc|cc|cc}
\hline
\multirow{2}{*}{Game} & \multirow{2}{*}{Method}
& \multicolumn{2}{c|}{Replicator}
& \multicolumn{2}{c|}{Projection}
& \multicolumn{2}{c}{MWU} \\
\cline{3-8}
& & $K=3$ & $K=4$ & $K=3$ & $K=4$ & $K=3$ & $K=4$ \\
\hline
\multirow{4}{*}{\begin{tabular}[c]{@{}c@{}}Zero-sum\\cyclic\end{tabular}}
& \methodLEADS{} (few-shot)
& (2.91 $\pm$ 0.37)E-1
& (1.12 $\pm$ 0.12)E-1
& \textbf{(3.87 $\pm$ 2.07)E-2}
& \textbf{(4.53 $\pm$ 0.45)E-2}
& (5.84 $\pm$ 0.31)E-1
& (2.14 $\pm$ 0.15)E-1 \\
& \methodCoDALOne{} (few-shot)
& (1.42 $\pm$ 0.13)E-1
& \underline{(9.22 $\pm$ 1.19)E-2}
& (3.66 $\pm$ 0.41)E-1
& (1.66 $\pm$ 0.21)E-1
& (2.58 $\pm$ 0.58)E-1
& (1.61 $\pm$ 0.23)E-1 \\
& \methodCoDALTwo{} (few-shot)
& \underline{(1.41 $\pm$ 0.12)E-1}
& (9.34 $\pm$ 0.99)E-2
& (3.47 $\pm$ 1.09)E-1
& (1.68 $\pm$ 0.21)E-1
& \underline{(2.58 $\pm$ 0.58)E-1}
& \underline{(1.59 $\pm$ 0.22)E-1} \\
\rowcolor{gray!15}
\cellcolor{white} & \methodPARmD{} (zero-shot)
& \textbf{(2.12 $\pm$ 0.73)E-2}
& \textbf{(1.73 $\pm$ 0.09)E-2}
& \underline{(1.37 $\pm$ 0.06)E-1}
& \underline{(1.08 $\pm$ 0.06)E-1}
& \textbf{(3.04 $\pm$ 0.29)E-2}
& \textbf{(2.58 $\pm$ 0.11)E-2} \\
\hline
\multirow{4}{*}{\begin{tabular}[c]{@{}c@{}}Potential\\congestion\end{tabular}}
& \methodLEADS{} (few-shot)
& (2.55 $\pm$ 0.48)E-2
& (2.22 $\pm$ 0.15)E-2
& \underline{(1.04 $\pm$ 0.24)E-2}
& \underline{(1.56 $\pm$ 0.10)E-2}
& (3.17 $\pm$ 0.60)E-2
& \underline{(3.99 $\pm$ 0.17)E-2} \\
& \methodCoDALOne{} (few-shot)
& \underline{(2.10 $\pm$ 0.11)E-2}
& (2.38 $\pm$ 0.19)E-2
& (3.24 $\pm$ 0.40)E-2
& (4.20 $\pm$ 0.74)E-2
& \underline{(2.79 $\pm$ 0.09)E-2}
& (4.25 $\pm$ 0.33)E-2 \\
& \methodCoDALTwo{} (few-shot)
& (2.10 $\pm$ 0.11)E-2
& \underline{(2.20 $\pm$ 0.16)E-2}
& (3.24 $\pm$ 0.33)E-2
& (4.44 $\pm$ 0.82)E-2
& (2.81 $\pm$ 0.09)E-2
& (4.29 $\pm$ 0.27)E-2 \\
\rowcolor{gray!15}
\cellcolor{white} & \methodPARmD{} (zero-shot)
& \textbf{(5.83 $\pm$ 0.28)E-3}
& \textbf{(5.13 $\pm$ 0.50)E-3}
& \textbf{(1.46 $\pm$ 0.13)E-4}
& \textbf{(8.73 $\pm$ 0.61)E-4}
& \textbf{(9.37 $\pm$ 0.65)E-3}
& \textbf{(6.57 $\pm$ 0.65)E-3} \\
\hline
\end{tabular}%
}
\end{table}

\subsection{Complete YouTube Results}
\label{app:youtube-complete}

\begin{table}[H]
\renewcommand{\underline}[1]{\begingroup\setbox0=\hbox{#1}\leavevmode\vtop{\offinterlineskip\copy0\kern0.25pt\hrule height0.3pt width\wd0}\endgroup}
\centering
\caption{YouTube experiment results on held-out countries as $q$ varies. Each entry is mean $\pm$ std across five independent repetitions, with results in each repetition averaged over ten randomly sampled country splits. One-step errors use observed engagement-derived payoff vectors, while rollout metrics use estimated country-specific payoff proxies.}
\label{tab:exp-youtube}
\scriptsize
\setlength{\tabcolsep}{2.5pt}
\resizebox{\linewidth}{!}{%
\begin{tabular}{c|ccc|ccc}
\hline
\multirow{2}{*}{Method} & \multicolumn{3}{c|}{$q=3$} & \multicolumn{3}{c}{$q=4$} \\
\cline{2-4}
\cline{5-7}
 & One-step ($\downarrow$) & Rollout RMSE ($\downarrow$) & Rollout@$T$ ($\downarrow$) & One-step ($\downarrow$) & Rollout RMSE ($\downarrow$) & Rollout@$T$ ($\downarrow$) \\
\hline
\methodSINDy{} & (9.53 $\pm$ 0.00)E-1 & (9.67 $\pm$ 0.00)E-1 & (9.70 $\pm$ 0.00)E-1 & (9.91 $\pm$ 0.00)E-1 & (1.01 $\pm$ 0.00)E+0 & (1.03 $\pm$ 0.00)E+0 \\
\methodSIAR{} & (3.91 $\pm$ 0.00)E-1 & (8.06 $\pm$ 0.00)E-1 & (8.30 $\pm$ 0.00)E-1 & (2.92 $\pm$ 0.00)E-1 & (6.69 $\pm$ 0.00)E-1 & (6.77 $\pm$ 0.00)E-1 \\
\methodNeuralODE{} & (1.13 $\pm$ 0.05)E-1 & (3.41 $\pm$ 0.26)E-1 & (4.03 $\pm$ 0.38)E-1 & \underline{(9.94 $\pm$ 0.29)E-2} & (3.31 $\pm$ 0.28)E-1 & (3.89 $\pm$ 0.45)E-1 \\
\methodLEADS{} & (1.81 $\pm$ 0.09)E-1 & (6.55 $\pm$ 0.26)E-1 & (7.18 $\pm$ 0.40)E-1 & (1.57 $\pm$ 0.12)E-1 & (6.08 $\pm$ 0.63)E-1 & (6.67 $\pm$ 0.77)E-1 \\
\methodCoDALOne{} & \underline{(1.13 $\pm$ 0.01)E-1} & (3.26 $\pm$ 0.12)E-1 & (3.74 $\pm$ 0.17)E-1 & (1.05 $\pm$ 0.04)E-1 & \underline{(3.25 $\pm$ 0.10)E-1} & (3.77 $\pm$ 0.14)E-1 \\
\methodCoDALTwo{} & (1.13 $\pm$ 0.01)E-1 & \underline{(3.26 $\pm$ 0.11)E-1} & \underline{(3.74 $\pm$ 0.15)E-1} & (1.06 $\pm$ 0.04)E-1 & (3.26 $\pm$ 0.10)E-1 & \underline{(3.77 $\pm$ 0.14)E-1} \\
\rowcolor{gray!15}
\methodPARmD{} & \textbf{(6.90 $\pm$ 0.02)E-2} & \textbf{(2.53 $\pm$ 0.07)E-1} & \textbf{(3.53 $\pm$ 0.12)E-1} & \textbf{(6.96 $\pm$ 0.01)E-2} & \textbf{(2.41 $\pm$ 0.16)E-1} & \textbf{(3.46 $\pm$ 0.29)E-1} \\
\hline
\end{tabular}%
}

\vspace{0.5em}
\resizebox{\linewidth}{!}{%
\begin{tabular}{c|ccc|ccc}
\hline
\multirow{2}{*}{Method} & \multicolumn{3}{c|}{$q=5$} & \multicolumn{3}{c}{$q=6$} \\
\cline{2-4}
\cline{5-7}
 & One-step ($\downarrow$) & Rollout RMSE ($\downarrow$) & Rollout@$T$ ($\downarrow$) & One-step ($\downarrow$) & Rollout RMSE ($\downarrow$) & Rollout@$T$ ($\downarrow$) \\
\hline
\methodSINDy{} & (9.72 $\pm$ 0.00)E-1 & (9.67 $\pm$ 0.00)E-1 & (9.77 $\pm$ 0.00)E-1 & (9.93 $\pm$ 0.00)E-1 & (1.01 $\pm$ 0.00)E+0 & (1.02 $\pm$ 0.00)E+0 \\
\methodSIAR{} & (2.24 $\pm$ 0.00)E-1 & (6.44 $\pm$ 0.00)E-1 & (7.03 $\pm$ 0.00)E-1 & (1.85 $\pm$ 0.00)E-1 & (5.71 $\pm$ 0.00)E-1 & (5.96 $\pm$ 0.00)E-1 \\
\methodNeuralODE{} & \underline{(9.70 $\pm$ 0.54)E-2} & (3.13 $\pm$ 0.20)E-1 & (3.81 $\pm$ 0.37)E-1 & \underline{(9.56 $\pm$ 0.40)E-2} & \underline{(2.99 $\pm$ 0.19)E-1} & \underline{(3.67 $\pm$ 0.36)E-1} \\
\methodLEADS{} & (1.40 $\pm$ 0.11)E-1 & (6.32 $\pm$ 0.83)E-1 & (7.19 $\pm$ 0.95)E-1 & (1.30 $\pm$ 0.09)E-1 & (6.07 $\pm$ 0.70)E-1 & (6.97 $\pm$ 0.75)E-1 \\
\methodCoDALOne{} & (1.00 $\pm$ 0.04)E-1 & (3.13 $\pm$ 0.18)E-1 & (3.69 $\pm$ 0.25)E-1 & (9.84 $\pm$ 0.45)E-2 & (3.10 $\pm$ 0.14)E-1 & (3.74 $\pm$ 0.29)E-1 \\
\methodCoDALTwo{} & (1.00 $\pm$ 0.04)E-1 & \underline{(3.12 $\pm$ 0.18)E-1} & \underline{(3.68 $\pm$ 0.25)E-1} & (9.84 $\pm$ 0.45)E-2 & (3.09 $\pm$ 0.15)E-1 & (3.74 $\pm$ 0.31)E-1 \\
\rowcolor{gray!15}
\methodPARmD{} & \textbf{(7.28 $\pm$ 0.01)E-2} & \textbf{(2.43 $\pm$ 0.13)E-1} & \textbf{(3.50 $\pm$ 0.19)E-1} & \textbf{(7.57 $\pm$ 0.00)E-2} & \textbf{(2.39 $\pm$ 0.07)E-1} & \textbf{(3.45 $\pm$ 0.11)E-1} \\
\hline
\end{tabular}%
}

\vspace{0.5em}
\resizebox{\linewidth}{!}{%
\begin{tabular}{c|ccc|ccc}
\hline
\multirow{2}{*}{Method} & \multicolumn{3}{c|}{$q=7$} & \multicolumn{3}{c}{$q=8$} \\
\cline{2-4}
\cline{5-7}
 & One-step ($\downarrow$) & Rollout RMSE ($\downarrow$) & Rollout@$T$ ($\downarrow$) & One-step ($\downarrow$) & Rollout RMSE ($\downarrow$) & Rollout@$T$ ($\downarrow$) \\
\hline
\methodSINDy{} & (9.67 $\pm$ 0.00)E-1 & (9.75 $\pm$ 0.00)E-1 & (9.78 $\pm$ 0.00)E-1 & (9.80 $\pm$ 0.00)E-1 & (9.98 $\pm$ 0.00)E-1 & (1.02 $\pm$ 0.00)E+0 \\
\methodSIAR{} & (1.63 $\pm$ 0.00)E-1 & (4.87 $\pm$ 0.00)E-1 & (5.35 $\pm$ 0.00)E-1 & (1.32 $\pm$ 0.00)E-1 & (4.86 $\pm$ 0.00)E-1 & (5.34 $\pm$ 0.00)E-1 \\
\methodNeuralODE{} & \underline{(8.93 $\pm$ 0.30)E-2} & (3.18 $\pm$ 0.17)E-1 & (3.87 $\pm$ 0.29)E-1 & \underline{(8.73 $\pm$ 0.55)E-2} & (2.91 $\pm$ 0.12)E-1 & (3.63 $\pm$ 0.30)E-1 \\
\methodLEADS{} & (1.24 $\pm$ 0.06)E-1 & (5.81 $\pm$ 0.52)E-1 & (6.68 $\pm$ 0.66)E-1 & (1.16 $\pm$ 0.10)E-1 & (5.29 $\pm$ 0.46)E-1 & (5.98 $\pm$ 0.62)E-1 \\
\methodCoDALOne{} & (9.47 $\pm$ 0.41)E-2 & (3.17 $\pm$ 0.15)E-1 & (3.78 $\pm$ 0.24)E-1 & (8.99 $\pm$ 0.50)E-2 & \underline{(2.87 $\pm$ 0.21)E-1} & (3.52 $\pm$ 0.35)E-1 \\
\methodCoDALTwo{} & (9.46 $\pm$ 0.41)E-2 & \underline{(3.16 $\pm$ 0.14)E-1} & \underline{(3.74 $\pm$ 0.22)E-1} & (9.00 $\pm$ 0.49)E-2 & (2.88 $\pm$ 0.21)E-1 & \underline{(3.51 $\pm$ 0.35)E-1} \\
\rowcolor{gray!15}
\methodPARmD{} & \textbf{(7.28 $\pm$ 0.01)E-2} & \textbf{(2.22 $\pm$ 0.06)E-1} & \textbf{(3.15 $\pm$ 0.10)E-1} & \textbf{(6.48 $\pm$ 0.01)E-2} & \textbf{(2.11 $\pm$ 0.08)E-1} & \textbf{(3.02 $\pm$ 0.11)E-1} \\
\hline
\end{tabular}%
}

\vspace{0.5em}
\resizebox{0.53\linewidth}{!}{%
\begin{tabular}{c|ccc}
\hline
\multirow{2}{*}{Method} & \multicolumn{3}{c}{$q=9$} \\
\cline{2-4}
 & One-step ($\downarrow$) & Rollout RMSE ($\downarrow$) & Rollout@$T$ ($\downarrow$) \\
\hline
\methodSINDy{} & (9.66 $\pm$ 0.00)E-1 & (1.03 $\pm$ 0.00)E+0 & (1.03 $\pm$ 0.00)E+0 \\
\methodSIAR{} & (1.35 $\pm$ 0.00)E-1 & (5.06 $\pm$ 0.00)E-1 & (6.77 $\pm$ 0.00)E-1 \\
\methodNeuralODE{} & \underline{(8.52 $\pm$ 0.32)E-2} & \underline{(3.30 $\pm$ 0.54)E-1} & \underline{(3.91 $\pm$ 0.89)E-1} \\
\methodLEADS{} & (1.17 $\pm$ 0.12)E-1 & (5.91 $\pm$ 0.48)E-1 & (6.62 $\pm$ 0.93)E-1 \\
\methodCoDALOne{} & (9.11 $\pm$ 0.63)E-2 & (3.50 $\pm$ 0.42)E-1 & (4.16 $\pm$ 0.47)E-1 \\
\methodCoDALTwo{} & (9.16 $\pm$ 0.61)E-2 & (3.54 $\pm$ 0.43)E-1 & (4.21 $\pm$ 0.47)E-1 \\
\rowcolor{gray!15}
\methodPARmD{} & \textbf{(6.82 $\pm$ 0.03)E-2} & \textbf{(2.38 $\pm$ 0.34)E-1} & \textbf{(3.41 $\pm$ 0.54)E-1} \\
\hline
\end{tabular}%
}

\end{table}

\subsection{Payoff-Dependent Response Perturbations}
\label{app:response-perturbation}

We test departures from a shared response under replicator dynamics with $K=4$.
Write each payoff function as $v(x)=Px+q$.
We first factor the original payoff-independent response matrix as
$$
G_0(x)=L_0(x)L_0(x)^\top.
$$
We then define the payoff-dependent perturbed response as
$$
G_{\alpha}(x) = 
L_0(x) \bigl(I_K+\alpha PP^\top\bigr) L_0(x)^\top,
\qquad \alpha\geq0.
$$
This recovers $G_0(x)$ at $\alpha=0$ and preserves positive definiteness at interior states.
\Cref{tab:response-perturbation-replicator-k4} highlights the
robustness of \methodPARmD{} to violations of the shared-response
assumption: it maintains the lowest rollout RMSE under moderate
perturbations ($\alpha \le 0.8$) and remains among the top two
methods under stronger perturbations ($\alpha=1.6, 3.2$).

\begin{table}[H]
\renewcommand{\underline}[1]{\begingroup\setbox0=\hbox{#1}\leavevmode\vtop{\offinterlineskip\copy0\kern0.25pt\hrule height0.3pt width\wd0}\endgroup}
\centering
\caption{Rollout RMSE under response perturbations. Each entry is mean $\pm$ std over five runs.}
\label{tab:response-perturbation-replicator-k4}
\scriptsize
\renewcommand{\arraystretch}{1.12}
\setlength{\tabcolsep}{3pt}
\resizebox{\linewidth}{!}{%
\begin{tabular}{c|c|ccccc}
\hline
Game & Method & $\alpha=0$ & $\alpha=0.4$ & $\alpha=0.8$ & $\alpha=1.6$ & $\alpha=3.2$ \\
\hline
 & \methodSINDy{} & (2.88 $\pm$ 0.00)E-1 & (5.76 $\pm$ 0.00)E-1 & (6.49 $\pm$ 0.00)E-1 & (7.47 $\pm$ 0.00)E-1 & (7.37 $\pm$ 0.00)E-1 \\
 & \methodSIAR{} & (2.11 $\pm$ 0.00)E-1 & (2.41 $\pm$ 0.00)E-1 & \underline{(2.77 $\pm$ 0.00)E-1} & \textbf{(3.29 $\pm$ 0.00)E-1} & \textbf{(4.65 $\pm$ 0.00)E-1} \\
 & \methodNeuralODE{} & \underline{(8.72 $\pm$ 0.62)E-2} & (2.15 $\pm$ 0.18)E-1 & (3.67 $\pm$ 0.40)E-1 & (5.82 $\pm$ 0.23)E-1 & (7.27 $\pm$ 0.14)E-1 \\
 & \methodLEADS{} & (9.71 $\pm$ 1.56)E-2 & (1.95 $\pm$ 0.10)E-1 & (3.12 $\pm$ 0.47)E-1 & (5.10 $\pm$ 0.72)E-1 & (7.20 $\pm$ 0.35)E-1 \\
 & \methodCoDALOne{} & (9.38 $\pm$ 0.94)E-2 & (1.89 $\pm$ 0.15)E-1 & (3.21 $\pm$ 0.21)E-1 & (5.45 $\pm$ 0.35)E-1 & (7.35 $\pm$ 0.27)E-1 \\
 & \methodCoDALTwo{} & (9.34 $\pm$ 0.99)E-2 & \underline{(1.79 $\pm$ 0.13)E-1} & (3.29 $\pm$ 0.15)E-1 & (5.44 $\pm$ 0.27)E-1 & (7.44 $\pm$ 0.24)E-1 \\
\rowcolor{gray!15}
\cellcolor{white}\multirow{-7}{*}{\begin{tabular}[c]{@{}c@{}}\cellcolor{white}Zero-sum\\\cellcolor{white}cyclic\end{tabular}}
 & \methodPARmD{} & \textbf{(1.73 $\pm$ 0.09)E-2} & \textbf{(7.53 $\pm$ 0.39)E-2} & \textbf{(1.90 $\pm$ 0.17)E-1} & \underline{(4.26 $\pm$ 0.11)E-1} & \underline{(6.47 $\pm$ 0.13)E-1} \\
\hline
 & \methodSINDy{} & (7.07 $\pm$ 0.00)E-2 & (7.00 $\pm$ 0.00)E-2 & (8.25 $\pm$ 0.00)E-2 & (1.16 $\pm$ 0.00)E-1 & (5.95 $\pm$ 0.00)E-1 \\
 & \methodSIAR{} & (5.17 $\pm$ 0.00)E-2 & (7.38 $\pm$ 0.00)E-2 & (1.27 $\pm$ 0.00)E-1 & (8.47 $\pm$ 0.00)E-2 & (1.13 $\pm$ 0.00)E-1 \\
 & \methodNeuralODE{} & \underline{(2.21 $\pm$ 0.07)E-2} & \underline{(2.16 $\pm$ 0.06)E-2} & \underline{(2.12 $\pm$ 0.07)E-2} & \underline{(2.08 $\pm$ 0.09)E-2} & \underline{(2.14 $\pm$ 0.10)E-2} \\
 & \methodLEADS{} & (2.42 $\pm$ 0.17)E-2 & (2.38 $\pm$ 0.17)E-2 & (2.35 $\pm$ 0.16)E-2 & (2.30 $\pm$ 0.13)E-2 & (2.31 $\pm$ 0.07)E-2 \\
 & \methodCoDALOne{} & (2.38 $\pm$ 0.19)E-2 & (2.34 $\pm$ 0.19)E-2 & (2.30 $\pm$ 0.19)E-2 & (2.22 $\pm$ 0.18)E-2 & (2.22 $\pm$ 0.14)E-2 \\
 & \methodCoDALTwo{} & (2.38 $\pm$ 0.20)E-2 & (2.34 $\pm$ 0.20)E-2 & (2.29 $\pm$ 0.19)E-2 & (2.21 $\pm$ 0.16)E-2 & (2.22 $\pm$ 0.14)E-2 \\
\rowcolor{gray!15}
\cellcolor{white}\multirow{-7}{*}{\begin{tabular}[c]{@{}c@{}}\cellcolor{white}Potential\\\cellcolor{white}congestion\end{tabular}}
 & \methodPARmD{} & \textbf{(5.13 $\pm$ 0.50)E-3} & \textbf{(4.85 $\pm$ 0.44)E-3} & \textbf{(5.11 $\pm$ 0.41)E-3} & \textbf{(6.84 $\pm$ 0.32)E-3} & \textbf{(1.09 $\pm$ 0.04)E-2} \\
\hline
\end{tabular}%
}
\end{table}

\section{Limitations and Broader Impact} \label{sec:limitations}

\paragraph{Limitations.}
\methodPARmD{} is designed for payoff-aware prediction in settings where changing payoffs are the main source of variation, and our experiments show that this structure is effective across several synthetic game classes and a real-world population dynamics dataset. The current analysis focuses on the natural regime in which the population response structure is stable across payoff functions and the held-out payoff functions are reasonably covered by the training payoff functions. Future work could broaden this scope to settings where a new incentive design also changes information, constraints, or adjustment behavior. In the real-data experiment, population states represent category shares among trending-video records, and payoffs are approximated using engagement-derived proxies. The behavioral interpretation of these constructed quantities is limited because the available data do not directly track strategy revisions of content creators at either the individual or population level.

\paragraph{Broader impact.}
This work is a methodological contribution and is not tied to any particular deployment. Payoff-aware prediction may help evaluate incentive designs before deployment, which can reduce costly trial-and-error interventions and improve transparency in mechanism design. As with any tool for studying incentives, deployment in sensitive settings should be accompanied by careful validation of payoff proxies, privacy-preserving data practices, and attention to downstream welfare and fairness effects.

\newpage
\section*{NeurIPS Paper Checklist}

\begin{enumerate}

\item {\bf Claims}
    \item[] Question: Do the main claims made in the abstract and introduction accurately reflect the paper's contributions and scope?
    \item[] Answer: \answerYes{} 
    \item[] Justification: The main claims in the abstract and introduction outline our main ideas, and accurately reflect the core contributions and scope of our work.

\item {\bf Limitations}
    \item[] Question: Does the paper discuss the limitations of the work performed by the authors?
    \item[] Answer: \answerYes{} 
    \item[] Justification: We discuss the limitations in \cref{sec:limitations}.

\item {\bf Theory assumptions and proofs}
    \item[] Question: For each theoretical result, does the paper provide the full set of assumptions and a complete (and correct) proof?
    \item[] Answer: \answerYes{} 
    \item[] Justification: Assumptions and proofs for all theoretical results are provided in \cref{sec:model,sec:proofs}.

    \item {\bf Experimental result reproducibility}
    \item[] Question: Does the paper fully disclose all the information needed to reproduce the main experimental results of the paper to the extent that it affects the main claims and/or conclusions of the paper (regardless of whether the code and data are provided or not)?
    \item[] Answer: \answerYes{} 
    \item[] Justification: Experimental settings and implementations are described in \cref{sec:experiments,sec:appendix-exp}. Codes are provided as the supplement material.

\item {\bf Open access to data and code}
    \item[] Question: Does the paper provide open access to the data and code, with sufficient instructions to faithfully reproduce the main experimental results, as described in supplemental material?
    \item[] Answer: \answerYes{} 
    \item[] Justification: Synthetic datasets can be reproduced by following the instructions, while the real-world dataset is public with citations. See \cref{sec:experiments,sec:appendix-exp} for details. Codes are provided as the supplement material.

\item {\bf Experimental setting/details}
    \item[] Question: Does the paper specify all the training and test details (e.g., data splits, hyperparameters, how they were chosen, type of optimizer) necessary to understand the results?
    \item[] Answer: \answerYes{} 
    \item[] Justification: Experimental settings and implementations are described in \cref{sec:experiments,sec:appendix-exp}.

\item {\bf Experiment statistical significance}
    \item[] Question: Does the paper report error bars suitably and correctly defined or other appropriate information about the statistical significance of the experiments?
    \item[] Answer: \answerYes{} 
    \item[] Justification: We report results using mean $\pm$ standard deviation across five independent runs.

\item {\bf Experiments compute resources}
    \item[] Question: For each experiment, does the paper provide sufficient information on the computer resources (type of compute workers, memory, time of execution) needed to reproduce the experiments?
    \item[] Answer: \answerYes{} 
    \item[] Justification: We report the compute resources needed to reproduce each method-setting run in \cref{app:baselines}.
    
\item {\bf Code of ethics}
    \item[] Question: Does the research conducted in the paper conform, in every respect, with the NeurIPS Code of Ethics \url{https://neurips.cc/public/EthicsGuidelines}?
    \item[] Answer: \answerYes{} 
    \item[] Justification: The research conducted in the paper conforms with the NeurIPS Code of Ethics.

\item {\bf Broader impacts}
    \item[] Question: Does the paper discuss both potential positive societal impacts and negative societal impacts of the work performed?
    \item[] Answer: \answerYes{} 
    \item[] Justification: We discuss potential positive and negative broader impacts in \cref{sec:limitations}.
    
\item {\bf Safeguards}
    \item[] Question: Does the paper describe safeguards that have been put in place for responsible release of data or models that have a high risk for misuse (e.g., pre-trained language models, image generators, or scraped datasets)?
    \item[] Answer: \answerNA{} 
    \item[] Justification: This paper poses no safeguard risks.

\item {\bf Licenses for existing assets}
    \item[] Question: Are the creators or original owners of assets (e.g., code, data, models), used in the paper, properly credited and are the license and terms of use explicitly mentioned and properly respected?
    \item[] Answer: \answerYes{} 
    \item[] Justification: All existing assets used in the paper are properly credited. Codes are are credited with original licenses provided.

\item {\bf New assets}
    \item[] Question: Are new assets introduced in the paper well documented and is the documentation provided alongside the assets?
    \item[] Answer: \answerNA{} 
    \item[] Justification: No new assets are introduced in the paper.

\item {\bf Crowdsourcing and research with human subjects}
    \item[] Question: For crowdsourcing experiments and research with human subjects, does the paper include the full text of instructions given to participants and screenshots, if applicable, as well as details about compensation (if any)? 
    \item[] Answer: \answerNA{} 
    \item[] Justification: This work does not involve crowdsourcing or research with human subjects.

\item {\bf Institutional review board (IRB) approvals or equivalent for research with human subjects}
    \item[] Question: Does the paper describe potential risks incurred by study participants, whether such risks were disclosed to the subjects, and whether Institutional Review Board (IRB) approvals (or an equivalent approval/review based on the requirements of your country or institution) were obtained?
    \item[] Answer: \answerNA{} 
    \item[] Justification: This work does not involve crowdsourcing or research with human subjects.

\item {\bf Declaration of LLM usage}
    \item[] Question: Does the paper describe the usage of LLMs if it is an important, original, or non-standard component of the core methods in this research? Note that if the LLM is used only for writing, editing, or formatting purposes and does \emph{not} impact the core methodology, scientific rigor, or originality of the research, declaration is not required.
    \item[] Answer: \answerNA{} 
    \item[] Justification: Core method development in this research does not involve LLMs.

\end{enumerate}

\noindent

\end{document}